%% file: main.tex
\documentclass[runningheads]{llncs}

\usepackage{eccv}

\usepackage{eccvabbrv}

\usepackage{graphicx}
\usepackage{booktabs}
\usepackage{kotex}
\usepackage[accsupp]{axessibility}  
\usepackage[most]{tcolorbox}
\usepackage{enumitem}

\usepackage{caption}
\usepackage[table]{xcolor} 
\usepackage{todonotes}
\usepackage{colortbl}
\usepackage{subcaption}
\input{math_commands.tex}

\definecolor{cvprblue}{rgb}{0.21,0.49,0.74}
\definecolor{cornellred}{rgb}{0.7, 0.11, 0.11}
\definecolor{cadmiumgreen}{rgb}{0.0, 0.42, 0.24}
\definecolor{aliceblue}{rgb}{0.91, 0.94, 0.97}
\definecolor{darkblue}{rgb}{0.83, 0.89, 0.97}
\definecolor{Red7}{rgb}{0.941, 0.243, 0.243}
\definecolor{Green7}{RGB}{55, 178, 77}
\definecolor{Blue9}{rgb}{0.098,0.3,0.9}
\definecolor{darkergreen}{RGB}{21, 152, 56}
\definecolor{red2}{RGB}{252, 54, 65}
\definecolor{myteal}{RGB}{0,78,77}
\definecolor{sectiongray}{HTML}{E6E6E6}
\definecolor{mypink}{RGB}{253,231,230}

\usepackage{orcidlink}

\hypersetup{
    colorlinks=true,
    linkcolor=cornellred,   
    urlcolor=Blue9,
    citecolor=cadmiumgreen
}

\begin{document}

\title{Representation Alignment for Just Image Transformers is not Easier than You Think}

\titlerunning{PixelREPA}

\author{$\text{Jaeyo Shin}^{\dagger}$\orcidlink{0009-0005-3931-938X}  \and
$\text{Jiwook Kim}^{\dagger}$\orcidlink{0009-0005-3796-8060} \and
Hyunjung Shim\orcidlink{0000-0001-6796-1058}
}

\authorrunning{J. Shin, J. Kim, and H. Shim.}

\institute{
$^{\dagger}$ indicates equal contributions.\\[1em]
KAIST AI, \\
291 Daehak-ro, Yuseong-gu, Daejeon 34141, Republic of Korea
\email{\{jaeyo\_shin,tom919,kateshim\}@kaist.ac.kr}
}

\maketitle

\renewcommand{\thefootnote}{}
\footnotetext{Our code is available at \url{https://github.com/kaist-cvml/PixelREPA}.}

\begin{abstract}
  Representation Alignment (REPA) has emerged as a simple way to accelerate Diffusion Transformers training in latent space.
  At the same time, pixel-space diffusion transformers such as Just image Transformers (JiT) have attracted growing attention because they remove a dependency on a pretrained tokenizer, and then avoid the reconstruction bottleneck of latent diffusion.
  This paper shows that the REPA can fail for JiT.
  REPA yields worse FID for JiT as training proceeds and collapses diversity on image subsets that are tightly clustered in the representation space of pretrained semantic encoder on ImageNet.
  We trace the failure to an information asymmetry: denoising occurs in the high dimensional image space, while the semantic target is strongly compressed, making direct regression a shortcut objective.
  We propose PixelREPA, which transforms the alignment target and constrains alignment with a Masked Transformer Adapter that combines a shallow transformer adapter with partial token masking.
  PixelREPA improves both training convergence and final quality.
  PixelREPA reduces FID from 3.66 to 3.17 for JiT-B$/16$ and improves Inception Score (IS) from 275.1 to 284.6 on ImageNet $256 \times 256$, while achieving $> 2\times$ faster convergence.
  Finally, PixelREPA-H$/16$ achieves FID$=1.81$ and IS$=317.2$.
  \keywords{Pixel space diffusion model \and Representation Alignment}
\end{abstract}

\begin{figure}[th]
  \centering
  \includegraphics[width=0.8\textwidth]{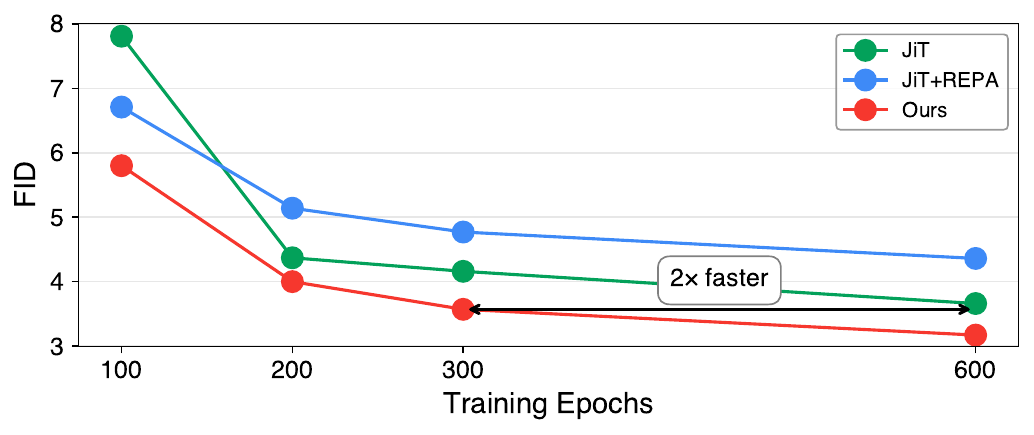}
  \vspace{-1em}
  \caption{
  \textbf{REPA degrades JiT performance.}
  As training progresses, JiT$+$REPA yields higher FID~\cite{heusel2017gans} ($\downarrow$) than vanilla JiT, indicating that REPA hinders pixel space diffusion training.
  PixelREPA prevents overfitting to the external semantic feature target, which accelerates convergence in JiT training.
  Remarkably, PixelREPA achieves $>2 \times$ faster convergence than the vanilla JiT.
  All evaluated models utilize JiT-B$/16$.
  }
  \label{fig:main}
  \vspace{-1em}
\end{figure}

\section{Introduction}
\label{sec:intro}
\input{sections/01_introduction}

\section{Preliminaries}
\label{sec:pre}
\input{sections/02_preliminaries}

\section{Motivation}
\label{sec:motivation}
\input{sections/03_motivations}

\section{PixelREPA: REPA for Pixel Space Diffusion Models}
\label{sec:meth}
\input{sections/04_method}

\section{Experiments}
\label{sec:meth}
\input{sections/05_experiments}

\section{Conclusion}
\label{sec:conclusion}
\input{sections/06_conclusion}

\clearpage  

\bibliographystyle{splncs04}
\bibliography{main}

\clearpage

\input{sections/07_supple}

\end{document}

%% file: math_commands.tex
\usepackage{amsmath,amsfonts,bm}

\def\eqref#1{equation~\ref{#1}}

\def\1{\bm{1}}

\def\rvx{{\mathbf{x}}}

\def\vzero{{\bm{0}}}

\def\vtheta{{\bm{\theta}}}
\def\vepsilon{{\bm{\epsilon}}}

\def\vh{{\bm{h}}}

\def\vv{{\bm{v}}}

\def\vx{{\bm{x}}}

\def\evbeta{{\beta}}

\def\evsigma{{\sigma}}

\def\mI{{\bm{I}}}

\DeclareMathAlphabet{\mathsfit}{\encodingdefault}{\sfdefault}{m}{sl}
\SetMathAlphabet{\mathsfit}{bold}{\encodingdefault}{\sfdefault}{bx}{n}

\def\gH{{\mathcal{H}}}

\def\gL{{\mathcal{L}}}

\def\gN{{\mathcal{N}}}

\def\gR{{\mathcal{R}}}

\def\gX{{\mathcal{X}}}

\newcommand{\pdata}{p_{\rm{data}}}

\newcommand{\E}{\mathbb{E}}
\newcommand{\Ls}{\mathcal{L}}



%% file: sections/01_introduction.tex
Diffusion models~\cite{sohl2015deep,song2019generative,ho2020denoising,rombach2022high} can be categorized by the choice of data space in which denoising is performed.
Latent Diffusion Models (LDMs)~\cite{rombach2022high} reduce computation by mapping pixels into a learned latent space via a pretrained image tokenizer. However, this choice couples the achievable generation quality to the capacity and reconstruction fidelity of the tokenizer: strong compression attenuates fine textures and small structures, imposing an upper bound on what the generator can express~\cite{gu2024rethinking,blau2018perception}.
Just Image Transformers (JiT)~\cite{li2025back} revisits pixel-space diffusion~\cite{ho2020denoising,song2019generative,dhariwal2021diffusion} and shows that a plain Vision Transformer (ViT)~\cite{dosovitskiy2020image} can be trained end-to-end on raw images without any latent tokenizer or auxiliary objectives such as adversarial~\cite{goodfellow2014generative} and perceptual~\cite{zhang2018unreasonable} losses, while still achieving strong generation performance.
By removing the dependency on the pretrained tokenizer, pixel-space diffusion eliminates the reconstruction bottleneck and opens a path toward fully self-contained diffusion pipelines that can, in principle, represent arbitrary high-frequency detail.

Training such models, however, remains expensive. 
In parallel with efforts on pixel-space diffusion, a complementary line of work seeks to accelerate latent Diffusion Transformers (DiT)~\cite{peebles2023scalable} training by injecting semantic structure from large representation encoders.
Representation Alignment (REPA)~\cite{yu2024representation} aligns intermediate DiT activations with features from an external semantic encoder such as DINOv2~\cite{oquab2023dinov2}, providing an explicit semantic target and dramatically speeds up convergence. Because pixel-space diffusion faces a similar, often more severe training cost, applying REPA to JiT is a natural next step.

However, we observe the opposite tendency in pixel space, as shown in \cref{fig:main} (JiT$+$REPA). 
REPA unexpectedly degrades performance when pixel-space diffusion training progresses.
This observation raises a natural question:
\emph{why does REPA accelerate latent-space diffusion yet hinder pixel-space diffusion?}

We trace the root cause to a fundamental \emph{information asymmetry} between the two spaces. 
In LDMs, the pretrained tokenizer compresses the image and suppresses much of the fine-scale, high-frequency variation~\cite{blau2018perception,jiang2021focal,esser2021taming}. 
The external semantic encoder is also a compressed representation that is largely insensitive to this fine detail~\cite{park2023self}. 
Because both the denoising space and the alignment target have already passed through \emph{information bottlenecks}~\cite{tishby2000information}, their degrees of freedom are roughly matched, and direct feature alignment is effective. 

In pixel space, however, denoising operates in the ambient image space with $O(H \times W)$ degrees of freedom, while the semantic encoder still produces a compact, bottleneck representation. 
Accordingly, many pixel-distinct images therefore map to similar regions in feature space of the semantic encoder, and this ambiguity grows with resolution. 
Forcing the diffusion model to regress toward such a compressed target leads to \emph{feature hacking}: the model overfits to the narrow external feature space and loses the ability to generate diverse images whose semantic features are highly similar. 
Our experiments confirm this analysis. 
REPA improves JiT at $32\times 32$ resolution where the pixel-feature gap is small, but consistently degrades performance at $256\times 256$ where this gap is large. 
Furthermore, JiT$+$REPA shows degraded FID compared to vanilla JiT specifically on image subsets that are tightly clustered in feature space of semantic encoder yet visually diverse in pixel space, directly evidencing feature hacking. 

These findings reveal that the target of alignment matters.
Standard REPA projects diffusion features into the semantic space through a point-wise Multi-Layer Perceptron (MLP) and matches them to feature space of the semantic encoder.
This is effectively a feature to pixel alignment: it asks the pixel-space model to conform to a compressed feature target. 
When the information gap between the two spaces is large, the original REPA formulation trivially minimizes direct regression to feature space of the semantic encoder, collapsing diversity. 
As a result, REPA encourages intermediate JiT representations to collapse toward semantic feature.
Later blocks must then reconstruct pixels from a compressed semantic feature.
This semantic to image direction is ill-posed in pixel space, because many distinct images map to similar semantic features~\cite{blau2018perception}.

We transform this target. 
Rather than forcing pixel representations to match a compressed target, we map them into the semantic feature space via a \textbf{\emph{shallow Transformer adapter}} and align them to transformed space induced by the adapter.
Concretely, we extract an intermediate representation from JiT encoder, pass it through a lightweight two-block Transformer adapter, and align the adapter output with features of the frozen semantic encoder.
This adapter is trained to transform intermediate JiT features toward the semantic target to prevent feature hacking.
This preserves the information needed for subsequent JiT blocks to map back to pixels while \emph{selectively} injecting semantic structure into JiT representation.
Furthermore, the adapter performs contextual aggregation via self-attention, so each token prediction can leverage information from neighboring tokens before matching $f(\cdot)$, reducing reliance on purely local cues. 

A critical design choice accompanies this adapter. 
Without additional constraints, we find that the adapter can still learn a trivial token-wise mapping that shortcuts directly to the compressed target -- empirically, an unmasked adapter improves over REPA but still falls short of vanilla JiT. 
To prevent this shortcut, we apply random \textbf{\emph{partial masking}} to the adapter input. 
Masking serves two complementary roles. 
First, by removing a subset of tokens, it forces the adapter to predict the target representation under partial observation, which requires genuine contextual reasoning rather than trivial per-token projection~\cite{he2022masked}. 
Second, masking acts as an information bottleneck on the pixel side: it reduces the effective degrees of freedom of the pixel representation before alignment, narrowing the information gap between pixel features and the compressed semantic target. 
This makes the two spaces more compatible--analogous to the role the tokenizer plays in latent diffusion-- without discarding information in the main denoising pathway. 
Together, the adapter and masking form the \textbf{\emph{Masked Transformer Adapter (MTA)}}, which turns alignment into a constrained prediction problem well-suited to high-resolution pixel-space diffusion.

This design differs from standard REPA in both the alignment module architecture and the training-time masking mechanism.
REPA aligns patch-wise projections of diffusion hidden states to pretrained visual features using a trainable projection head, implemented as a MLP.
Our approach replaces this MLP projection with a shallow Transformer adapter and introduces masking on the adapter input, motivated by the pixel space failure mode where a strongly compressed external target can cause direct alignment to overemphasize feature matching. Importantly, MTA is applied only on the alignment branch and does not modify the main denoising pathway; it is used only during training and therefore incurs no additional cost at inference.

In this study, we propose \textbf{\emph{PixelREPA}}, a REPA-style alignment framework designed for pixel space diffusion by replacing MLP into MTA.
On ImageNet $256 \times 256$, PixelREPA-B$/16$ reduces FID~\cite{heusel2017gans} from 3.66 to 3.17 against JiT-B$/16$, and it achieves over $2\times$ faster convergence.
PixelREPA-H$/16$ further reaches FID 1.81, outperforming vanilla JiT-H$/16$ at 1.86 and even JiT-G$/16$ at 1.82, which has nearly $2\times$ more parameters.
These results show that PixelREPA improves both training efficiency and final generation quality at high resolution.

In summary, the core contributions of this study are as follows:
\begin{tcolorbox}[
  colback=gray!8,      
  colframe=myteal,
  boxrule=0.6pt,       
  arc=3mm,             
  left=2.5mm,right=2.5mm,top=2mm,bottom=2mm
]
\begin{itemize}[label=\raisebox{0.2ex}{\tiny$\bullet$}, leftmargin=*, itemsep=2pt, topsep=2pt]
  \item
  We find REPA degrades high resolution pixel-space diffusion training and induces \emph{feature hacking}, where direct regression to a compressed semantic target collapses diversity among samples with similar encoder features.

  \item
  We propose PixelREPA, which transforms the alignment target and constrains the alignment branch with a Masked Transformer Adapter that combines a shallow Transformer adapter with partial token masking.

  \item
  PixelREPA improves both convergence speed and generation quality on ImageNet $256 \times 256$, reducing FID from 3.66 to 3.17 for B$/16$ and reaching 1.81 for H/16.
\end{itemize}
\end{tcolorbox}

%% file: sections/02_preliminaries.tex
\begin{figure}[t]
  \centering
  \includegraphics[height=5.6cm]{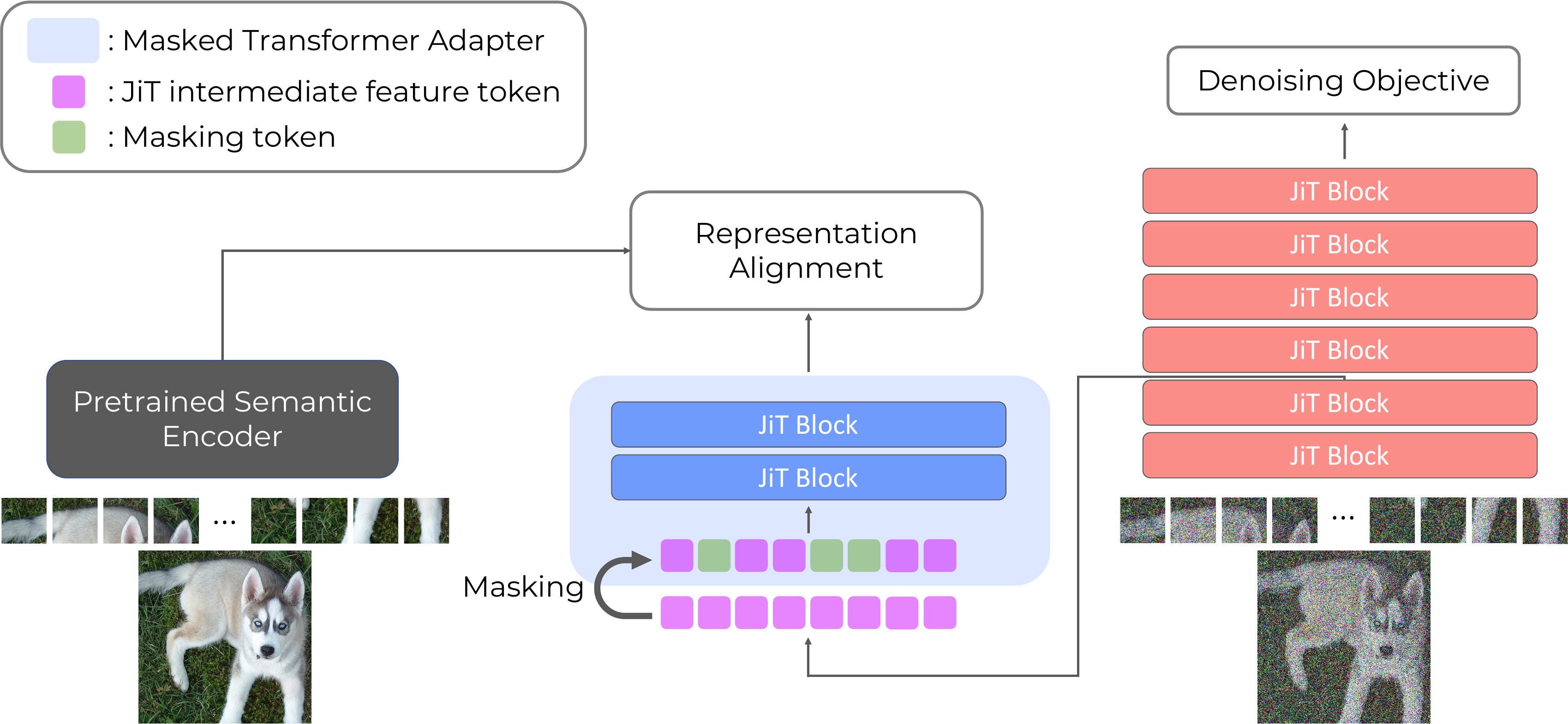}
  \vspace{-1em}
  \caption{\textbf{Overall Framework of PixelREPA.}
  PixelREPA masks a subset of tokens in an intermediate diffusion feature map.
  The full token sequence, with only a subset masked,  are then transformed by a shallow Transformer adapter and aligned to features from a frozen pretrained semantic encoder. 
  This transforms the alignment target and reduces overfitting to the external semantic representation.
  }
  \label{fig:framework}
  \vspace{-1em}
\end{figure}

\subsection{Diffusion and Flow-based Generative Models}

\subsubsection{DDPM.}
Diffusion models were popularized through Denoising Diffusion Probabilistic Models (DDPM)~\cite{ho2020denoising}, which consists of a forward noising process and a learned reverse denoising process. 
Given a data sample $\vx \sim \pdata$ and Gaussian noise $\vepsilon \sim \gN(\vzero, \mI)$ for timestep $t \in \{0, \dots, T\}$, the diffusion process is defined by two main trajectories, a forward process and a reverse process.
The forward process $q(\vx_t \mid \vx_{t-1})$ gradually corrupts the sample by adding noise according to a variance schedule $\evbeta_t$: $q(\vx_t \mid \vx_{t-1}) = \gN \!\left(\vx_t; \sqrt{1-\evbeta_t} \vx_{t-1}, \evbeta_t \mI\right).$
The reverse process $p_{\vtheta}(\vx_{t-1} \mid \vx_t)$ is trained to denoise the corrupted sample and recover the original data:
$p_{\vtheta}(\vx_{t-1} \mid \vx_t) = \gN \!\left(\vx_{t-1}; \frac{1}{\sqrt{\alpha_t}} \left(\vx_t - \frac{\evbeta_t}{\sqrt{1-\bar{\alpha}_t}} \vepsilon_{\vtheta}(\vx_t, t)\right), \evsigma_t^2 \mI\right),$
where $\alpha_t := 1 - \evbeta_t$ and $\bar{\alpha}_t := \prod_{s=1}^t \alpha_s$.
A neural network model is $\vepsilon_{\vtheta}(\cdot)$ and $\evsigma_t^2$ denotes a variance schedule.
Finally, the model is trained to predict the added noise by minimizing the following training objective:
\begin{align}
\Ls_{\text{DDPM}}
&= \E_{\vx,\vepsilon,t}\Big[\big\lVert \vepsilon - \vepsilon_{\vtheta}(\vx_t,t)\big\rVert_2^2\Big].
\end{align}

\subsubsection{Flow-based Geneartive Models.}
From a continuous-time perspective, diffusion models can also be formulated as an ODE-based flow~\cite{albergo2022building,lipman2022flow,liu2022flow}. 
In this perspective, a noisy sample $\vx_t=a_t\vx + b_t\vepsilon$ is an interpolation between data $\vx$ and noise $\vepsilon$, with pre-defined noise schedules $a_t$ and $b_t$ and timestep $t\in[0,1]$.
A flow velocity at timestep $t$ is defined as the time-derivative of $\vx_{t}$ as $\vv_{t} = \vx_{t}' = a_{t}' \vx + b_{t}' \vepsilon$.
Under linear schedules $a_t=t$ and $b_t=1-t$, the corresponding velocity can be represented as $\vv = \vx-\vepsilon$. 
Flow-based models learn a velocity field that deterministically transports samples from noise to clean data, via the following velocity-matching objective:
\begin{align}
\Ls_{\text{flow}}
&= \E_{\vx,\vepsilon,t}\Big[\big\lVert \vv_{\vtheta}(\vx_t,t) - \vv \big\rVert_2^2\Big].
\end{align}

\subsection{Pixel-space Diffusion}
Latent diffusion model (LDM)~\cite{rombach2022high} is the common choice for high-resolution generation, which denoises in a compressed autoencoder latent space.
LDM is efficient because operating in the lower-dimensional latent space reduces computation and memory, enabling faster training and sampling at high resolutions.
However, there is a reconstruction bottleneck: generated image quality is bounded by the autoencoder~\cite{blau2018perception}, and strong compression can remove fine textures and small structures in latent space~\cite{jiang2021focal}.
Also, since the autoencoder is trained for reconstruction rather than generation, this mismatch can surface as artifacts such as overly smooth textures or slight color shifts.
It further adds an extra component to train and maintain, and decoding latents back to pixels adds overhead at sampling time. 
These limitations make pixel-space diffusion attractive. 

Recent works have revisited diffusion directly in pixel space and shown that strong results are possible without an external autoencoder. 
SiD2~\cite{hoogeboom2025simpler} scales pixel-space diffusion model with sigmoid loss weighting and a streamlined U-ViT backbone. 
More recently, JiT~\cite{li2025back} achieves performance comparable to latent-space diffusion by employing a pure Transformer architecture. 
JiT shows that clean image prediction ($\vx$-prediction) is necessary, regardless of prediction type. Formally, JiT uses $\vx$-prediction~\cite{salimans2022progressive} and velocity-matching objective:
\begin{align}
\gL_{\text{JiT}} = \mathbb{E}_{\vx,\vepsilon,t}\Big[\big\lVert \tilde{\vv}_{\vtheta}(\vx_t,t) - \vv \big\rVert_2^2\Big],
\end{align}
where $\tilde{\vv}_{\vtheta}(\vx_t, t) = (\rvx_{\vtheta}(\vx_t, t) - \vx_t) / (1-t)$ and $\rvx_{\vtheta}(\cdot)$ is the $\vx$-prediction network.

\subsection{Representation Alignment for Generation}
Recently, REPA~\cite{yu2024representation} has emerged an effective approach for accelerating training and improving sample quality in DiT~\cite{peebles2023scalable} and SiT~\cite{ma2024sit}.
REPA aligns intermediate diffusion features with semantic representations from a frozen pretrained encoder $f(\cdot)$.
The alignment objective is simply defined as:
\begin{align}
\gL_{\text{REPA}} = -\mathbb{E}_{\vx,\vepsilon,t}\Big[\frac{1}{N}\sum_{n=1}^N\text{cossim}(f(\vx)^{[n]}, h_\phi(\vh_t^{[n]}))\Big],
\end{align}
where $n$ is a patch index, $N$ is the number of patches, $\vh_t$ denotes an intermediate feature of diffusion Transformers at timestep $t$, $h_\phi(\cdot)$ indicates a projection function, and $\text{cossim}(\cdot, \cdot)$ represents a cosine-similarity function.

Given its simplicity and effectiveness, several subsequent studies have been conducted.
For instance, REPA-E~\cite{leng2025repa} utilizes this alignment for the end-to-end joint tuning of a VAE and a diffusion model, and Wang \etal~\cite{wang2025repa} introduce an early termination strategy, coupled with attention alignment.
Furthermore, this approach has been successfully extended to various tasks, including video generation~\cite{zhang2025videorepa, lee2025improving}, 3D-aware generation~\cite{wu2025geometry,kim2024dreamcatalyst}, and unified model training~\cite{ma2025janusflow}.

%% file: sections/03_motivations.tex
We begin with our main findings and show experimental analysis to verify these findings.
\Cref{fig:main} shows na\"ively applying REPA~\cite{yu2024representation} to JiT~\cite{li2025back}, a pixel-space diffusion model, leads to a performance degradation.
REPA is a simple regularization strategy that has been shown to accelerate training convergence and improve final performance in latent space diffusion transformers such as DiT~\cite{peebles2023scalable} and SiT~\cite{ma2024sit}.
These advantages provide a clear motivation to apply REPA to JiT.
However, JiT$+$REPA underperforms vanilla JiT on ImageNet~\cite{deng2009imagenet} $256 \times 256$ as training progresses.
This gap raises a natural question: \emph{why does REPA facilitate learning in latent space diffusion, yet struggle in pixel-space diffusion?}

Before diving into this question, we first revisit the key differences between latent space and pixel space.
The key differences between latent space and pixel space fall into two aspects~\cite{esser2021taming,rombach2022high}: (1) \emph{dimensionality of representation} and (2) \emph{perceptual compression}.

We first focus on \emph{dimensionality}.
Latent diffusion~\cite{rombach2022high} performs denoising in a compact token grid whose spatial size and channel capacity are reduced relative to the image, which substantially lowers the degrees of freedom that the denoiser must model. 
Pixel-space diffusion instead denoises the target in the ambient image space. 
For an image of resolution $H \times W$, this space contains $O(H \times W)$ degrees of freedom. 
As $H$ and $W$ increase, the number of local variations grows rapidly, and many of these variations correspond to fine scale intensity changes rather than semantic changes. 
This high dimensional continuous geometry makes a mapping from semantic features to fine detailed images as highly ill-posed.

Second, latent representations~\cite{rombach2022high} introduce an explicit \emph{perceptual compression}.
The pretrained tokenizer maps an image into a compact code representation that prioritizes salient, reconstructable content~\cite{rombach2022high}. 
As a result, much of the fine grained detail and high frequency variation is attenuated in the latent~\cite{jiang2021focal}. 
Pixel space retains these details in the denoising signal, including textures and micro patterns that are weakly tied to semantics. 
This discrepancy leads to different learning dynamics between latent and pixel-space diffusion.

\begin{figure}[t]
    \centering
        \begin{subfigure}[b]{0.48\textwidth}
            \centering
            \includegraphics[width=\textwidth]{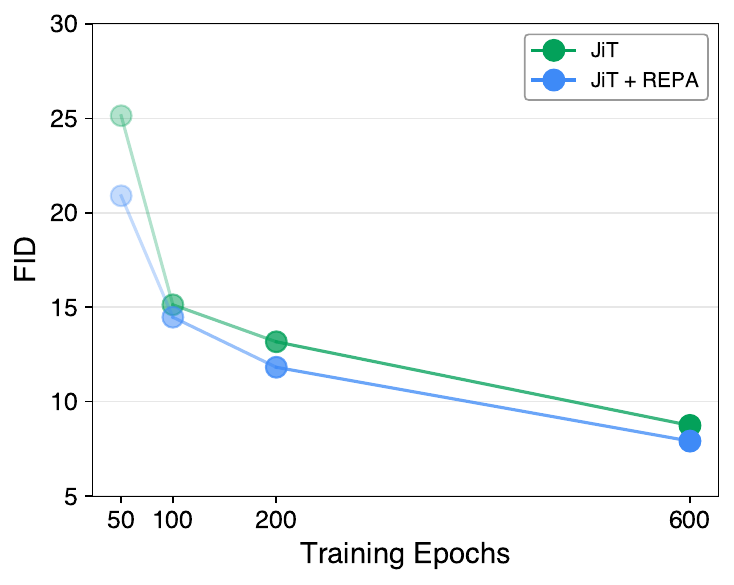}
            \caption{\textbf{ImageNet $32 \times 32$}}
            \label{fig:32x32}
        \end{subfigure}
        \hfill
        \begin{subfigure}[b]{0.48\textwidth}
            \centering
            \includegraphics[width=\textwidth]{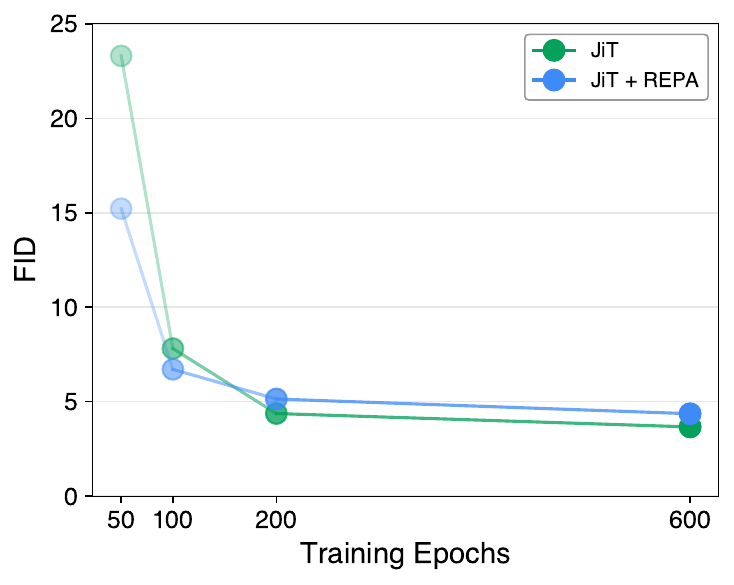}
            \caption{\textbf{ImageNet $256 \times 256$}}
            \label{fig:256x256}
        \end{subfigure}
        \caption{
        \textbf{REPA accelerates pixel diffusion training at low resolution, whereas it degrades training at high resolution.}
        This figure illustrates FID scores across different resolutions comparing JiT and JiT$+$REPA. 
        Results show (a) ImageNet $32 \times 32$ and (b) ImageNet $256 \times 256$ with varying training epochs. 
        }
        \label{fig:resolution_comparison}
        \vspace{-1em}
\end{figure}

\subsection{Dimensionality of Representation}
We now return to the main question and analyze it through the lens of these two differences.
We first investigate whether the performance degradation stems from \emph{dimensionality of representation}. 
\Cref{fig:resolution_comparison} compares JiT and JiT$+$REPA on ImageNet $32 \times 32$ and $256 \times 256$.
This setup is designed to identify the effect of \emph{dimensionality} on JiT$+$REPA by varying resolution.
\Cref{fig:32x32} shows REPA improves over vanilla JiT at low resolution.
In contrast, \cref{fig:256x256} shows REPA degrades performance as training progresses at high resolution. 
These results suggest that REPA becomes ineffective as the degrees of freedom increase, while remaining beneficial in low dimensional settings.
As a result, this experiment presents a following finding:

\begin{tcolorbox}[
  colback=gray!8,      
  colframe=myteal,
  boxrule=0.6pt,       
  arc=3mm,             
  left=2.5mm,right=2.5mm,top=2mm,bottom=2mm
]
    \textbf{Finding 1.}
    The failure of REPA in pixel space emerges as dimensionality of representation space increases.
\end{tcolorbox}

\begin{figure}[t]
    \centering
    \begin{minipage}[h]{0.48\textwidth}
        \centering
        \includegraphics[width=\textwidth]{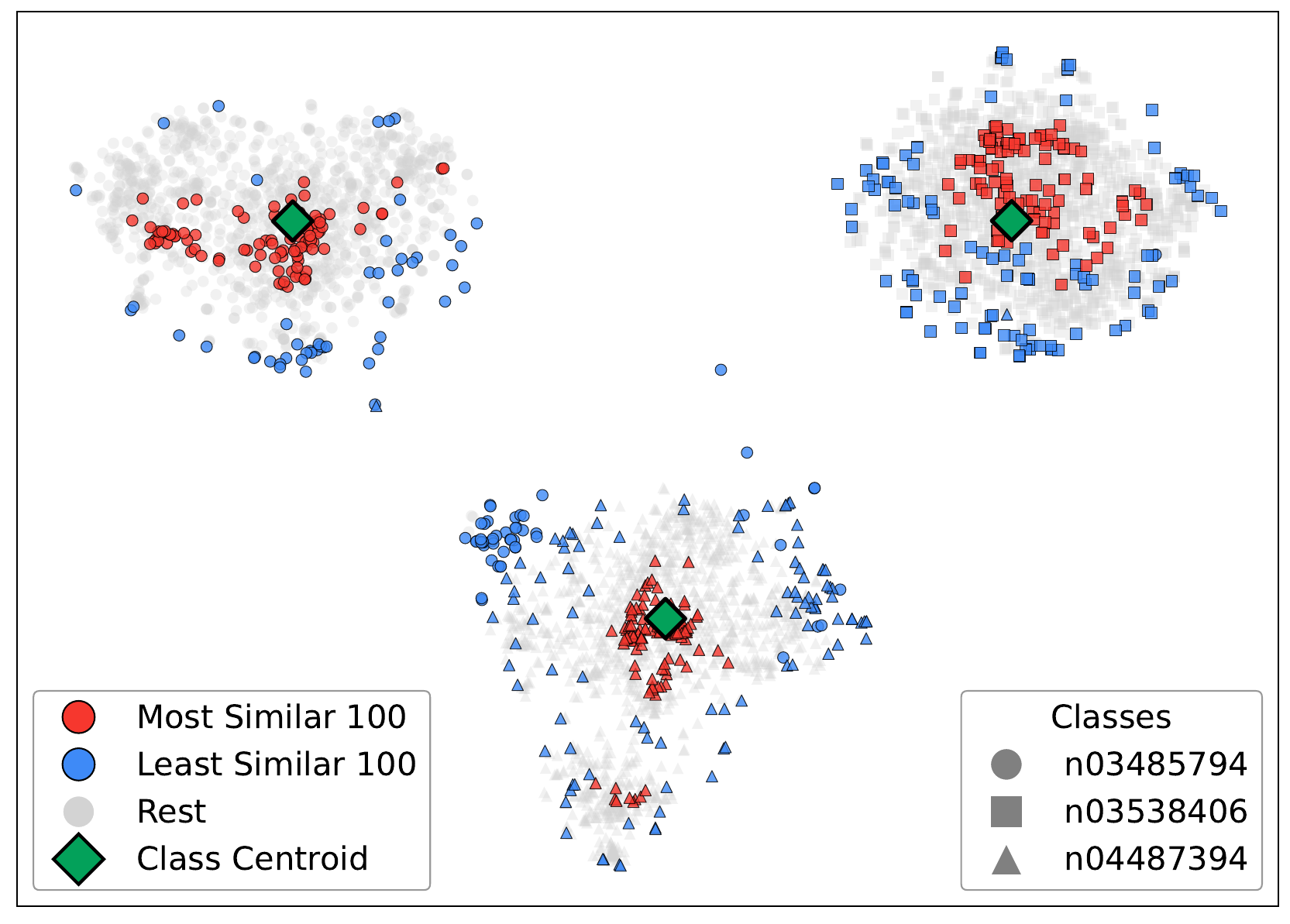}
        
        \caption{\textbf{Visualization of semantic representation distribution by class with t-SNE~\cite{van2008visualizing}}.
        For each of classes, we compute a centroid in the semantic feature space based on feature similarity. 
        We mark the 100 samples most similar to the centroid as red dots and the 100 samples least similar to the centroid as blue dots.
        }
        \label{fig:tsne}
    \end{minipage}
    \hfill
    \begin{minipage}[h]{0.48\textwidth}
    \begin{subfigure}[h]{0.48\textwidth}
        \centering
        \includegraphics[width=\textwidth]{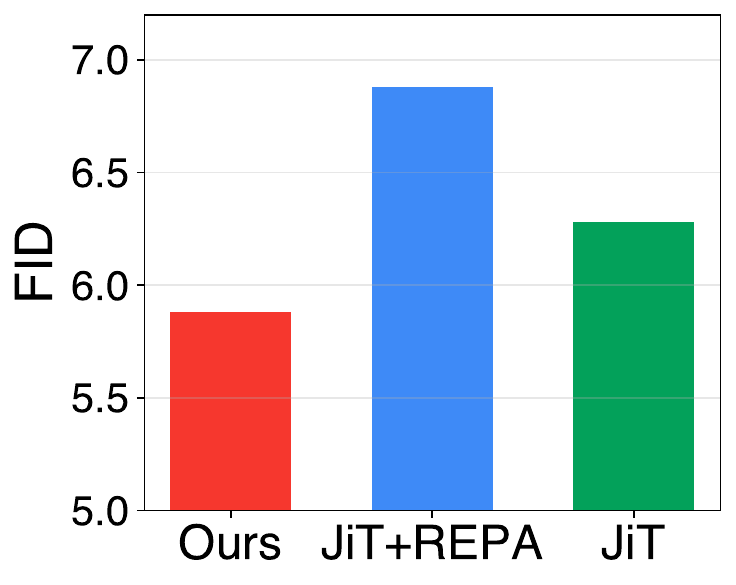}
        \caption{Most Similar 100}
        \label{fig:most_similar}
    \end{subfigure}
    \begin{subfigure}[h]{0.48\textwidth}
        \centering
        \includegraphics[width=\textwidth]{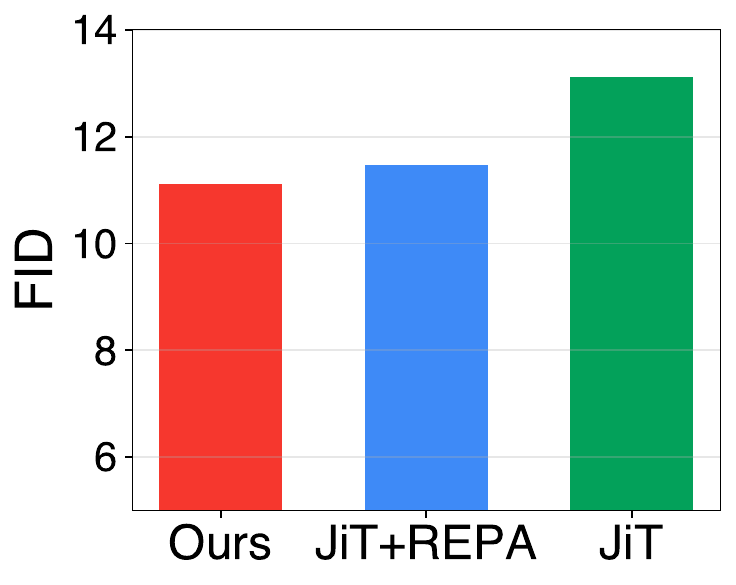}
        \caption{Least Similar 100}
        \label{fig:least_similar}
    \end{subfigure}
    \caption{
    \textbf{REPA degrades generation diversity compared to the vanilla JiT on the most similar 100 samples for each class}.
    This figure shows FID scores across different training data selection strategies.
    We compute FID across randomly selected 100 classes using 100 samples per class.
    Vanilla JiT achieves lower FID on the Most Similar 100 subset, whereas JiT$+$REPA achieves lower FID on the Least Similar 100 subset.
    Ours shows the best FID on both settings.
    }
    \label{fig:training_comparison}
    \end{minipage}
    \vspace{-1em}
\end{figure}

\subsection{Perceptual Compression}
We next investigate \emph{perceptual compression}. 
The latent space induced by a pretrained image tokenizer suppresses fine grained detail and high frequency variation compared to pixel space.
This \emph{perceptual compression} makes the latent denoising space more compatible with the representation space of a pretrained semantic encoder.
As a result, REPA leads to faster convergence and improved performance when aligning the semantic representation to LDMs.
In contrast, the alignment degrades performance in high resolution pixel-space diffusion.

We hypothesize this degradation arises because many semantically similar yet visually distinct images map to similar regions in the feature space of the pretrained encoder as high resolution pixel space has substantially more degrees of freedom.
To verify this, we compare vanilla JiT and JiT$+$REPA across samples that are close to, or far from, a mode in the external semantic feature space. 
For each ImageNet class, we compute a class centroid in the feature space of the external semantic encoder $f(\cdot)$ as \cref{fig:tsne}.
This centroid serves as a proxy for a dense semantic mode, where many semantically similar images concentrate in the encoder representation space.
We then extract two subsets: the most similar 100 samples to the centroid and the least similar 100 samples from the centroid.
The most similar subset contains images that differ in pixel space, yet remain tightly clustered in $f(\cdot)$ space.
Conversely, the least similar subset is widely scattered in $f(\cdot)$ space, indicating low similarity under the external semantic representation.

\begin{figure}[t]
    \centering
    \setlength{\tabcolsep}{0.1em} 
    \subcaptionbox{\textbf{Most Similar 100}\label{fig:most_similar_100}}{%
        \begin{tabular}{@{}cccccccc@{}}
            \includegraphics[width=0.12\textwidth]{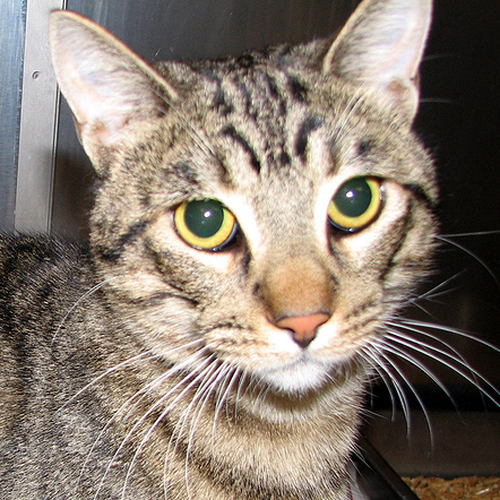} &
            \includegraphics[width=0.12\textwidth]{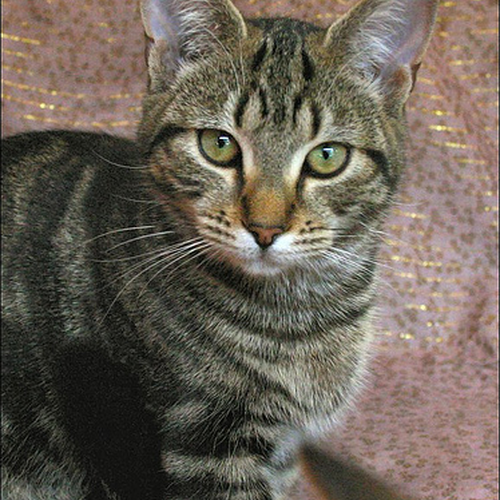} &
            \includegraphics[width=0.12\textwidth]{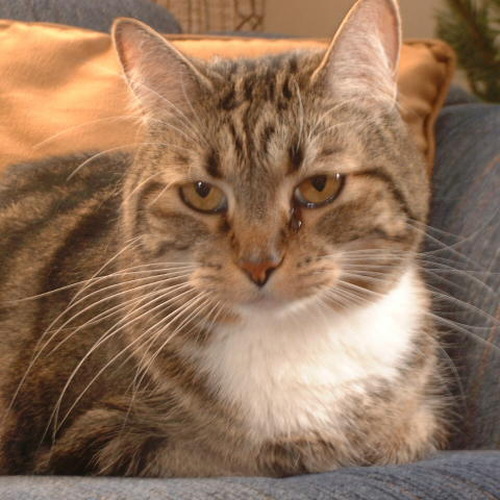} &
            \includegraphics[width=0.12\textwidth]{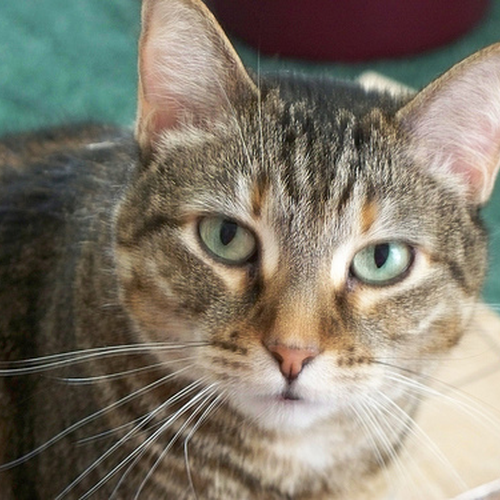} &
            \includegraphics[width=0.12\textwidth]{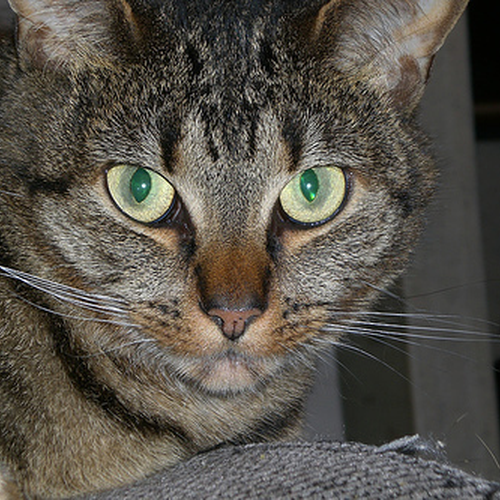} &
            \includegraphics[width=0.12\textwidth]{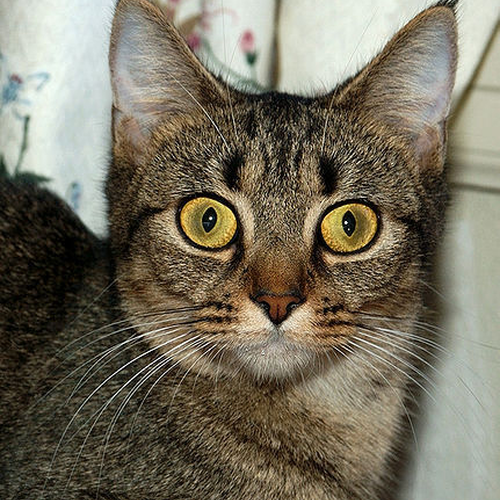} &
            \includegraphics[width=0.12\textwidth]{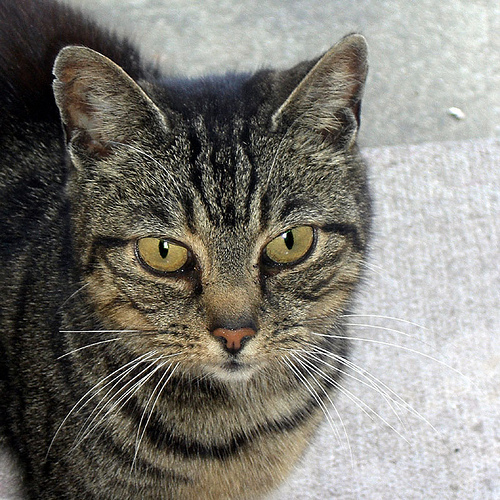} &
            \includegraphics[width=0.12\textwidth]{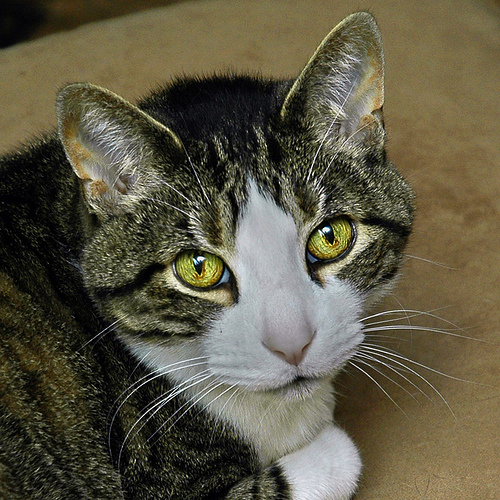}
        \end{tabular}
    } 
    
    \subcaptionbox{\textbf{Least Similar 100}\label{fig:least_similar_100}}{%
        \begin{tabular}{@{}cccccccc@{}}
            \includegraphics[width=0.12\textwidth]{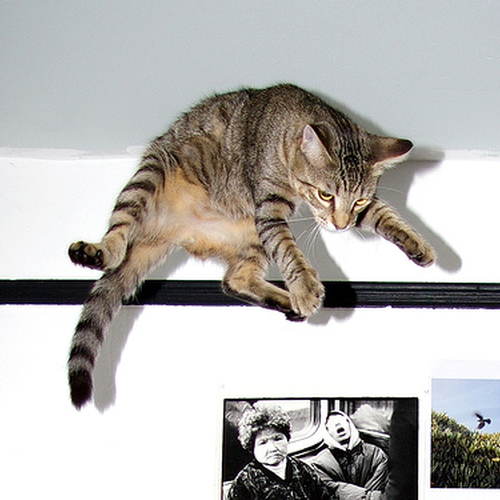} &
            \includegraphics[width=0.12\textwidth]{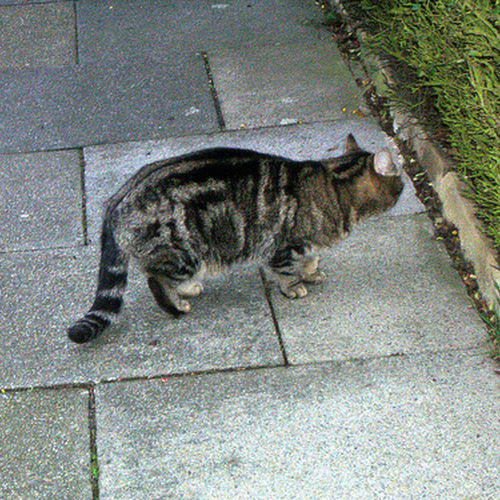} &
            \includegraphics[width=0.12\textwidth]{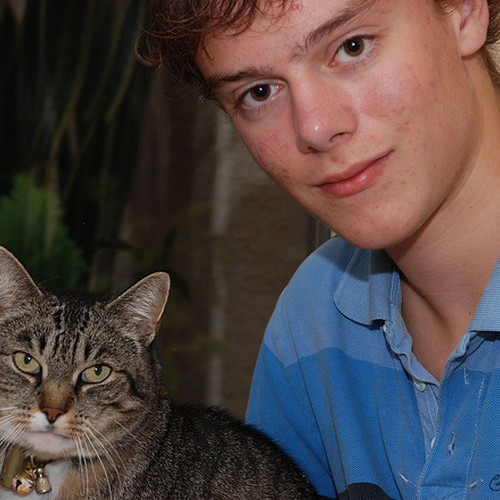} &
            \includegraphics[width=0.12\textwidth]{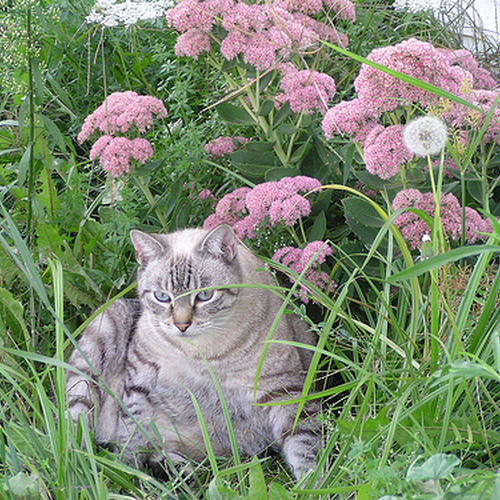} &
            \includegraphics[width=0.12\textwidth]{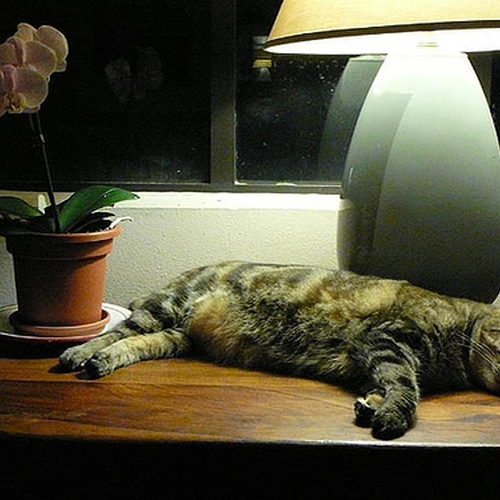} &
            \includegraphics[width=0.12\textwidth]{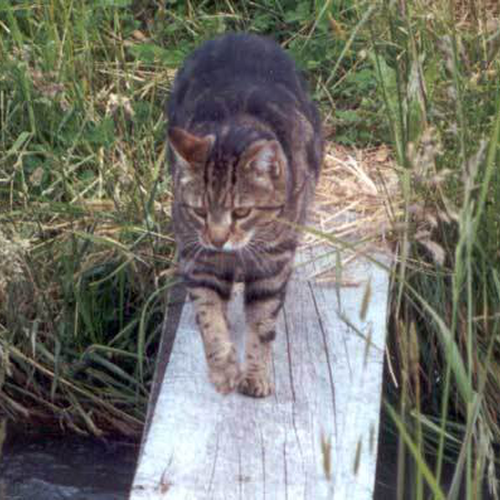} &
            \includegraphics[width=0.12\textwidth]{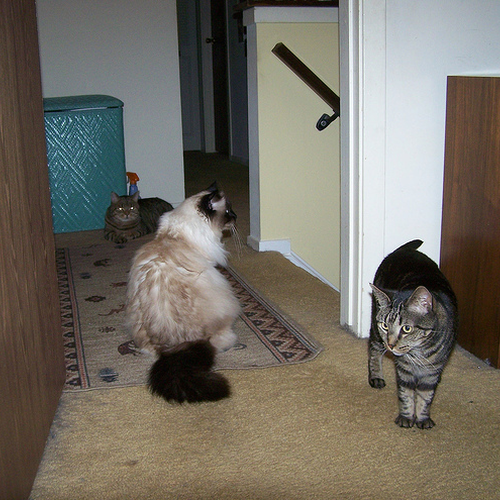} &
            \includegraphics[width=0.12\textwidth]{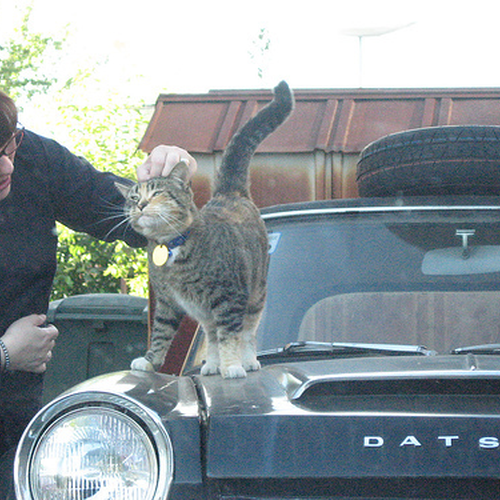}
        \end{tabular}
    }

    \caption{\textbf{Most Similar 100 and Least Similar 100 images in the external semantic feature space.} (a) Most Similar 100 images to the class centroid in feature space of the external semantic encoder. (b) Least Similar 100 images to the centroid.}
    \label{fig:viz_sim}
    \vspace{-1em}
\end{figure}

\Cref{fig:viz_sim} visualizes the two subsets defined in the external semantic feature space. 
As shown in \cref{fig:viz_sim}, the most similar 100 samples share similar global structure and composition, while differing mainly in fine scale details. 
In contrast, the least similar 100 samples differ substantially from both structure and content. 
These visualizations suggest that the most similar 100 images cluster tightly and map to highly similar semantic features under the encoder, whereas the least similar 100 images are scattered at semantic space and map to distinct semantic features.
These images are perturbed with diffusion noise at $t=0.2$ to retain part of the original image signal.
Each model then denoises these noisy images.

As shown in \cref{fig:training_comparison}, vanilla JiT achieves lower FID than JiT$+$REPA on the most similar 100 subset, while opposite holds on the least similar 100 subset.
This asymmetry is the signature of feature hacking. 
The most similar 100 subset is precisely where \emph{feature hacking} manifests: images are pixel-diverse yet semantically clustered, so the alignment loss drives them toward a narrow region of the feature space near the mode.
On the least similar subset, where semantic targets are well-separated, alignment is informative and REPA benefits. 
This confirms that failure of REPA in pixel space is not a uniform degradation but a structured one: it harms generation quality specifically where the feature space is most ambiguous.
We refer to this as \emph{feature hacking}.   
Our second finding is:
\begin{tcolorbox}[
  colback=gray!8,      
  colframe=myteal,
  boxrule=0.6pt,       
  arc=3mm,             
  left=2.5mm,right=2.5mm,top=2mm,bottom=2mm
]
    \textbf{Finding 2.}
    In high resolution pixel-space diffusion, REPA induces \textit{feature hacking} and hinders training on sets of images whose semantic features are highly similar under the external semantic encoder.
\end{tcolorbox}

%% file: sections/04_method.tex
Our analysis identifies two causes behind REPA~\cite{yu2024representation} failure in pixel-space diffusion: (1) the \emph{dimensionality of representation}, and (2) the \emph{perceptual compression}.
Both stem from alignment target of REPA. 
REPA projects the intermediate features of JiT~\cite{li2025back} through a point-wise MLP into the representation space of pretrained semantic encoder and aligns it there.
This pulls a compressed semantic representation toward the diffusion space. 
In latent diffusion, where the diffusion feature is already compact, this gap is manageable. 
In pixel space, the JiT features carry far richer information than $f(\cdot)$. 
Then, the MLP enforces the JiT features to conform to the compressed $f(\cdot)$, without learning the fine-grained structure needed for high-quality pixel generation. 
Later diffusion blocks must then reconstruct diverse pixel outputs from a compressed semantic code—an ill-posed mapping since many distinct images share similar $f(\cdot)$.

We address this by transforming the alignment target and constraining the alignment pathway.
Rather than the JiT representations targets to learn $f(\cdot)$, we transform the semantic target through a dedicated module and \emph{align intermediate features into the transformed space induced by the module}.
This module, namely the Masked Transformer Adapter (MTA), consists of two components: (1) a \emph{shallow transformer adapter} and (2) a \emph{partial masking strategy}.
The shallow transformer adapter performs contextual aggregation over diffusion tokens, so each token prediction can leverage information from other tokens.
The partial masking applies random token masking to the adapter input, which regularizes alignment to discouraging overly direct regression to $f(\cdot)$ and acts as an information bottleneck. 
\Cref{fig:framework} describes the overall framework of PixelREPA.

\subsection{Shallow Transformer Adapter}
We introduce a shallow Transformer adapter for transforming alignment target. 
The adapter transforms an intermediate diffusion feature from the JiT encoder into the external semantic feature space for matching $f(\cdot)$.
This adapter consists of two Transformer blocks with self-attention. 
The critical difference from the original MLP-based alignment is twofold.

First, the adapter \emph{selectively} learns the transformation from the JiT features into $f(\cdot)$ to transform alignment target.
This means the alignment objective no longer pressures the JiT intermediate representation to match a compressed target. 
Instead, the adapter learns to extract semantic content from the JiT representation and project it into the space of $f(\cdot)$.
As a result, the alignment target of the JiT features is \emph{inversely} transformed feature by the adapter from the space of $f(\cdot)$.
The JiT features remain free to encode the full range of pixel-level detail needed for high-quality generation, while the adapter selectively distills the semantic signal for alignment.

Second, the adapter performs contextual aggregation via self-attention. 
Each token prediction incorporates information from neighboring tokens before matching $f(\cdot)$, rather than being mapped in isolation. 
This forces the adapter to build semantic predictions from a broader spatial context, producing a more structured transformation than a point-wise mapping. Contextual aggregation also reduces reliance on per-token correspondence, weakening the trivial regression pathway that underlies feature hacking.

The adapter remains lightweight by using only two transformer blocks, while providing a more structured and stable alignment pathway than an MLP head in high resolution pixel space diffusion.

In the perspective of self-supervised learning~\cite{bengio2013representation}, JiT can be composited of two functions as $\rvx_\theta = g_\theta \circ f_\theta$, where the encoder $f_\theta : \gX \to \gH$ and the decoder $g_\theta : \gH \to \gX$.
The encoder $f_\theta(\vx_t)$ maps a noisy image $\vx_t$ to the intermediate representation $\vh_t$, as $f_\theta(\vx_t) = \vh_t \in \gH$.
Image space, hidden space, and semantic feature space are denoted as $\gX, \gH,$ and $\gR$, respectively.
Finally, the transformer adapter $d_\phi : \gH \to \gR$ transforms $\vh_t$ to predict the external semantic feature.

\subsection{Partial Masking Strategy}
The shallow Transformer adapter transforms the alignment target and introduces contextual aggregation, but is not sufficient on its own. 
Without additional constraints, the adapter can still learn a near-trivial mapping from the JiT representation to $f(\cdot)$ and our experiments confirm this: an unmasked adapter (mask ratio 0.0) achieves FID 4.68 at 200 epochs, which improves over JiT$+$REPA (5.14) but falls behind vanilla JiT (4.37), as demonstrated in \cref{table:jit_repa_ours}. 
This intermediate result reveals that transforming the alignment target reduces but does not eliminate the shortcut, because the adapter can still exploit per-token correspondence between $\vh_t$ and $f(\vx)$ when all tokens are visible.

We propose a partial masking strategy on the adapter input to address this residual shortcut.  
During training, we randomly mask a fraction $r$ of tokens in the intermediate JiT feature map before passing the shallow transformer adapter. 
The adapter must then predict the full semantic target from partial observations, using the neighboring tokens as context. 

Masking serves two complementary roles. 
(1) Shortcut prevention: By removing a subset of input tokens, masking breaks the per-token correspondence between the JiT presentations and semantic features. 
This requires genuine contextual reasoning and prevents the trivial regression pathway. 
(2) Information bottleneck~\cite{tishby2000information} on the pixel side: Masking reduces the effective degrees of freedom of the adapter input from $O(N \cdot d)$ to $O((1-r)\cdot N\cdot d)$, where $N$ is the number of tokens, $d$ is the hidden dimension, and $r$ is the mask ratio. 
This narrows the information gap between the pixel representation and the compressed target, analogous to the dimensionality reduction that tokenizer performs in latent diffusion, but applied selectively to the alignment pathway rather than to the denoising process itself. 
The main denoising pathway retains the full, unmasked token sequence. 

Finally, PixelREPA is defined as follows:
\begin{align}
\gL_{\text{PixelREPA}} := -\mathbb{E}_{\vx, \boldsymbol{\epsilon}, t} \left[ \frac{1}{N} \sum_{n=1}^N \text{cossim}(f(\vx)^{[n]}, d_\phi(m \odot \vh_t^{[n]}) \right],
\end{align}
where $m$ denotes a patch-wise mask.
The final objective function becomes:
\begin{align}
\gL := \gL_{\text{JiT}} + \lambda\gL_{\text{PixelREPA}},
\end{align}
where $\lambda>0$ is a regularization hyperparameter.

%% file: sections/05_experiments.tex
\begin{table}[t]
\centering
\small
\setlength{\tabcolsep}{6pt}
\renewcommand{\arraystretch}{1.15}

\begingroup
\setlength{\arrayrulewidth}{0.2pt} 
\caption{\textbf{Quantitative comparison of diffusion models on ImageNet 256$\times$256.} FID and IS are evaluated with 50K samples.}
\begin{tabular}{l | c | c c}
\specialrule{0.9pt}{0pt}{0pt}
\textbf{Model} & \textbf{params} & \textbf{FID$\downarrow$} & \textbf{IS$\uparrow$} \\
\specialrule{0.6pt}{0pt}{0pt}

\rowcolor{sectiongray}
\multicolumn{4}{l}{\textit{Latent-space Diffusion}} \\
DiT-XL$/2$~\cite{pernias2023wurstchen}                          & 675+49M   & 2.27 & 278.2 \\
SiT-XL$/2$~\cite{ma2024sit}                                     & 675+49M   & 2.06 & 277.5 \\
REPA~\cite{yu2024representation}, SiT-XL$/2$                    & 675+49M   & 1.42 & 305.7 \\
LightningDiT-XL$/$2~\cite{yao2025reconstruction}                & 675+49M   & 1.35 & 295.3 \\
DDT-XL$/2$~\cite{wang2025ddt}                                   & 675+49M   & 1.26 & 310.6 \\
RAE~\cite{zheng2025diffusion}, DiT$^{\text{DH}}$-XL$/2$         & 839+415M  & 1.13 & 262.6 \\

\specialrule{0.6pt}{0pt}{0pt}
\rowcolor{sectiongray}
\multicolumn{4}{l}{\textit{Pixel-space Diffusion}} \\
ADM-G~\cite{dhariwal2021diffusion}                            & 559M      & 7.72 & 172.7 \\
RIN~\cite{jabri2022scalable}                                  & 320M      & 3.95 & 216.0 \\
SiD~\cite{hoogeboom2023simple}                                & 2B        & 2.44 & 256.3 \\
VDM++~\cite{kingma2023understanding}, UViT$/2$                                                 & 2B        & 2.12 & 267.7 \\
SiD2~\cite{hoogeboom2025simpler}, UViT$/2$                      & -         & 1.73 & -     \\
SiD2~\cite{hoogeboom2025simpler}, UViT$/1$                      & -         & 1.38 & -     \\
PixelFlow~\cite{chen2025pixelflow}, XL$/4$                      & 677M      & 1.98 & 282.1 \\
PixNerd~\cite{wang2025pixnerd}, XL$/16$                         & 700M      & 2.15 & 297.0 \\

\arrayrulecolor{gray!60}
\specialrule{0.3pt}{0pt}{0pt} 
\rowcolor{gray!3}
JiT-B$/16$~\cite{li2025back}                                    & 131M      & 3.66 & 275.1 \\
\rowcolor{gray!3}
JiT-L$/16$~\cite{li2025back}                                    & 459M      & 2.36 & 298.5 \\
\rowcolor{gray!3}
JiT-H$/16$~\cite{li2025back}                                    & 953M      & 1.86 & 303.4 \\
\rowcolor{gray!3}
JiT-G$/16$~\cite{li2025back}                                    & 2B        & 1.82 & 292.6 \\

\specialrule{0.3pt}{0pt}{0pt}
\rowcolor{gray!10}
PixelREPA-B$/16$                                                & 131M      & 3.17 & 284.6 \\
\rowcolor{gray!10}
PixelREPA-L$/16$                                                & 459M      & 2.11  & 309.5 \\
\rowcolor{gray!10}
PixelREPA-H$/16$                                                & 953M      & 1.81 & 317.2 \\

\arrayrulecolor{black}
\specialrule{0.9pt}{0pt}{0pt}
\end{tabular}
\label{table:models_comparison}
\endgroup
\vspace{-1em}
\end{table}

\subsection{Setup}
\subsubsection{Implementation details.}
Our implementation and configuration strictly follow the implementation of JiT~\cite{li2025back}.
Specifically, each model configuration follows JiT paper across all sizes, except for MTA.
We use a 2-layer transformer adapter, with masking ratio $r = 0.2$, regardless of the model size. 
For external semantic encoder, we employ DINOv2~\cite{oquab2023dinov2} as REPA ~\cite{yu2024representation}.
The intermediate features input to the MTA were obtained from the layer directly prior to the in-context start block, specific to each model size.
We fix a regularization hyperparameter $\lambda=0.1$ for every model size.
Our experiments are conducted on 8 NVIDIA H200 GPUs.
The detailed configurations are described in the supplementary materials.

\subsubsection{Evaluation.}
We evaluate FID~\cite{heusel2017gans} and Inception Score (IS)~\cite{salimans2016improved} with 50K samples, which is a common setting. 
Following JiT, we use a 50-step Heun~\cite{heun1900neue} ODE solver with CFG~\cite{ho2022classifier} interval $[0.1, 1]$~\cite{kynkaanniemi2024applying}.

\begin{table}[t]
    \centering
    \small
    \begin{minipage}[t]{0.4\columnwidth}
        \centering
        \setlength{\tabcolsep}{6pt}
        \renewcommand{\arraystretch}{1.15}
        \caption{
        \textbf{Ablation study for masking ratio.}
        This comparison is evaluated on ImageNet $256\times 256$ by FID with same model size B$/16$.
        Red colored box is the best result.
        }
        \resizebox{\linewidth}{!}{%
        \begin{tabular}{l | ccccc}
        \toprule
        \textbf{Mask ratio} & \textbf{0.1} & \textbf{0.2} & \textbf{0.3} & \textbf{0.4} & \textbf{0.5} \\
        \midrule
        \textbf{200-ep} & 4.26 & \cellcolor{mypink} 4.00 & 4.38 & 4.32 & 4.58 \\
        \bottomrule
        \end{tabular}
        }
        \label{table:mask_ratio}
    \end{minipage}
    \hfill
    \begin{minipage}[t]{0.56\columnwidth}
        \centering
        \small
        \setlength{\tabcolsep}{8pt}
        \renewcommand{\arraystretch}{1.15}
        \caption{
        \textbf{Ablation study for masking.}
        All models are trained on ImageNet $256\times256$, evaluated by FID. 
        The size of all models are fixed on B$/16$. 
        $\text{PixelREPA}^\dagger$ is PixelREPA without masking.
        Red colored box indicates the best result.}
        \resizebox{\linewidth}{!}{%
        \begin{tabular}{l | c c c c}
        \toprule
         \textbf{Model} & \textbf{JiT} & \textbf{JiT$+$REPA} & \textbf{$\text{PixelREPA}^\dagger$} & \textbf{PixelREPA} \\
        \midrule
        \textbf{200-ep} & 4.37 & 5.14 & 4.68 & \cellcolor{mypink} 4.00 \\
        \bottomrule
        \end{tabular}
        }
        \label{table:jit_repa_ours}
    \end{minipage}
\end{table}

\begin{figure}[t]
  \centering
  \includegraphics[width=0.7\textwidth]{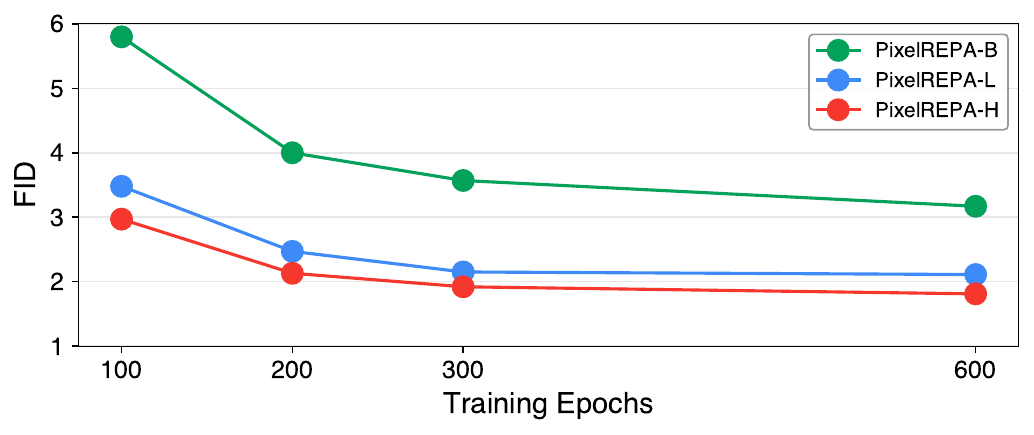}
  \caption{
  \textbf{Scalability.}
  As model size grows, PixelREPA shows better performance.
  }
  \label{fig:scalability}
\end{figure}

\subsection{Analysis}
\subsubsection{Comparisons on ImageNet 256$\times$256.}
As shown in \cref{table:models_comparison}, PixelREPA consistently outperforms JiT across all model scales.
For the B$/16$ architecture, PixelREPA reduces the FID from 3.66 to 3.17 (13.4$\%$) improvement. 
This trend holds for larger models: the L$/16$ variant yields 10.6$\%$ relative gain, and the H$/16$ variant shows further 2.7$\%$ improvement.
These consistent improvements confirm that PixelREPA is robust to the scalability.
Notably, PixelREPA-H$/16$ surpasses JiT-G$/16$, nearly $2\times$ larger model, demonstrating more effective parameter utilization.
Furthermore, PixelREPA achieves competitive results against recent pixel-space diffusion models without any modifications of transformer architecture, showcasing its robustness in pixel-level generation.

\subsubsection{Effectiveness of partial masking.}
First, we verify the effectiveness of partial marking by comparing whether mask is used.
MTA without masking surpasses JiT$+$REPA but falls behind the JiT baseline as \cref{table:jit_repa_ours}.
This shows our transformer adapter improves the generation performance than the standard REPA but suffers from accelerating JiT training.
Therefore, partial masking is essential to mitigate feature hacking.
This constraint discourages shortcut learning and reduces overfitting to the external semantic feature.

To investigate the effect of partial masking ratio, we compare PixelREPA by varying mask ratios.
As shown in \cref{table:mask_ratio}, the best performance is achieved at mask ratio $r=0.2$. 
However, further increasing mask ratio $r$ to 0.5 leads to performance degradation.
Masking removes supervision on the masked subset and then blocks the gradient signal to the JiT.
As a result, large mask ratio hinders JiT blocks to learn semantic features due to excessive information bottleneck, leading to the degraded performance.
Based on this analysis, we use a mask ratio of $0.2$ for all PixelREPA models.

\subsubsection{REPA vs. PixelREPA on JiT.}
PixelREPA effectively overcomes the inferior performance of REPA on JiT.
As illustrated in \cref{fig:256x256,table:jit_repa_ours}, JiT$+$REPA improves upon JiT early in training. 
However, this trend reverses with prolonged training, ultimately resulting in a 17.6\% degradation in FID (5.14 vs. 4.37) compared to the baseline at 200 epochs.
In contrast, PixelREPA consistently outperforms JiT, achieving an 8.5\% improvement over vanilla JiT at 200 epoch.
This suggests that PixelREPA stabilizes the alignment for pixel-space diffusion by reducing overfitting in the limited target feature space, making it a suitable alternative to the standard REPA in the high resolution image setting.

\Cref{fig:training_comparison} shows PixelREPA achieves the best FID scores at both the most and the least similar 100 samples.
However, JiT$+$REPA struggles on the most similar 100 setting as discussed at \cref{sec:motivation}.
This verifies PixelREPA is robust to synthesize images at the near centroids and thus mitigates feature hacking.

\subsubsection{Scalability.}
We investigate the scalability of PixelREPA by varying model size. 
PixelREPA achieves lower FID as the model scales up, as \cref{table:models_comparison,fig:scalability}. 
The improvement is consistent across training epochs at each model size, indicating better sample quality and diversity. 
PixelREPA also consistently outperforms vanilla JiT at matched model sizes.

%% file: sections/06_conclusion.tex
We revisit representation alignment for JiT and identify a failure mode of standard REPA at high resolution, where alignment to a compressed semantic target leads to feature hacking and degraded training. 
PixelREPA addresses this issue by transforming the alignment target and constraining the alignment pathway with a shallow Transformer adapter and partial token masking. The resulting Masked Transformer Adapter stabilizes optimization, scales with model size, and improves ImageNet $256 \times 256$ results across JiT backbones. 
PixelREPA reduces FID from 3.66 to 3.17 for B/16 and achieves 1.81 for H/16.

%% file: sections/07_supple.tex
\appendix
\section{Implementation Details}
\label{sec:appen_imple}

\begin{table}[h]
  \centering
  \small
  \setlength{\tabcolsep}{8pt}
  \renewcommand{\arraystretch}{1.15}
  \begin{tabular}{l|ccc}
    \toprule
      & \textbf{PixelREPA-B} & \textbf{PixelREPA-L} & \textbf{PixelREPA-H}\\
    \specialrule{0.6pt}{0pt}{0pt}
    \rowcolor{sectiongray}
    \multicolumn{4}{l}{\textit{Architecture}} \\
    depth       & 12   & 24   & 32   \\
    hidden dim  & 768  & 1024 & 1280 \\
    heads       & 12   & 16   & 16   \\
    image size  & \multicolumn{3}{c}{256} \\
    patch size  & \multicolumn{3}{c}{$\text{(image size)} / 16$} \\
    in-context class tokens & \multicolumn{3}{c}{32} \\
    in-context start block  & 4 & 8 & 10 \\
    alignment depth & \multicolumn{3}{c}{$\text{(in-context start block)} - 1$} \\
    block number of MTA & \multicolumn{3}{c}{2} \\
    \specialrule{0.6pt}{0pt}{0pt}
    \rowcolor{sectiongray}
    \multicolumn{4}{l}{\textit{Training}} \\
    epochs      & \multicolumn{3}{c}{50, 100, 200, 300, 600} \\
    optimizer   & \multicolumn{3}{c}{Adam~\cite{diederik2014adam}, $(\beta_{1}, \beta_{2}) = (0.9, 0.95)$} \\
    batch size  & \multicolumn{3}{c}{1024} \\
    learning rate   & \multicolumn{3}{c}{$2 \times 10^{-4}$} \\
    ema decay   & \multicolumn{3}{c}{{0.9996, 0.9998, 0.9999}} \\

    \specialrule{0.6pt}{0pt}{0pt}
    \rowcolor{sectiongray}
    \multicolumn{4}{l}{\textit{Sampling}} \\
    ODE solver  & \multicolumn{3}{c}{Heun} \\
    ODE steps   & \multicolumn{3}{c}{50} \\
    time steps  & \multicolumn{3}{c}{linear in $[0.0, 1.0]$} \\
    CFG interval  & \multicolumn{3}{c}{$[0.1, 1.0]$} \\
    
    \bottomrule
  \end{tabular}
  \caption{\textbf{Model configuration details.}}
  \label{table:appen_model_configuration}
\end{table}

\begin{figure}[h]
  \centering
  \includegraphics[width=0.8\textwidth]{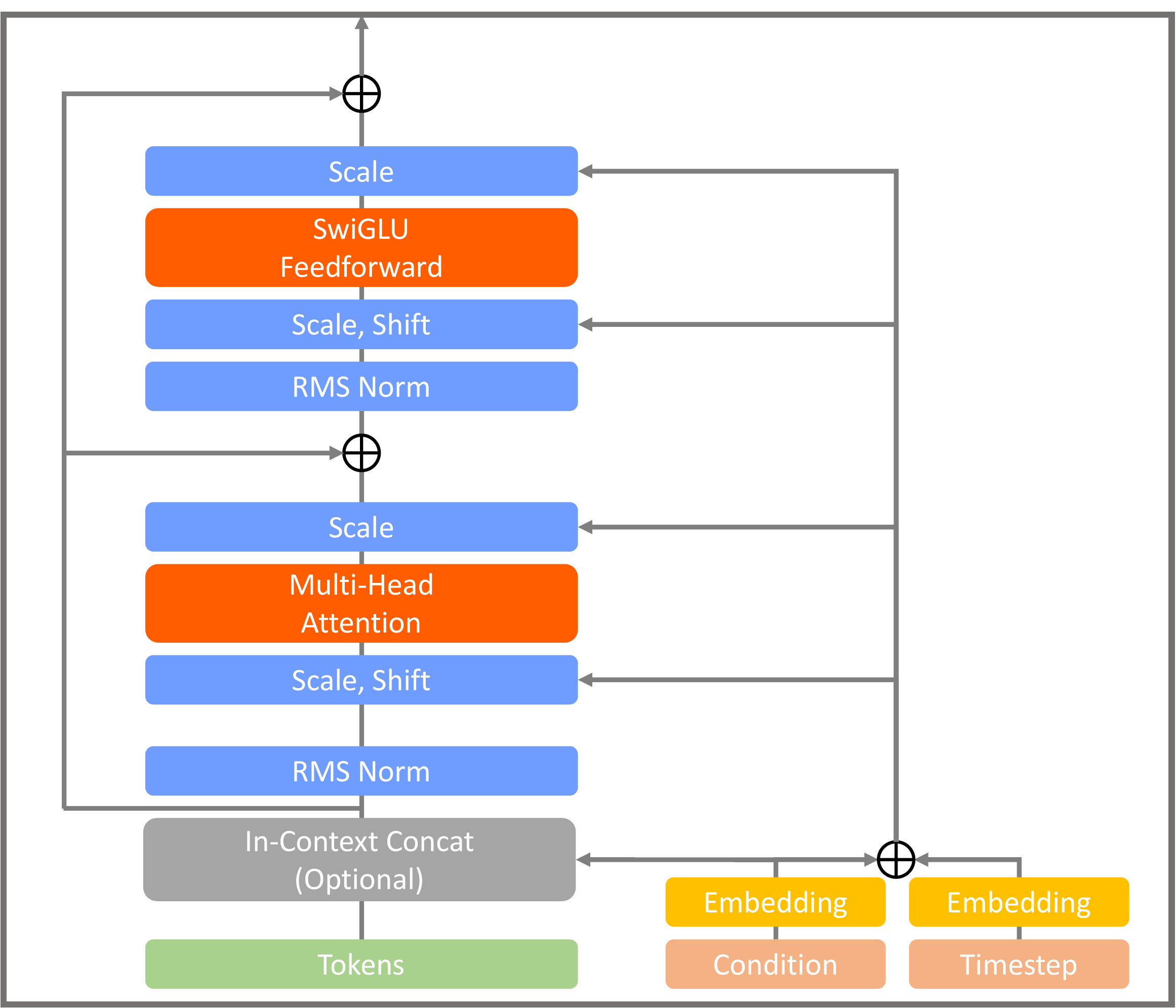}
  \caption{
  \textbf{Implementation of the JiT block.}
  The in-context concatenation is operated only after the predefined in-context start block.
  }
  \label{fig:appen_JiT}
\end{figure}

PixelREPA strictly follows the original configurations details of Just image Transformers (JiT)~\cite{li2025back} as \cref{table:appen_model_configuration}.
We use Adam optimizer~\cite{diederik2014adam} with constant learning rate of $2 \times 10^{-4}$ and $(\beta_{1}, \beta_{2}) = (0.9, 0.95)$.
This exactly same setting shows our Masked Transformers Adapter (MTA) is effective for JiT.

\Cref{fig:appen_JiT} illustrates the JiT block. 
This architecture is closely related to Diffusion Transformers (DiT)~\cite{peebles2023scalable} and Scalable Interpolant Transformers (SiT)~\cite{ma2024sit}. 
JiT uses AdaLN-Zero modulation in each attention block as DiT and SiT.
A key difference is that JiT adopts \textit{in-context concatenation}, unlike DiT and SiT.
Specifically, JiT concatenates condition embeddings and tokens from a previous block as \cref{fig:appen_JiT}.
This operation is applied only after a predefined in-context start block, whose index is listed in \cref{table:appen_model_configuration}. 
Since in-context concatenation strongly injects conditional information, we consistently apply representation alignment at the block immediately before the in the context start block.
Furthermore, MTA is consisted of two JiT blocks.

\section{Qualitative Results}
\label{sec:appen_qual}

\begin{figure*}[h]
\centering
\setlength{\tabcolsep}{10pt}  
\renewcommand{\arraystretch}{1.0}

\begin{tabular}{@{}cc@{}}

\begin{minipage}[t]{0.48\textwidth}
  \centering
  \includegraphics[width=\linewidth]{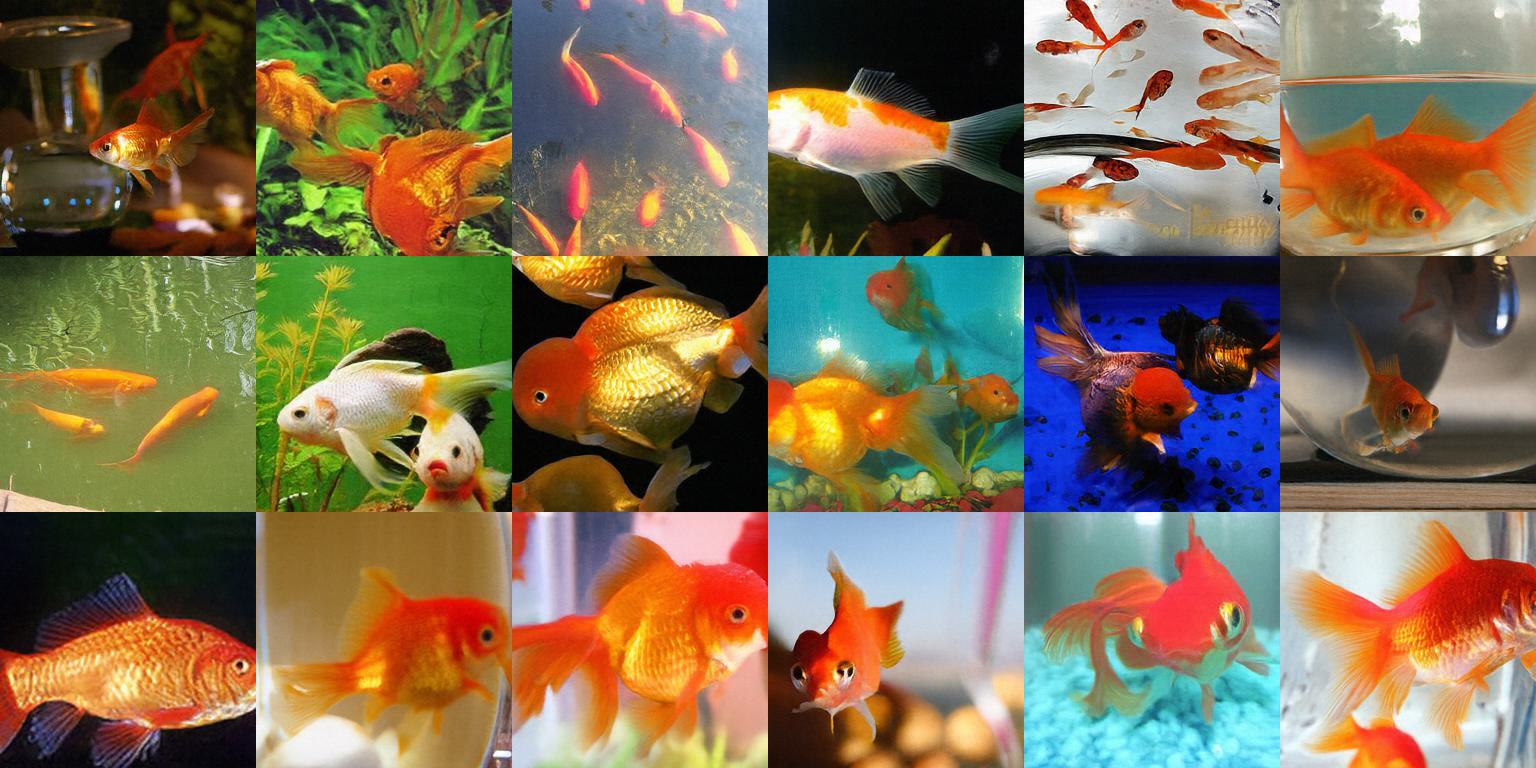}
  \vspace{-2.1em}
  \caption*{class n01443537}
\end{minipage}
&
\begin{minipage}[t]{0.48\textwidth}
  \centering
  \includegraphics[width=\linewidth]{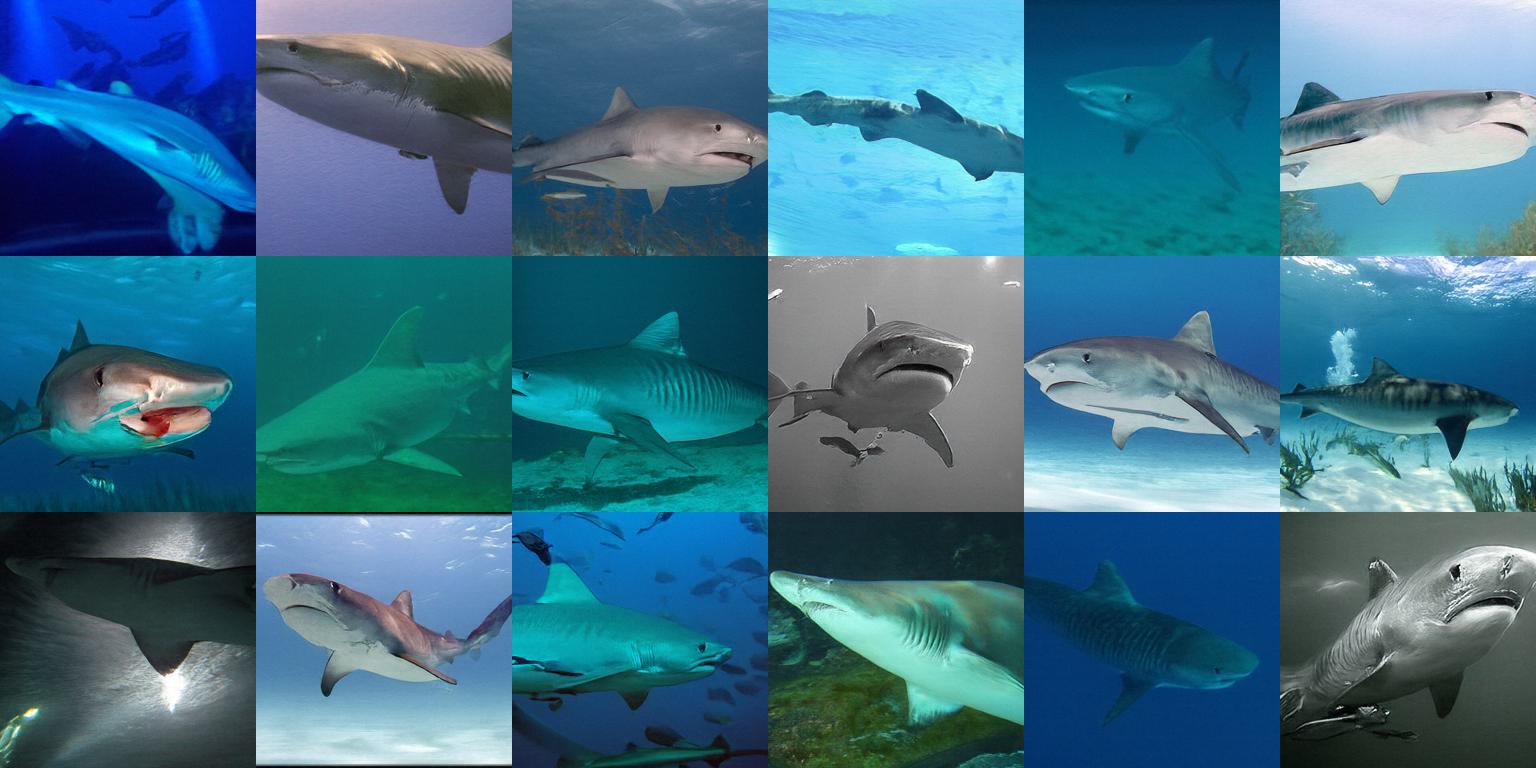}
  \vspace{-2.1em}
  \caption*{class n01491361}
\end{minipage}
\\[1.5pt] 

\begin{minipage}[t]{0.48\textwidth}
  \centering
  \includegraphics[width=\linewidth]{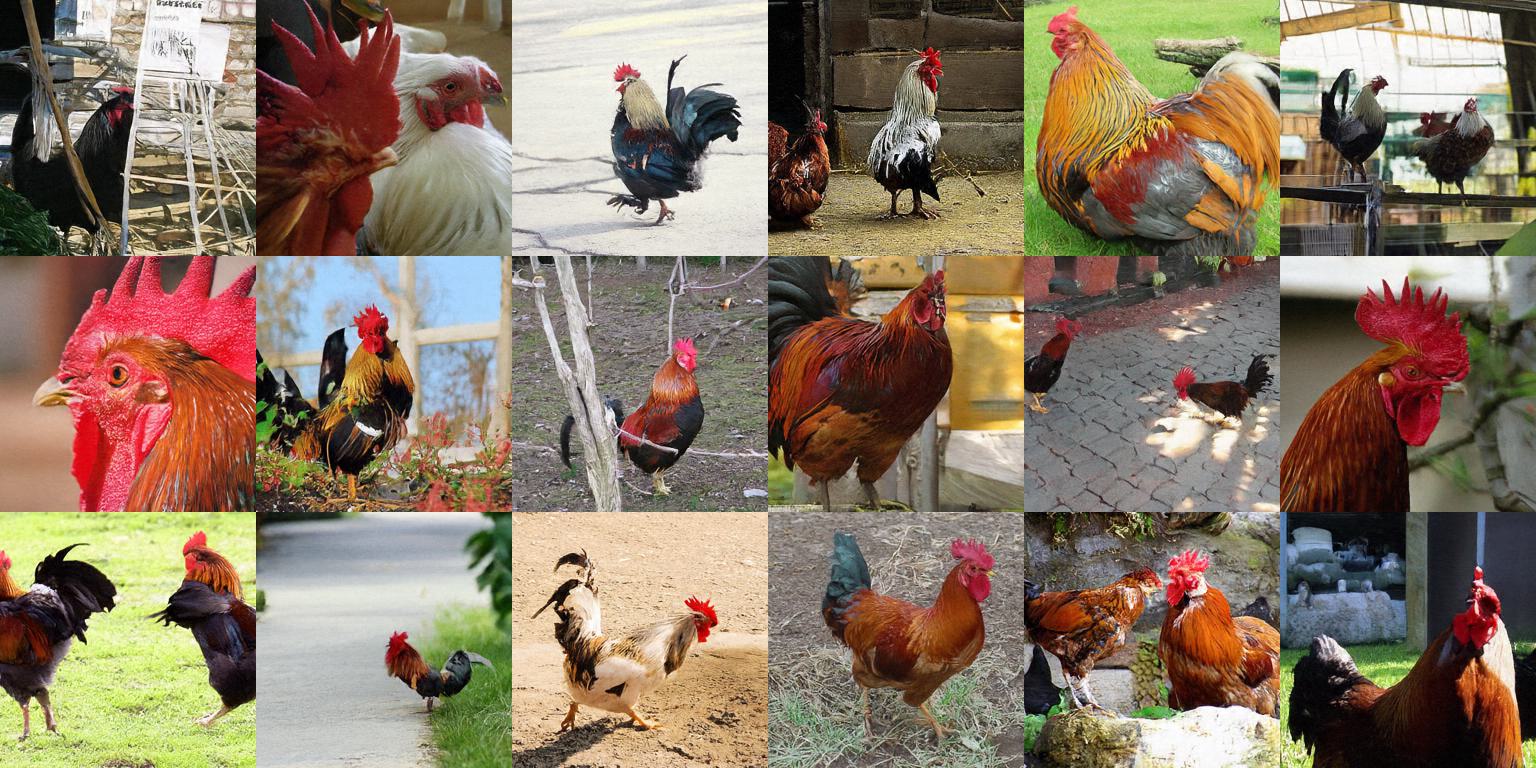}
  \vspace{-2.1em}
  \caption*{class n01514668}
\end{minipage}
&
\begin{minipage}[t]{0.48\textwidth}
  \centering
  \includegraphics[width=\linewidth]{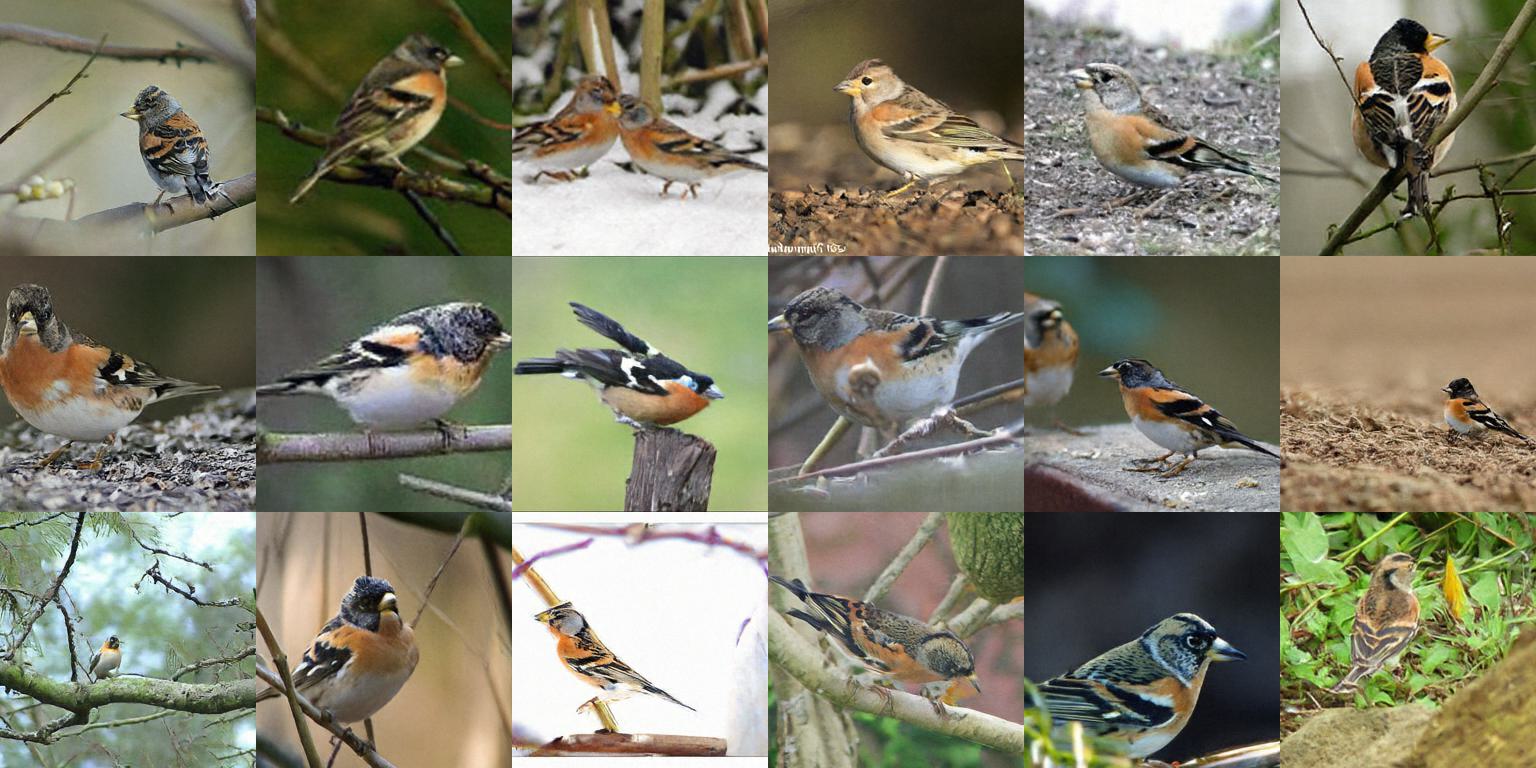}
  \vspace{-2.1em}
  \caption*{class n01530575}
\end{minipage}
\\[1.5pt] 

\begin{minipage}[t]{0.48\textwidth}
  \centering
  \includegraphics[width=\linewidth]{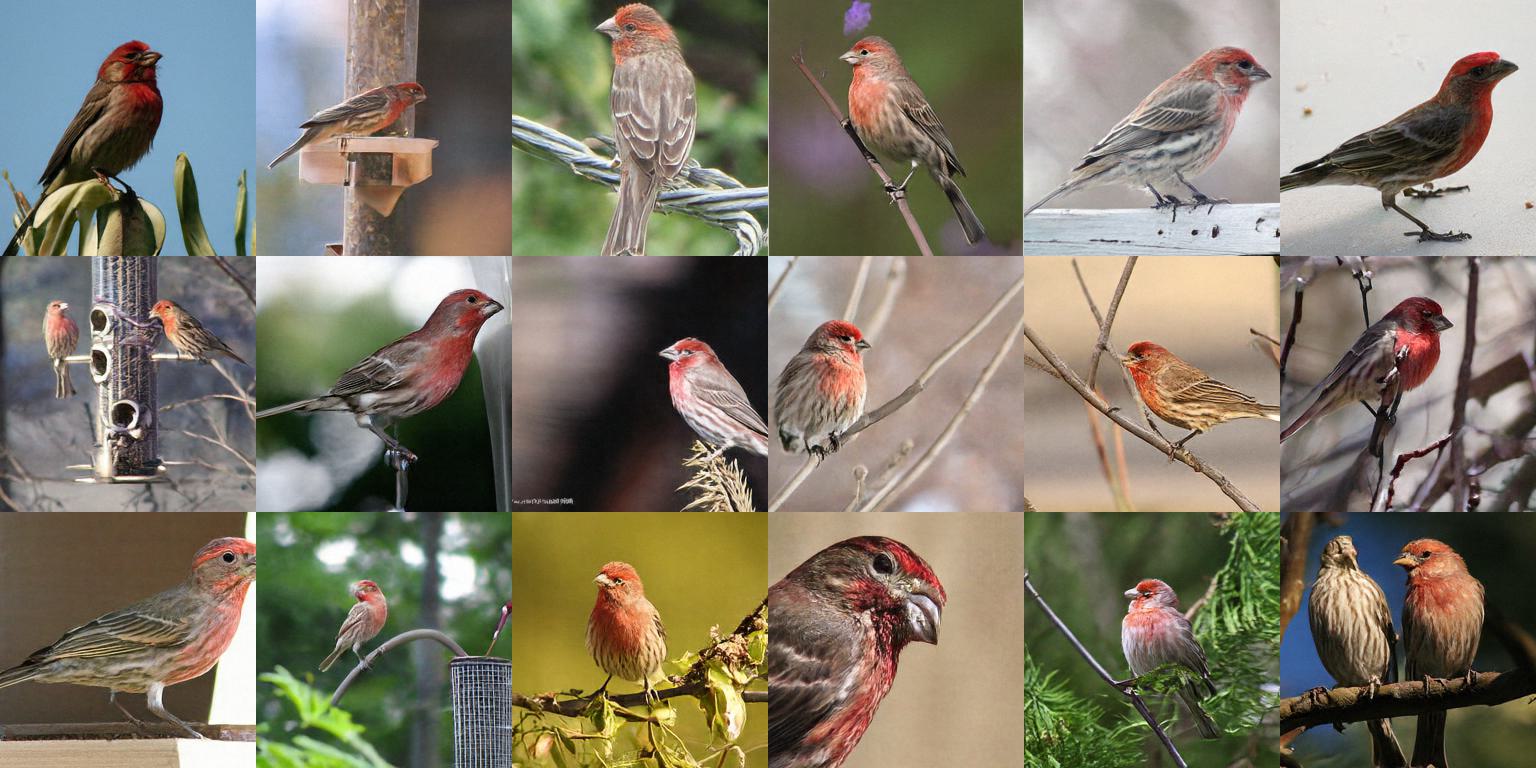}
  \vspace{-2.1em}
  \caption*{class n01532829}
\end{minipage}
&
\begin{minipage}[t]{0.48\textwidth}
  \centering
  \includegraphics[width=\linewidth]{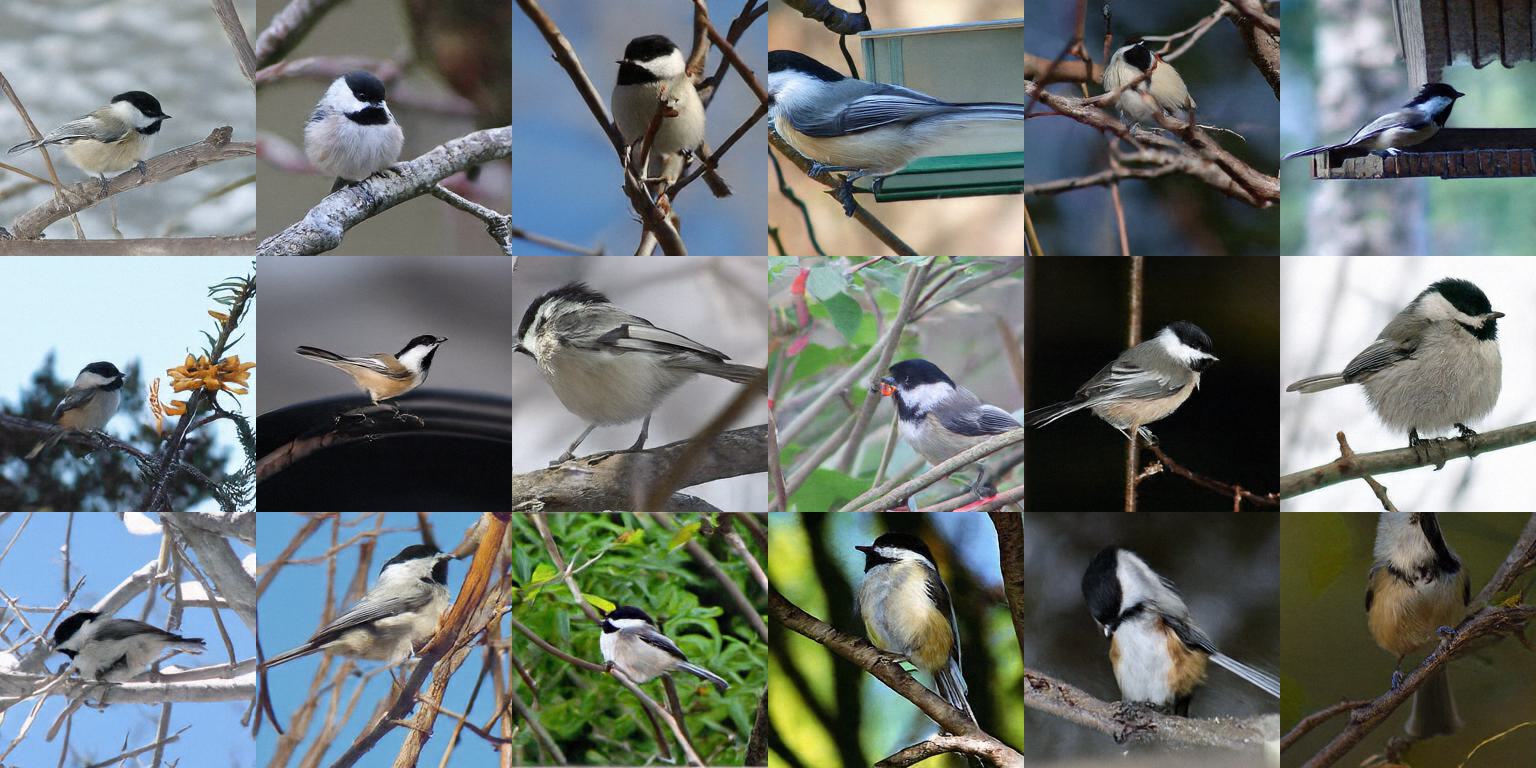}
  \vspace{-2.1em}
  \caption*{class n01592084}
\end{minipage}
\\[1.5pt] 

\begin{minipage}[t]{0.48\textwidth}
  \centering
  \includegraphics[width=\linewidth]{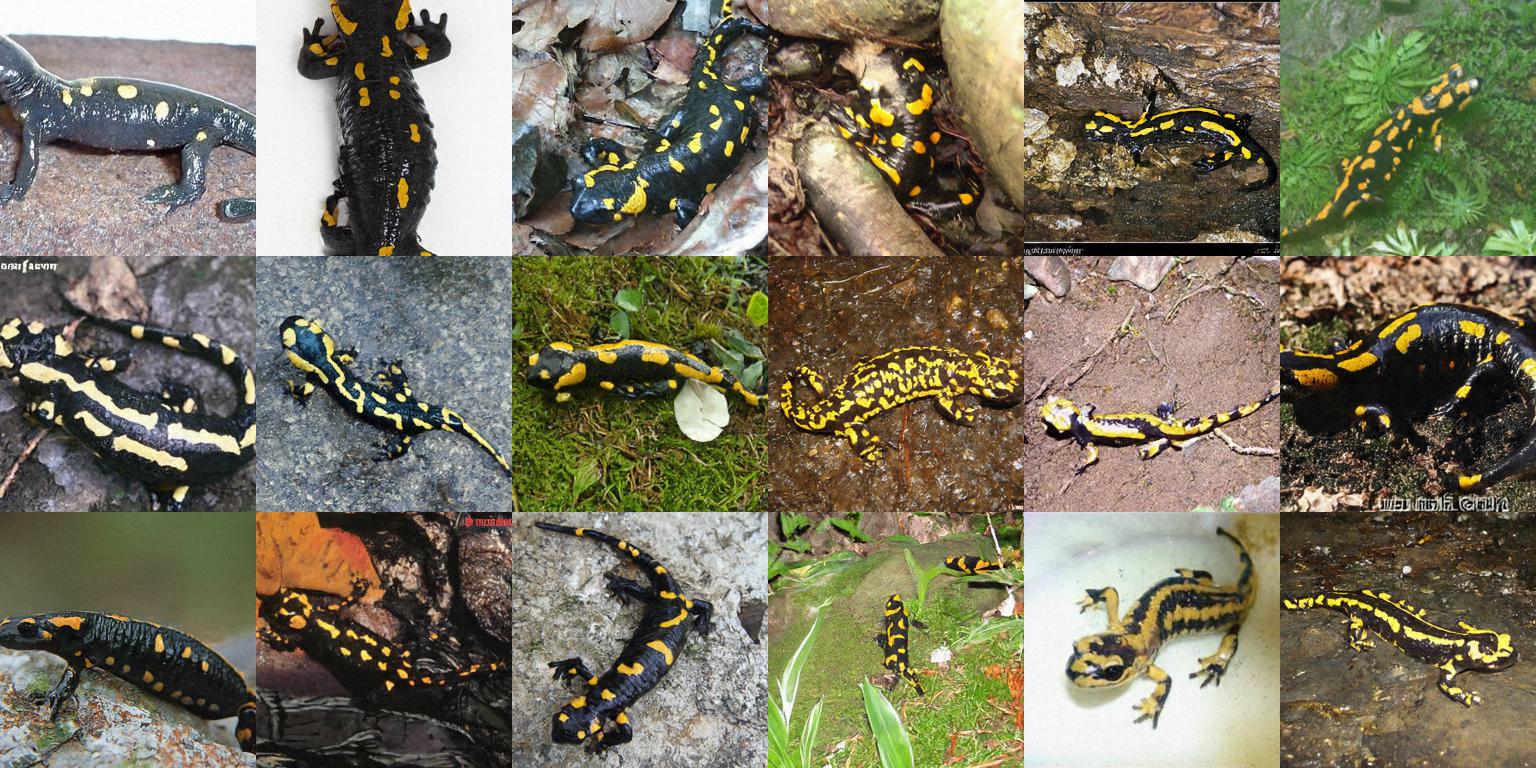}
  \vspace{-2.1em}
  \caption*{class n01629819}
\end{minipage}
&
\begin{minipage}[t]{0.48\textwidth}
  \centering
  \includegraphics[width=\linewidth]{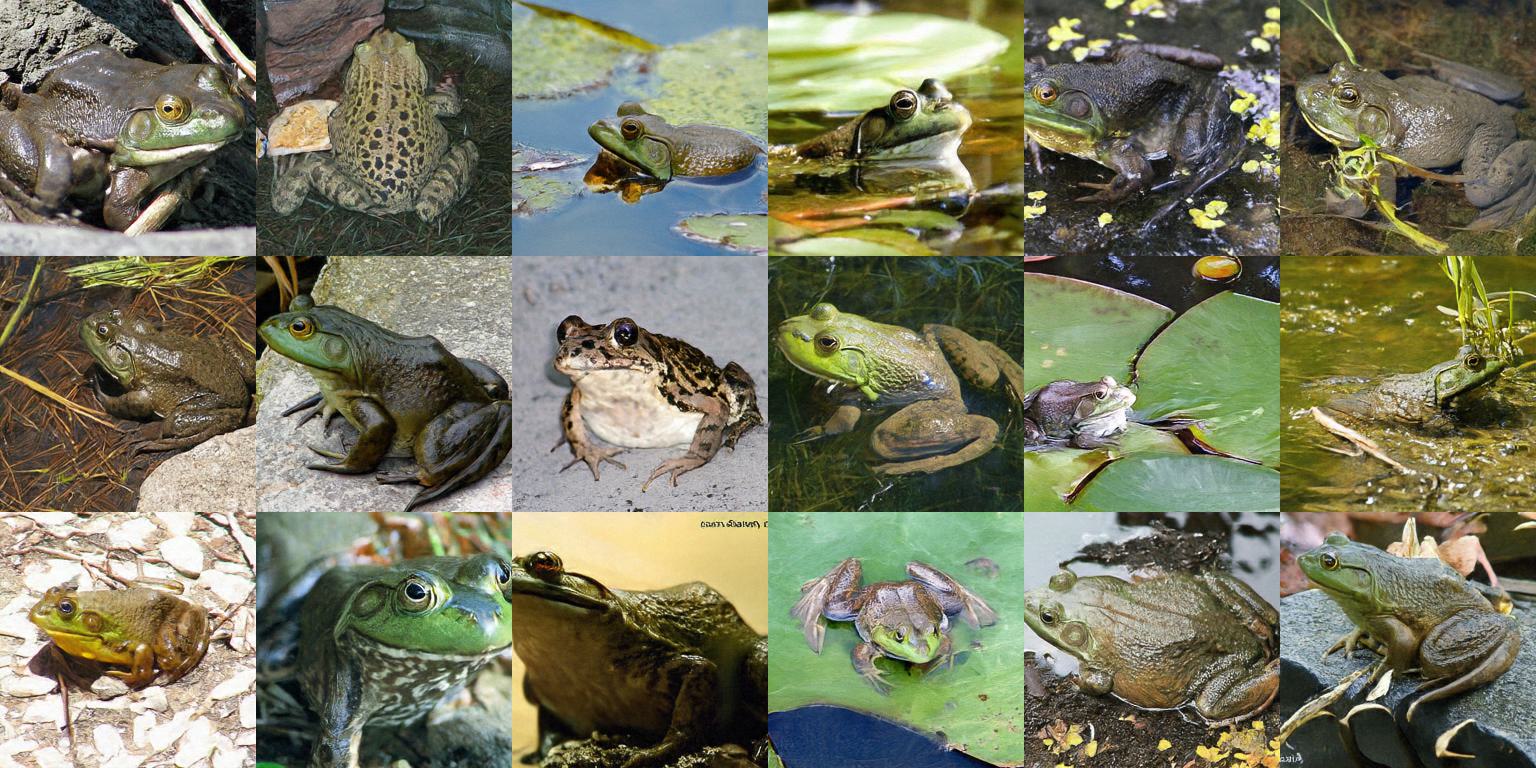}
  \vspace{-2.1em}
  \caption*{class n01641577}
\end{minipage}
\\[1.5pt] 

\begin{minipage}[t]{0.48\textwidth}
  \centering
  \includegraphics[width=\linewidth]{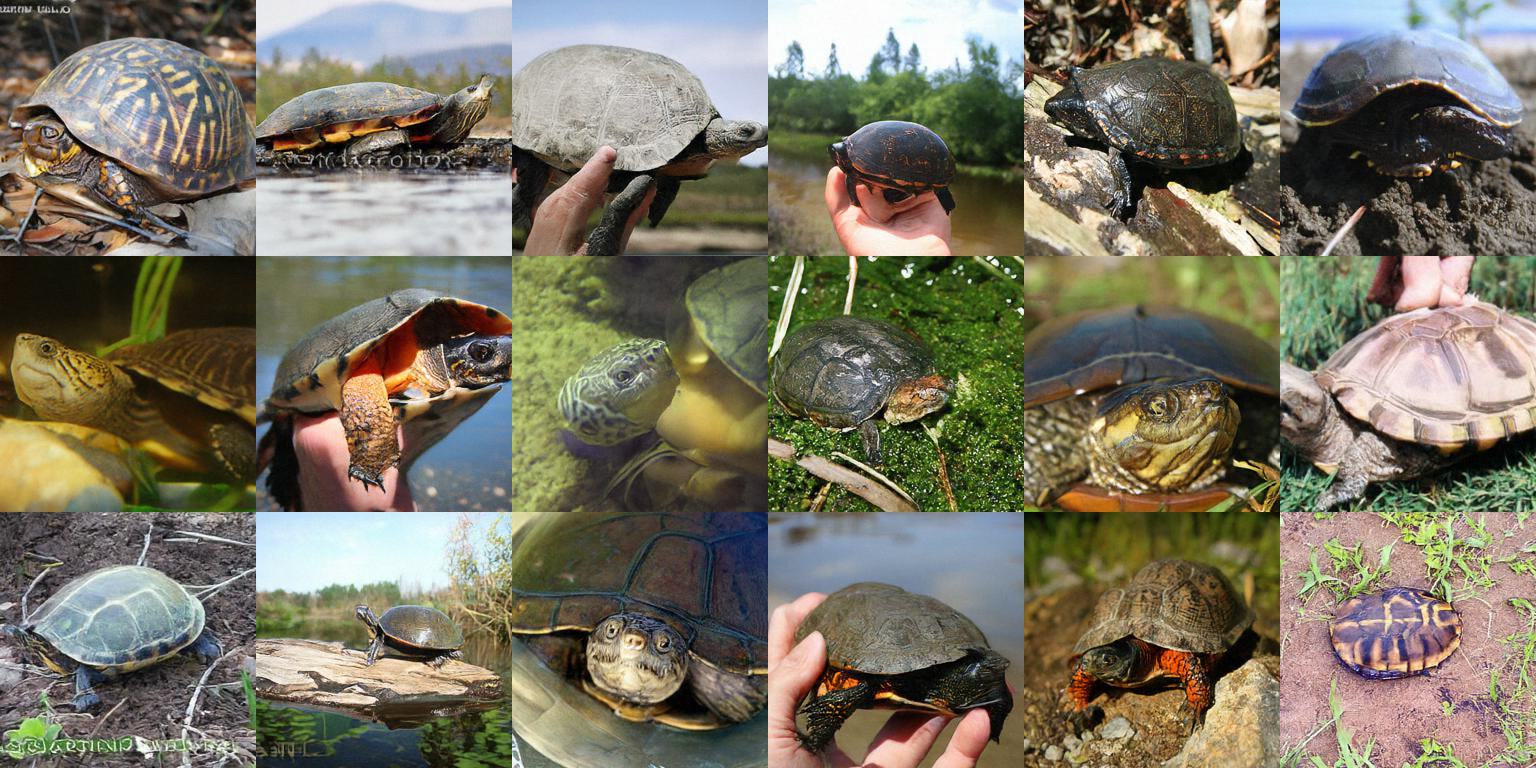}
  \vspace{-2.1em}
  \caption*{class n01667114}
\end{minipage}
&
\begin{minipage}[t]{0.48\textwidth}
  \centering
  \includegraphics[width=\linewidth]{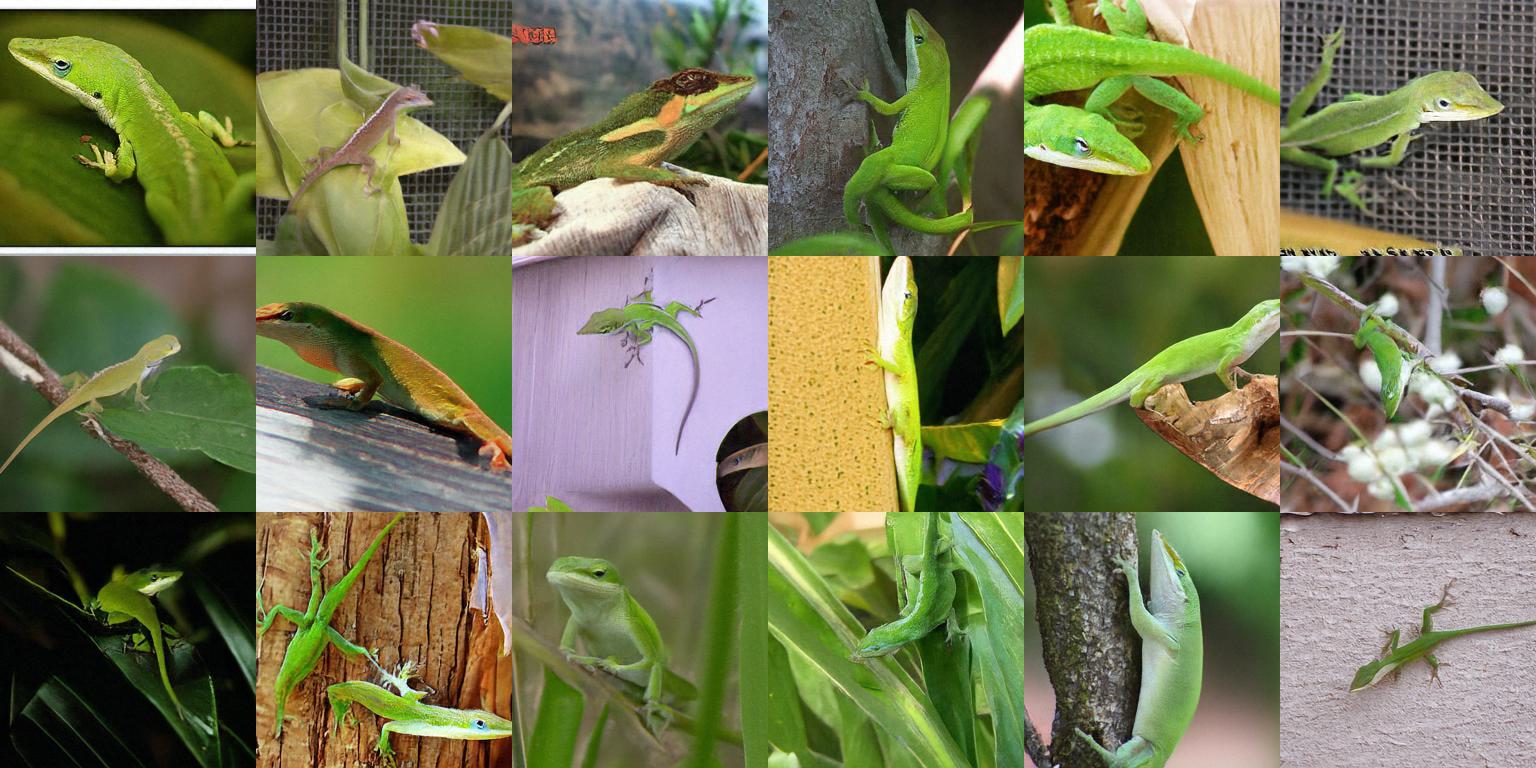}
  \vspace{-2.1em}
  \caption*{class n01682714}
\end{minipage}
\\[1.5pt] 

\end{tabular}

\caption{\textbf{Uncurated samples of PixelREPA/H-16 on ImageNet 256$\times$256~\cite{deng2009imagenet}.}}
\label{fig:appen_uncurated_1}
\end{figure*}

\begin{figure*}[h]
\centering
\setlength{\tabcolsep}{10pt}  
\renewcommand{\arraystretch}{1.0}

\begin{tabular}{@{}cc@{}}

\begin{minipage}[t]{0.48\textwidth}
  \centering
  \includegraphics[width=\linewidth]{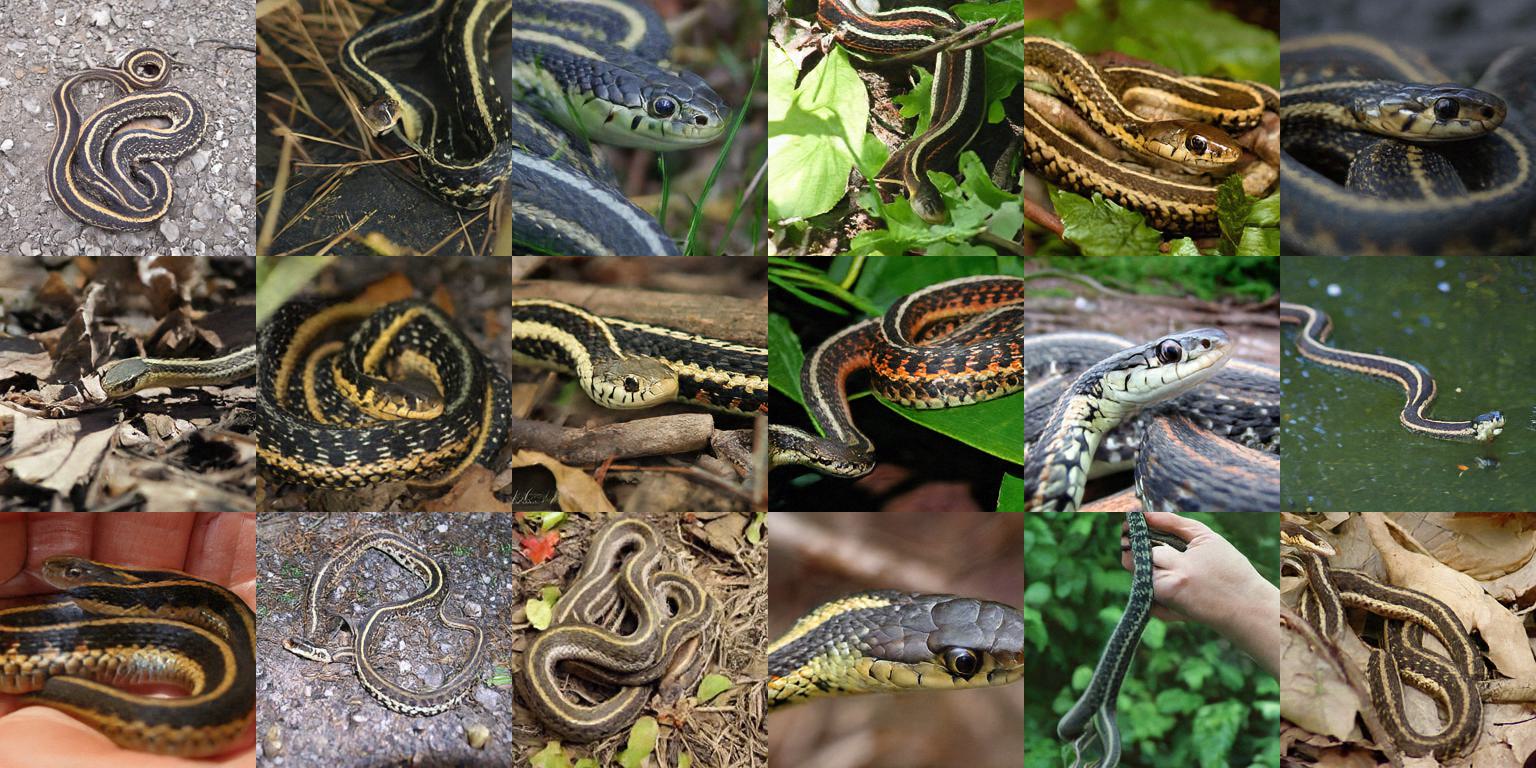}
  \vspace{-2.1em}
  \caption*{class n01735189}
\end{minipage}
&
\begin{minipage}[t]{0.48\textwidth}
  \centering
  \includegraphics[width=\linewidth]{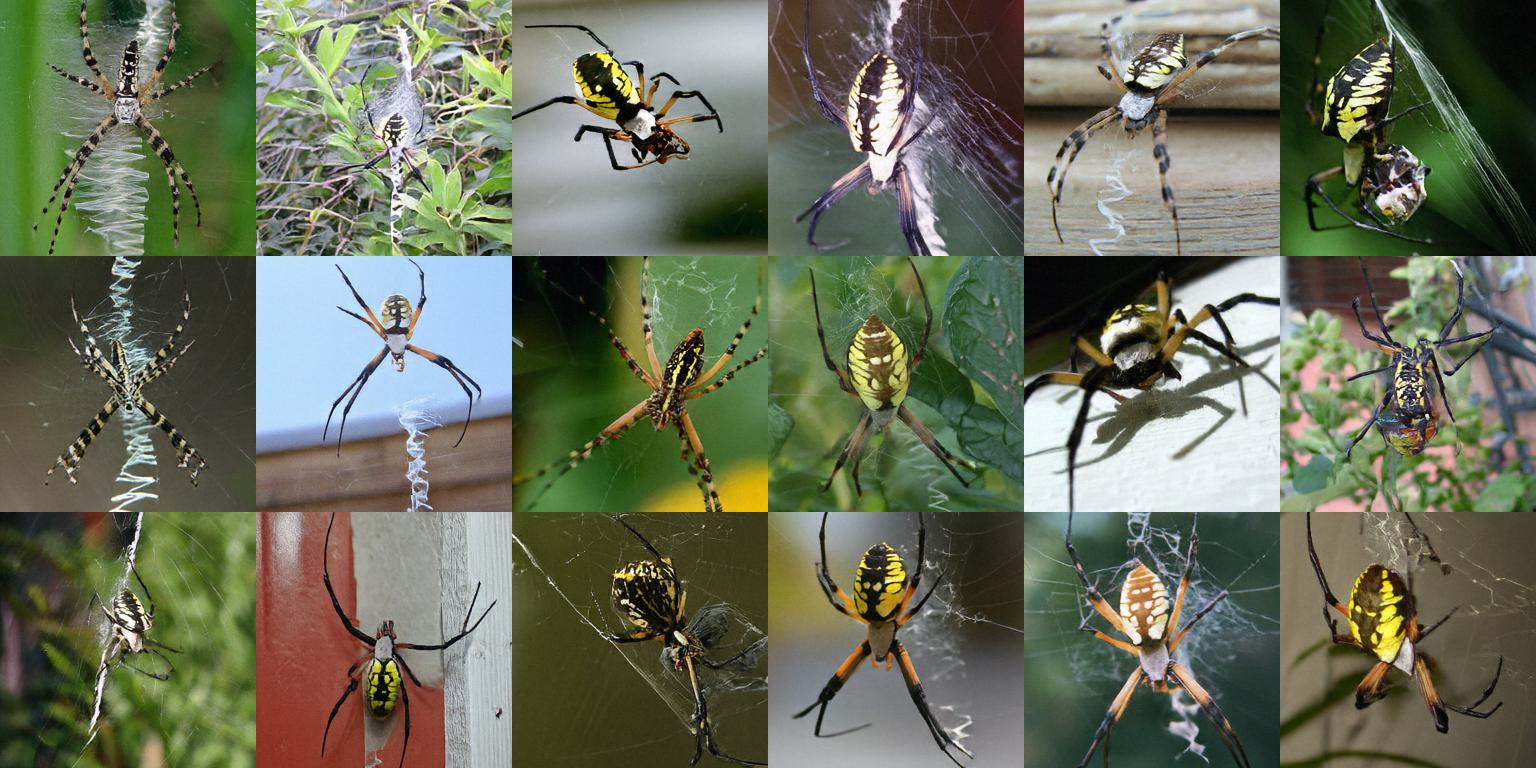}
  \vspace{-2.1em}
  \caption*{class n01773157}
\end{minipage}
\\[1.5pt] 

\begin{minipage}[t]{0.48\textwidth}
  \centering
  \includegraphics[width=\linewidth]{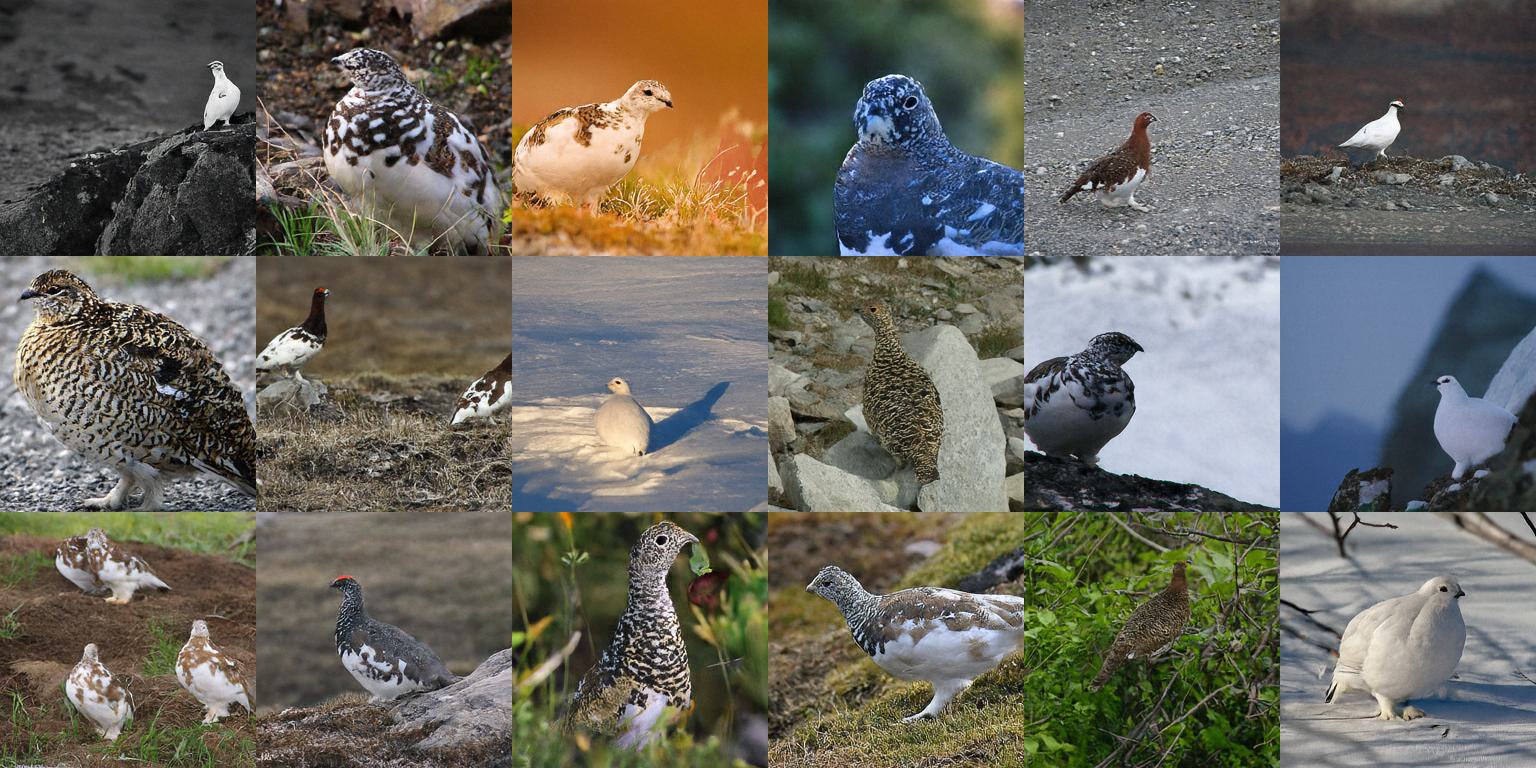}
  \vspace{-2.1em}
  \caption*{class n01796340}
\end{minipage}
&
\begin{minipage}[t]{0.48\textwidth}
  \centering
  \includegraphics[width=\linewidth]{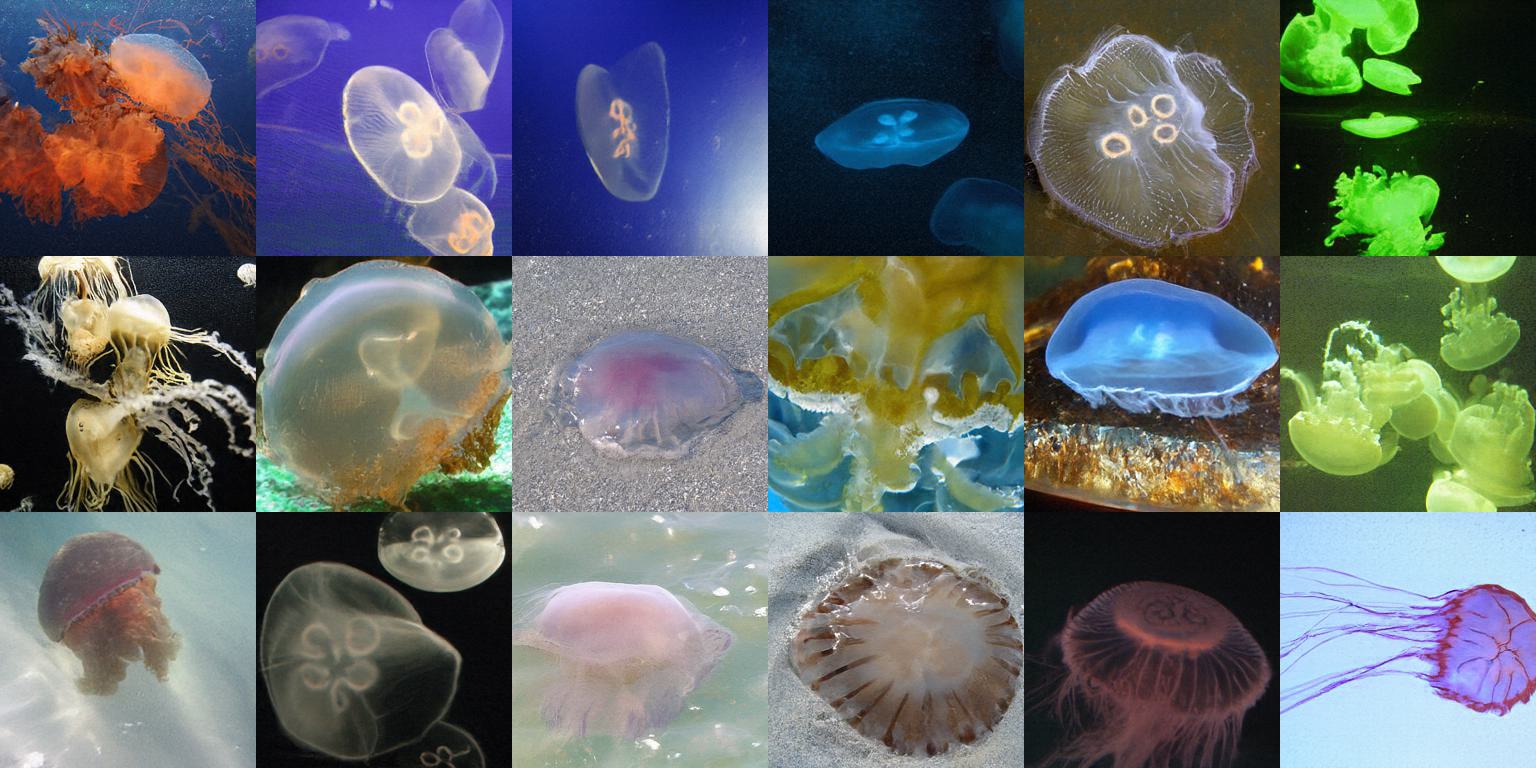}
  \vspace{-2.1em}
  \caption*{class n01910747}
\end{minipage}
\\[1.5pt] 

\begin{minipage}[t]{0.48\textwidth}
  \centering
  \includegraphics[width=\linewidth]{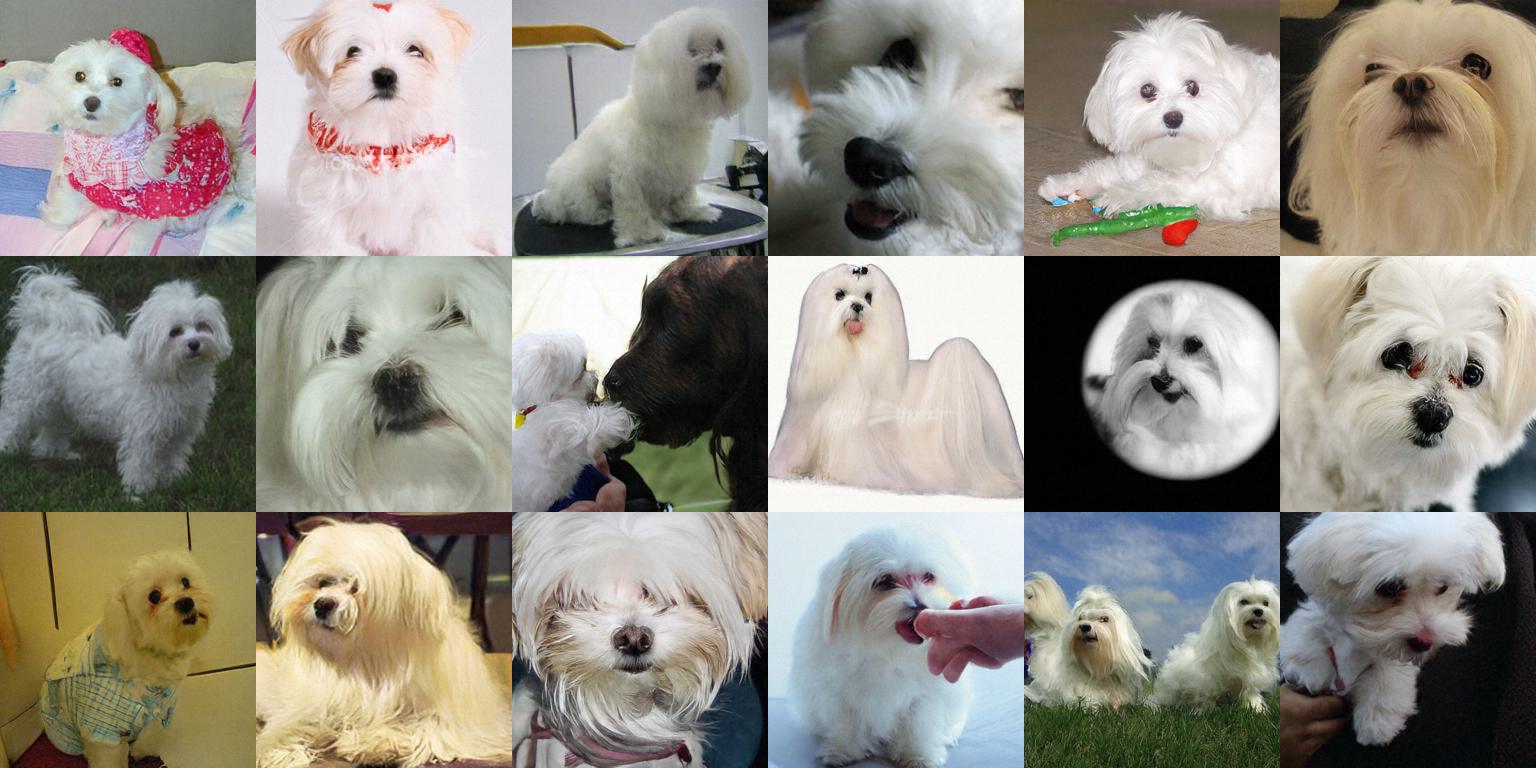}
  \vspace{-2.1em}
  \caption*{class n02085936}
\end{minipage}
&
\begin{minipage}[t]{0.48\textwidth}
  \centering
  \includegraphics[width=\linewidth]{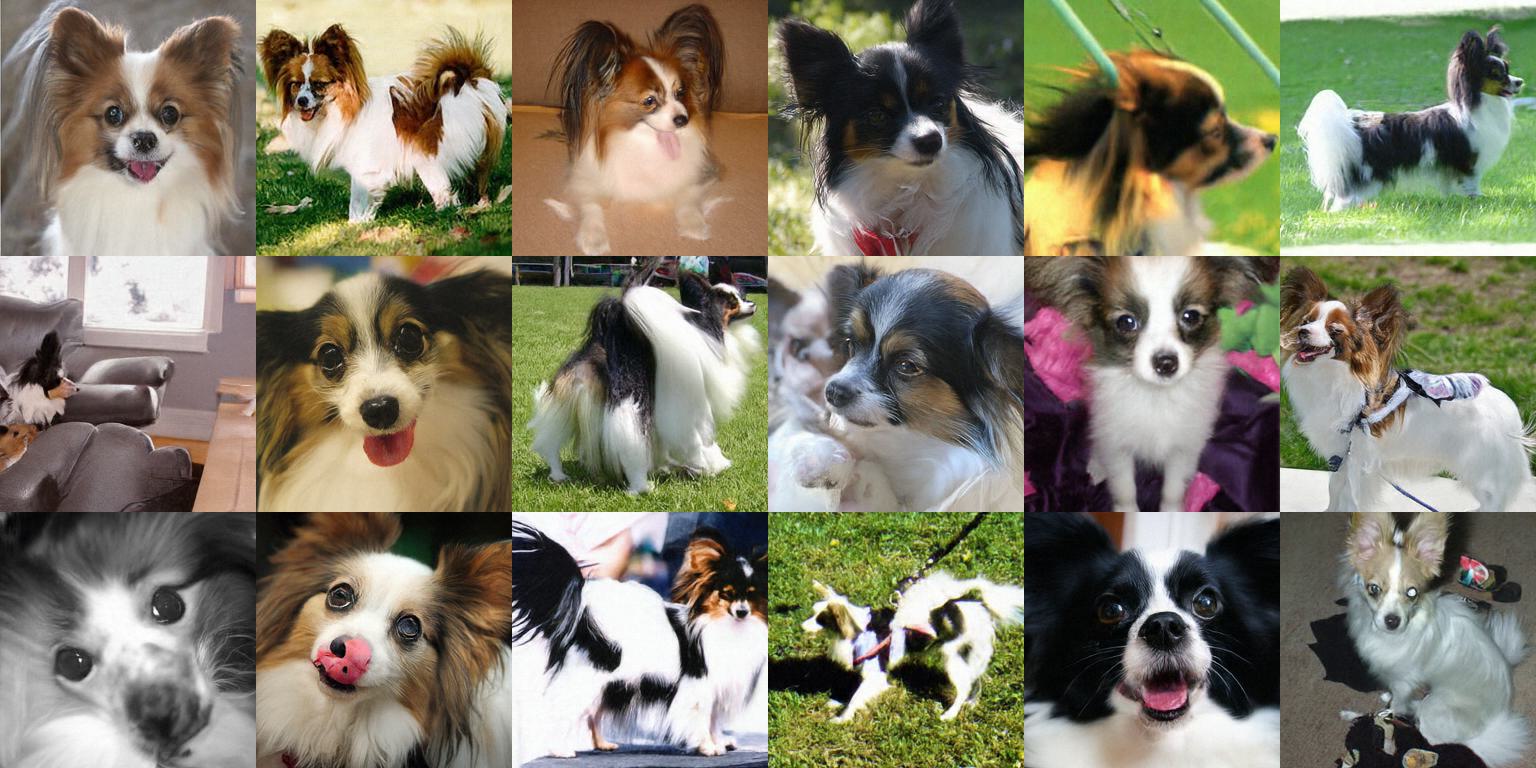}
  \vspace{-2.1em}
  \caption*{class n02086910}
\end{minipage}
\\[1.5pt] 

\begin{minipage}[t]{0.48\textwidth}
  \centering
  \includegraphics[width=\linewidth]{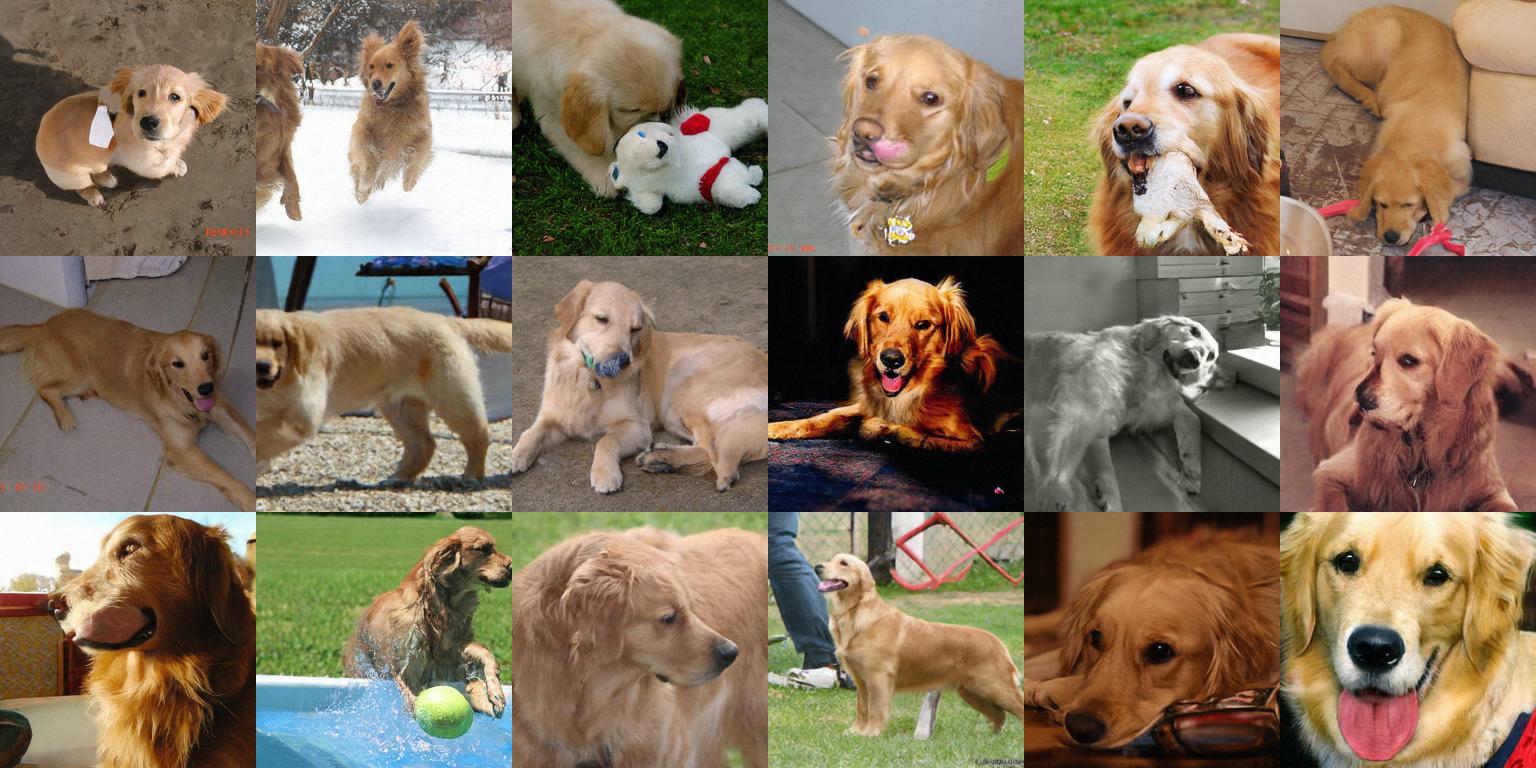}
  \vspace{-2.1em}
  \caption*{class n02099601}
\end{minipage}
&
\begin{minipage}[t]{0.48\textwidth}
  \centering
  \includegraphics[width=\linewidth]{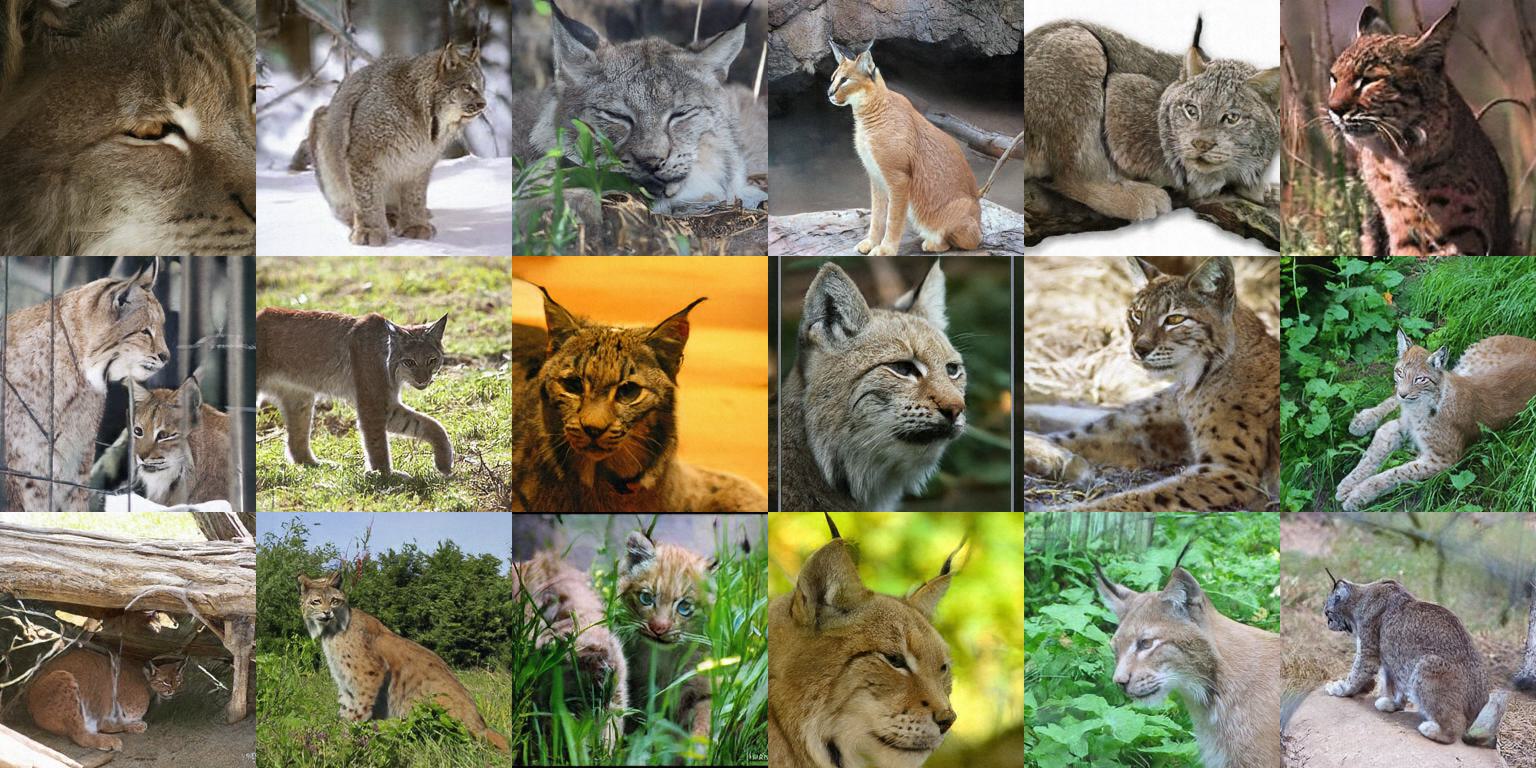}
  \vspace{-2.1em}
  \caption*{class n02127052}
\end{minipage}
\\[1.5pt] 

\begin{minipage}[t]{0.48\textwidth}
  \centering
  \includegraphics[width=\linewidth]{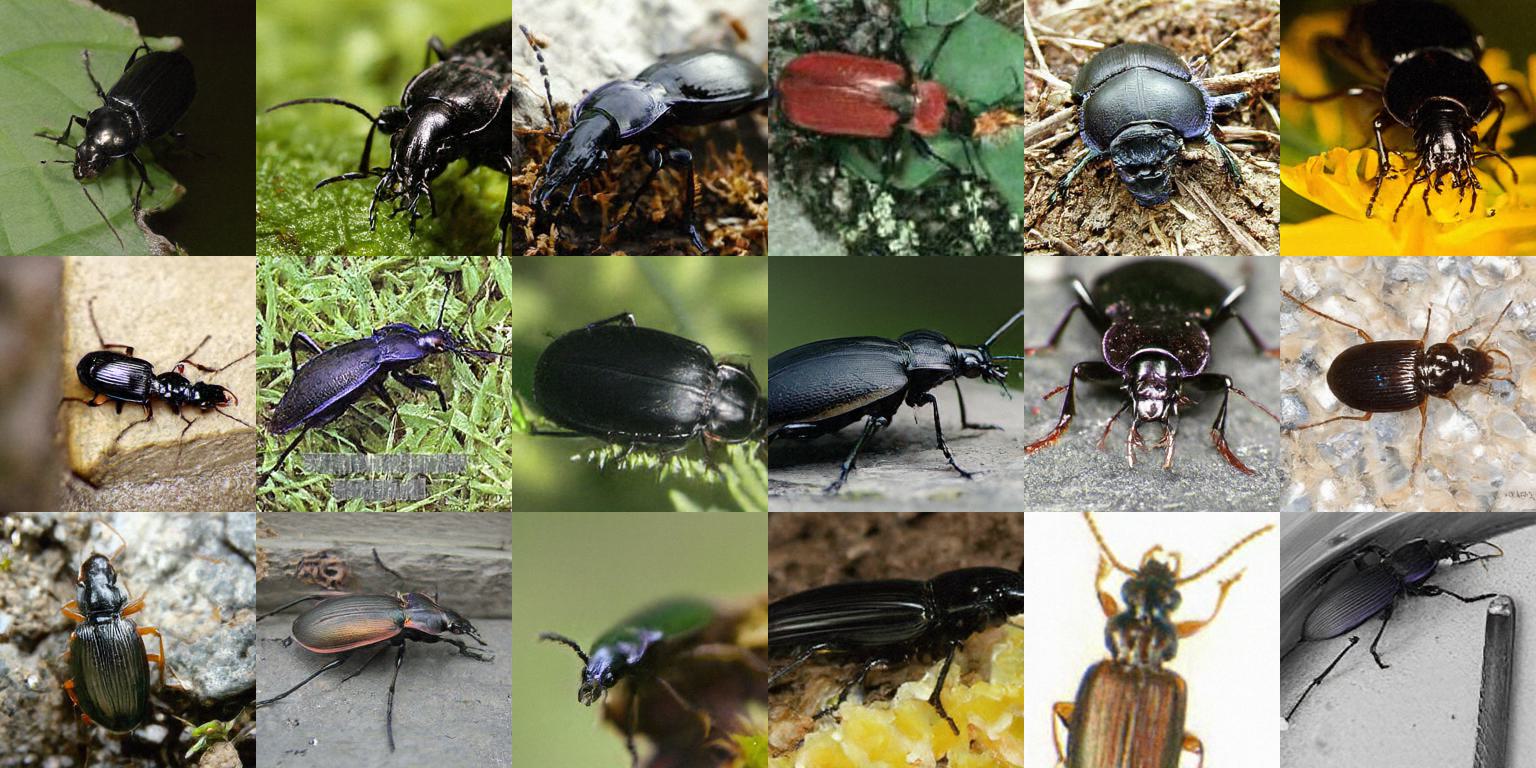}
  \vspace{-2.1em}
  \caption*{class n02167151}
\end{minipage}
&
\begin{minipage}[t]{0.48\textwidth}
  \centering
  \includegraphics[width=\linewidth]{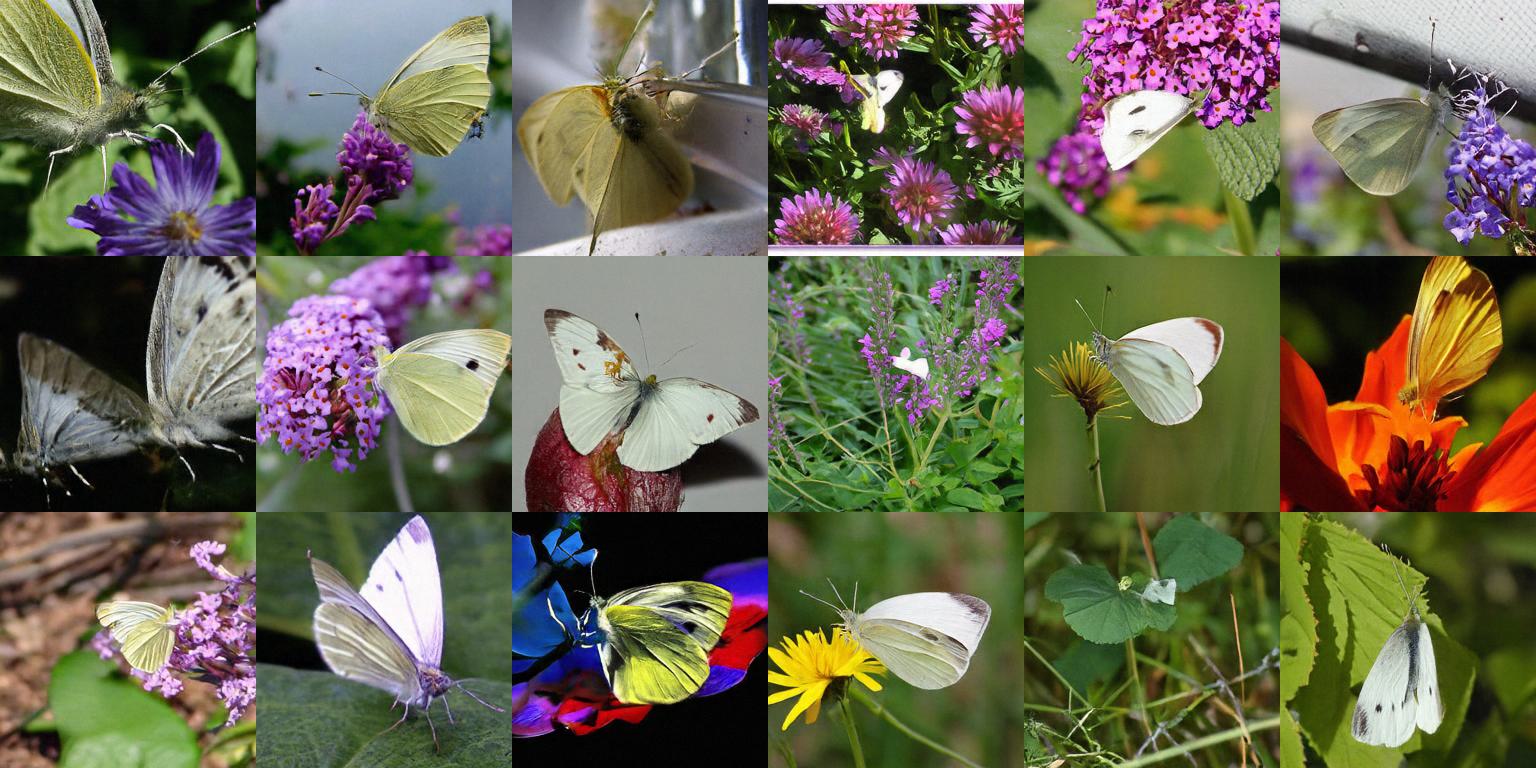}
  \vspace{-2.1em}
  \caption*{class n02280649}
\end{minipage}
\\[1.5pt] 

\end{tabular}

\caption{\textbf{Uncurated samples of PixelREPA/H-16 on ImageNet 256$\times$256.}}
\label{fig:appen_uncurated_2}
\end{figure*}

\begin{figure*}[h]
\centering
\setlength{\tabcolsep}{10pt}  
\renewcommand{\arraystretch}{1.0}

\begin{tabular}{@{}cc@{}}

\begin{minipage}[t]{0.48\textwidth}
  \centering
  \includegraphics[width=\linewidth]{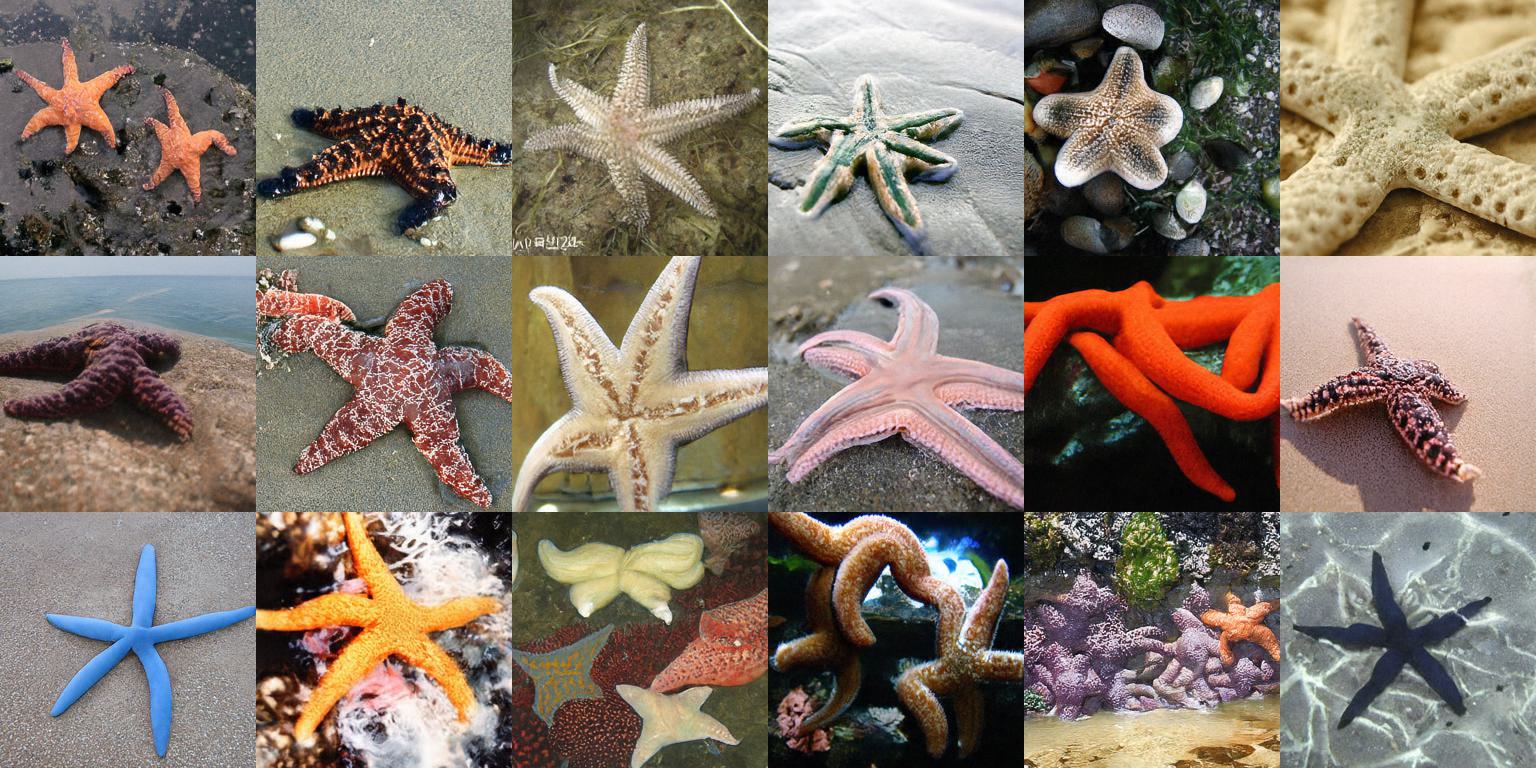}
  \vspace{-2.1em}
  \caption*{class n02317335}
\end{minipage}
&
\begin{minipage}[t]{0.48\textwidth}
  \centering
  \includegraphics[width=\linewidth]{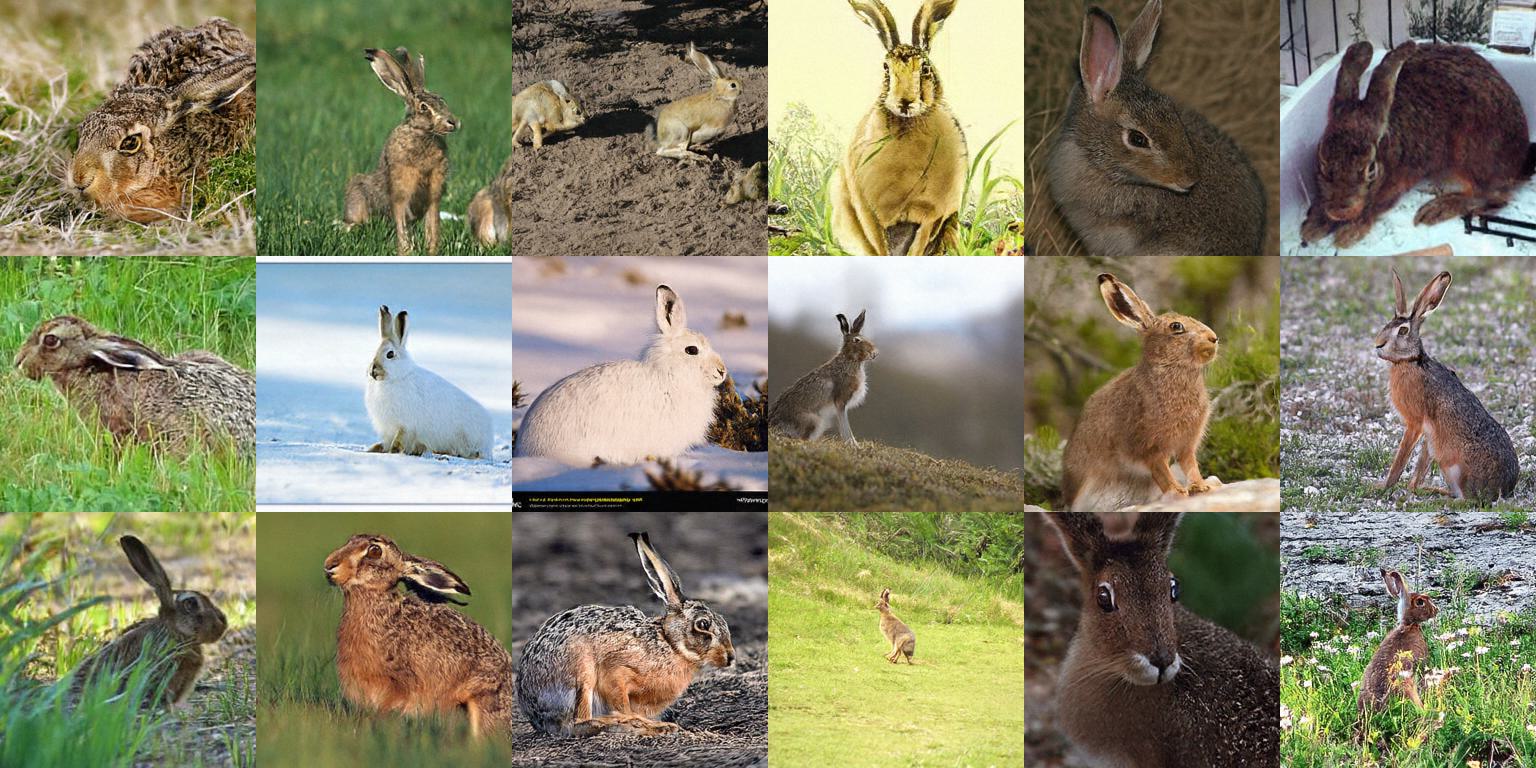}
  \vspace{-2.1em}
  \caption*{class n02326432}
\end{minipage}
\\[1.5pt] 

\begin{minipage}[t]{0.48\textwidth}
  \centering
  \includegraphics[width=\linewidth]{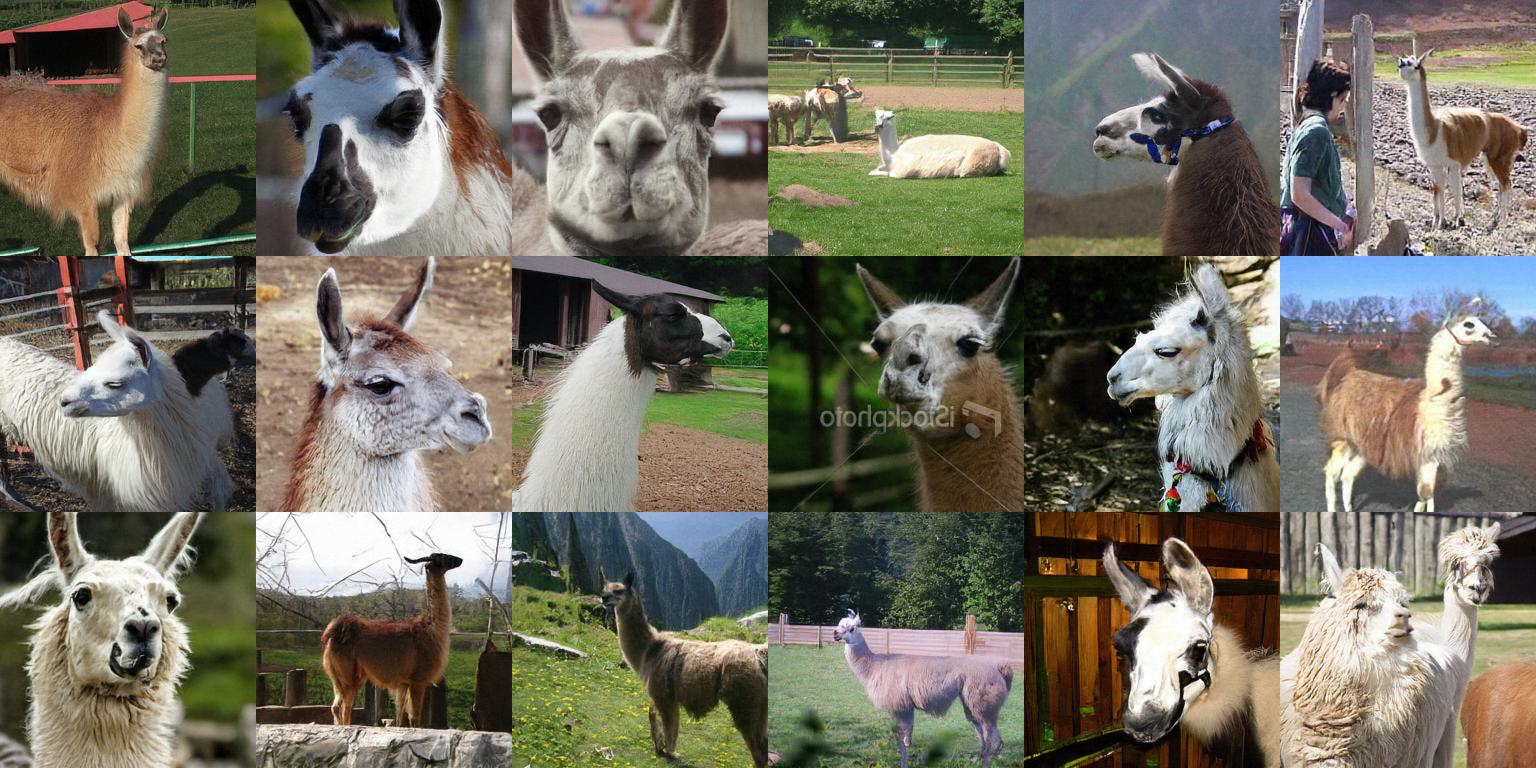}
  \vspace{-2.1em}
  \caption*{class n02437616}
\end{minipage}
&
\begin{minipage}[t]{0.48\textwidth}
  \centering
  \includegraphics[width=\linewidth]{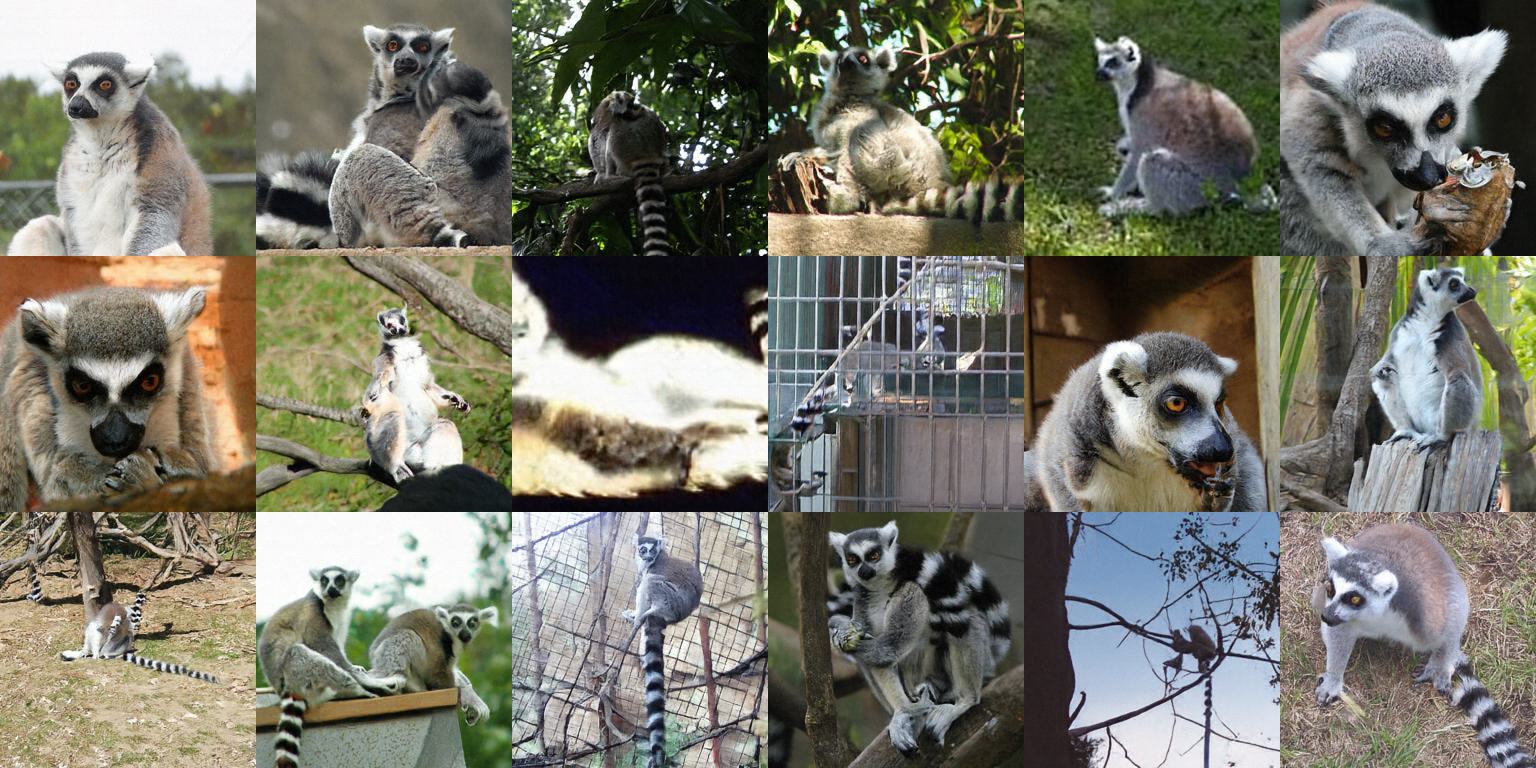}
  \vspace{-2.1em}
  \caption*{class n02497673}
\end{minipage}
\\[1.5pt] 

\begin{minipage}[t]{0.48\textwidth}
  \centering
  \includegraphics[width=\linewidth]{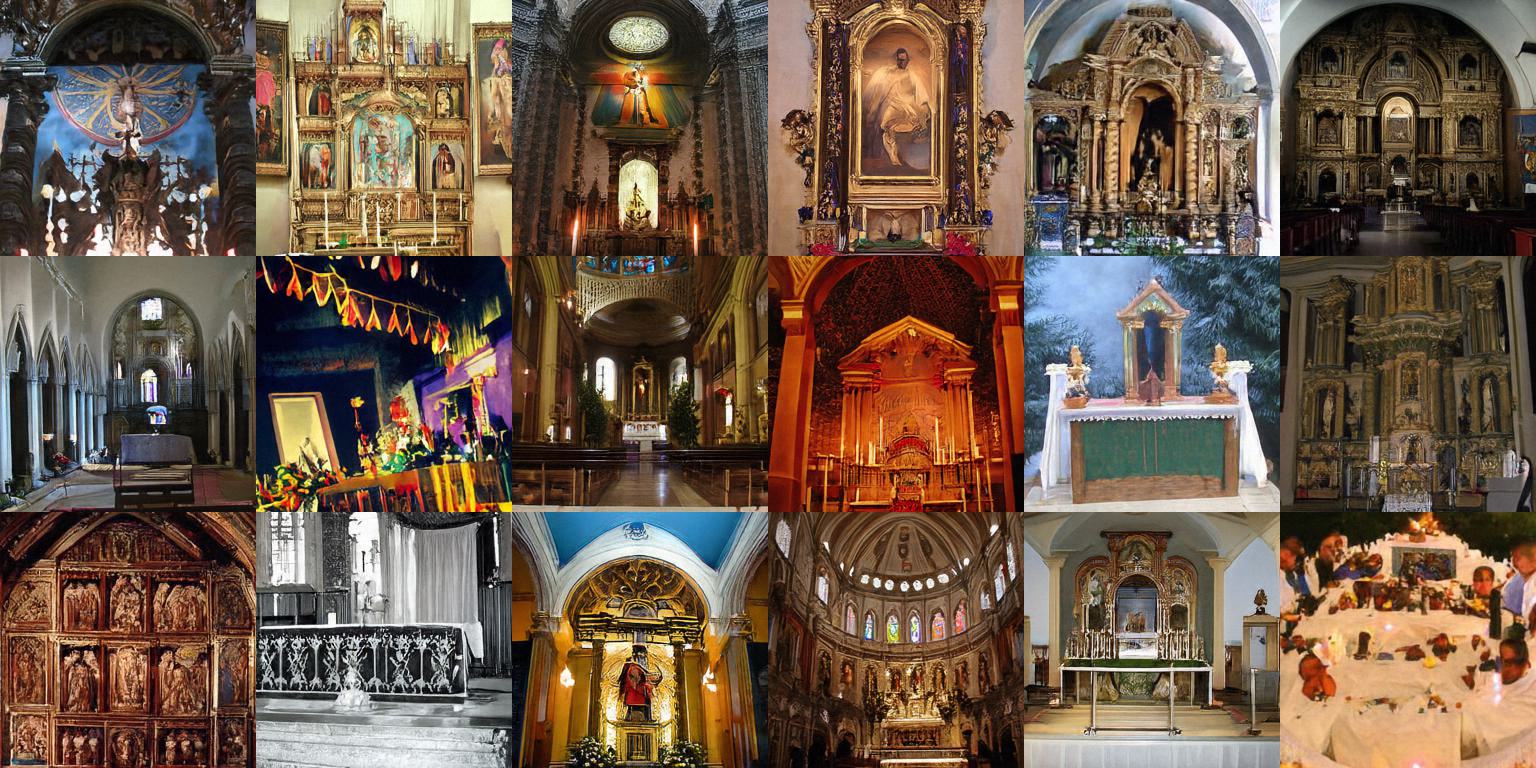}
  \vspace{-2.1em}
  \caption*{class n02699494}
\end{minipage}
&
\begin{minipage}[t]{0.48\textwidth}
  \centering
  \includegraphics[width=\linewidth]{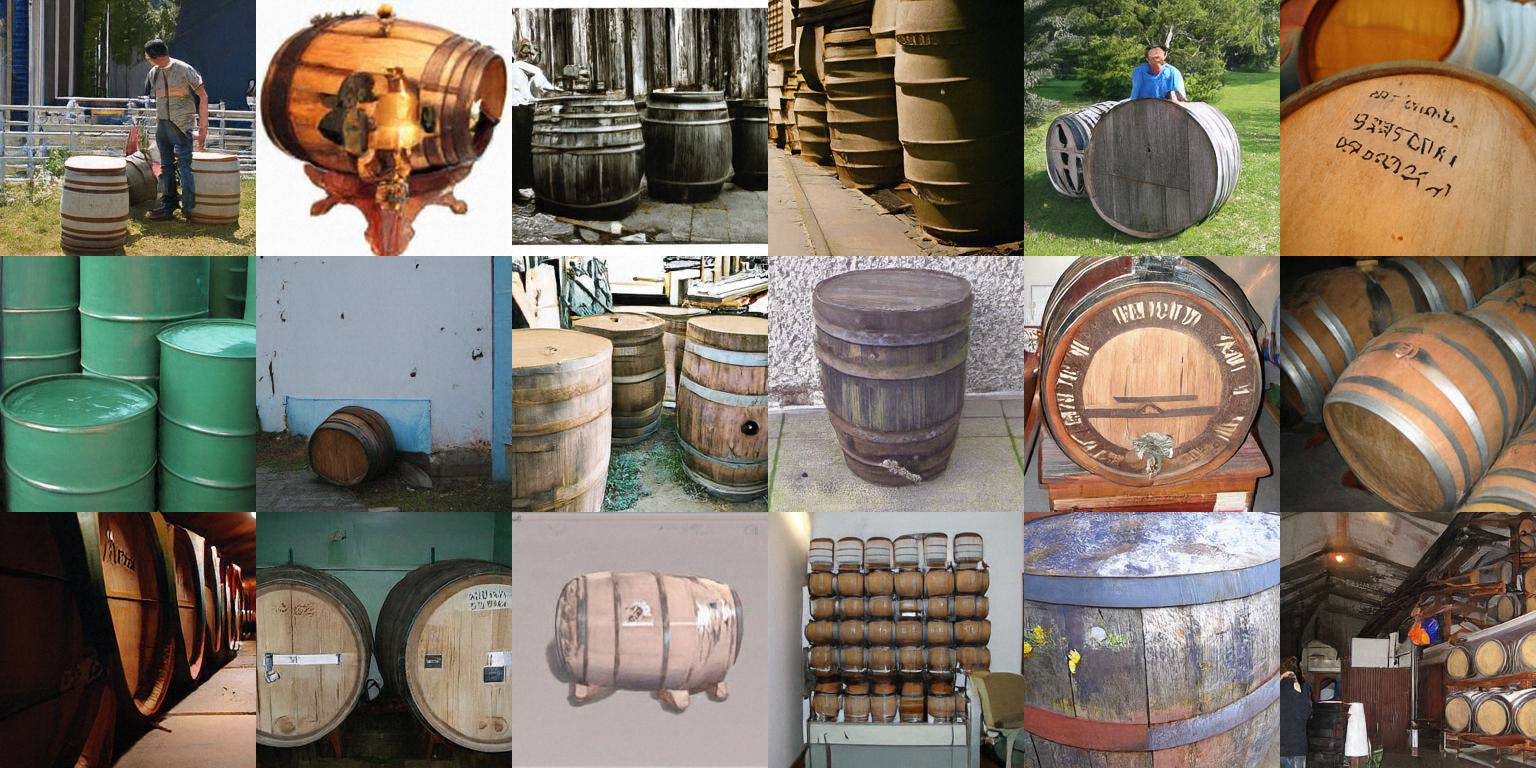}
  \vspace{-2.1em}
  \caption*{class n02795169}
\end{minipage}
\\[1.5pt] 

\begin{minipage}[t]{0.48\textwidth}
  \centering
  \includegraphics[width=\linewidth]{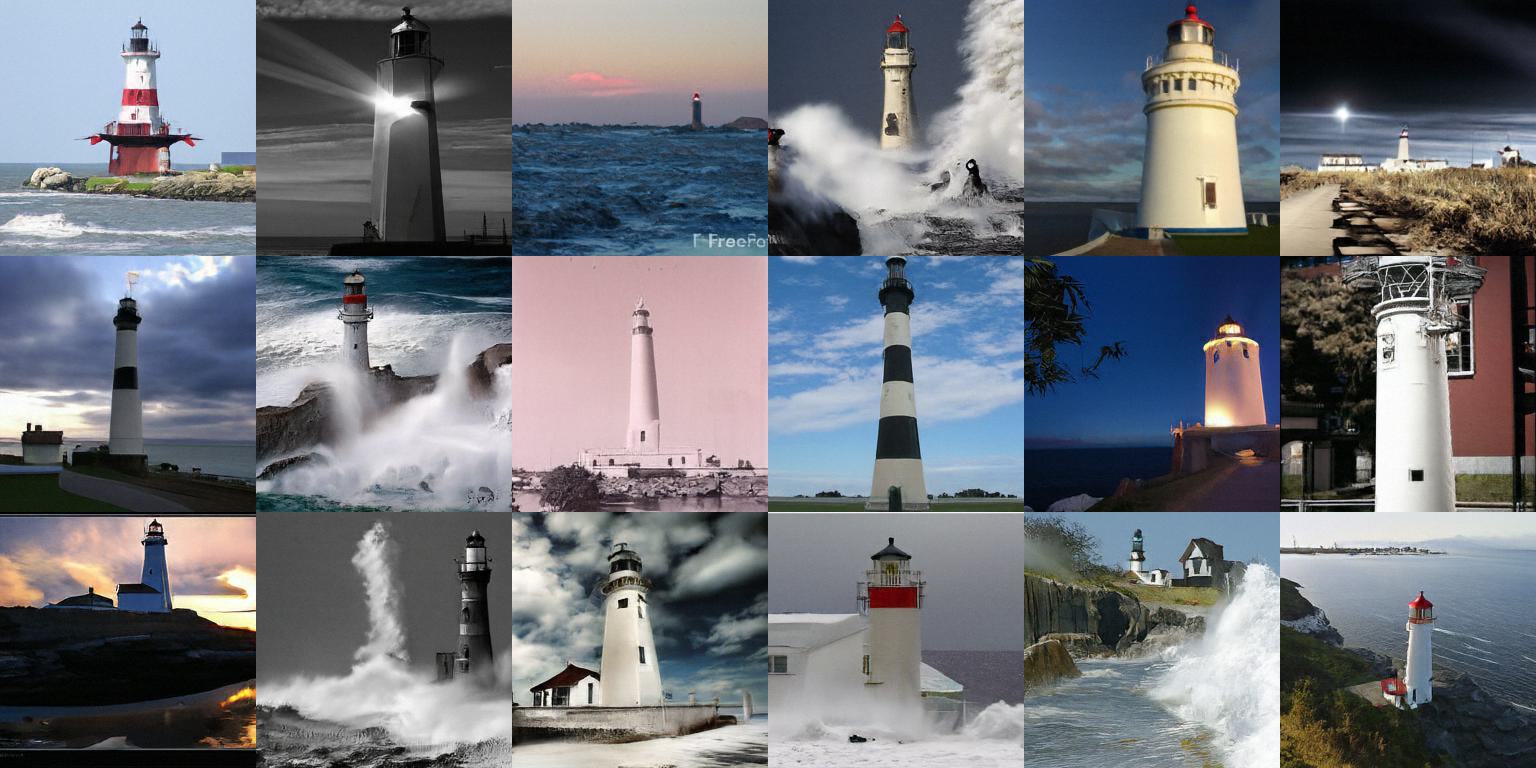}
  \vspace{-2.1em}
  \caption*{class n02814860}
\end{minipage}
&
\begin{minipage}[t]{0.48\textwidth}
  \centering
  \includegraphics[width=\linewidth]{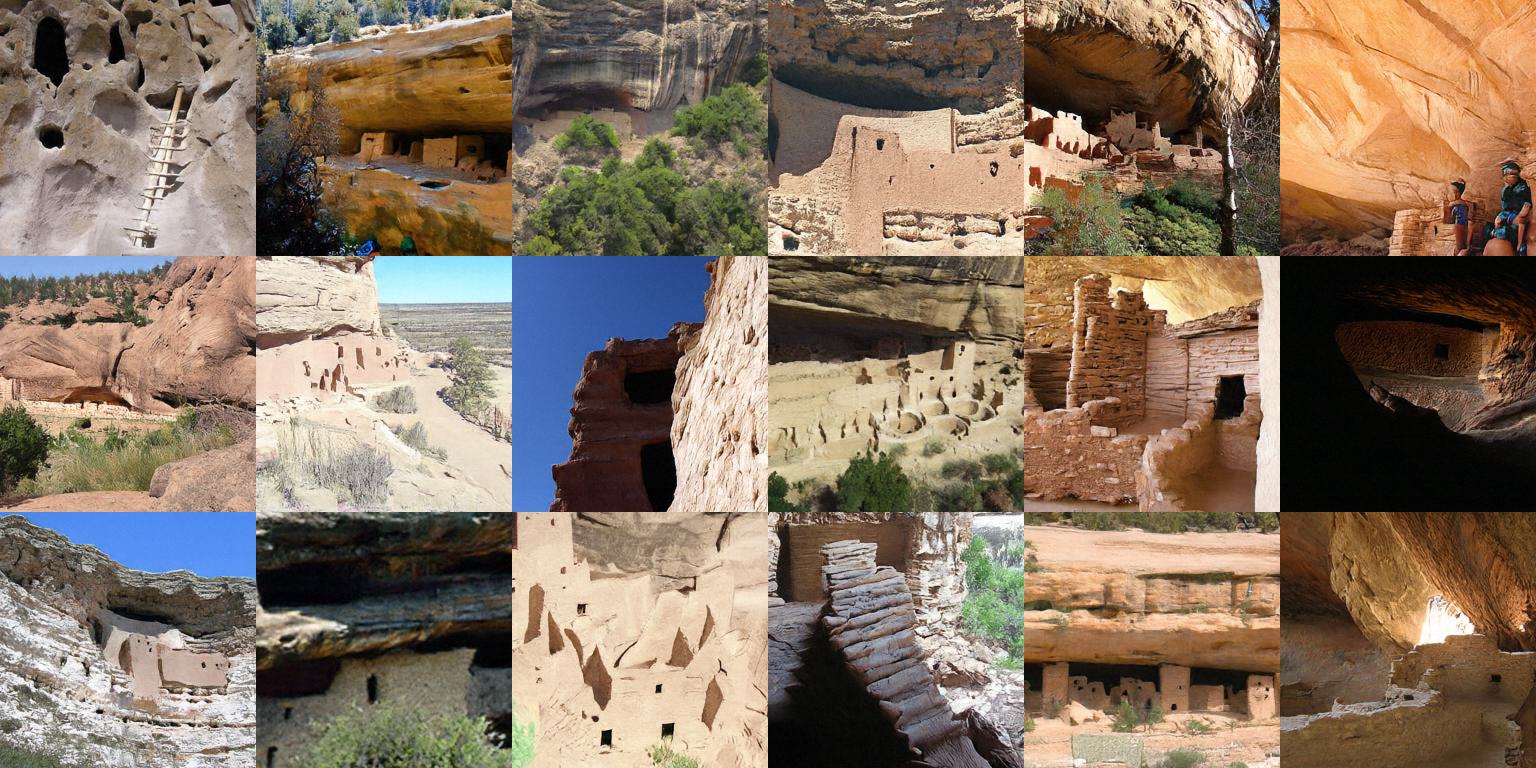}
  \vspace{-2.1em}
  \caption*{class n03042490}
\end{minipage}
\\[1.5pt] 

\begin{minipage}[t]{0.48\textwidth}
  \centering
  \includegraphics[width=\linewidth]{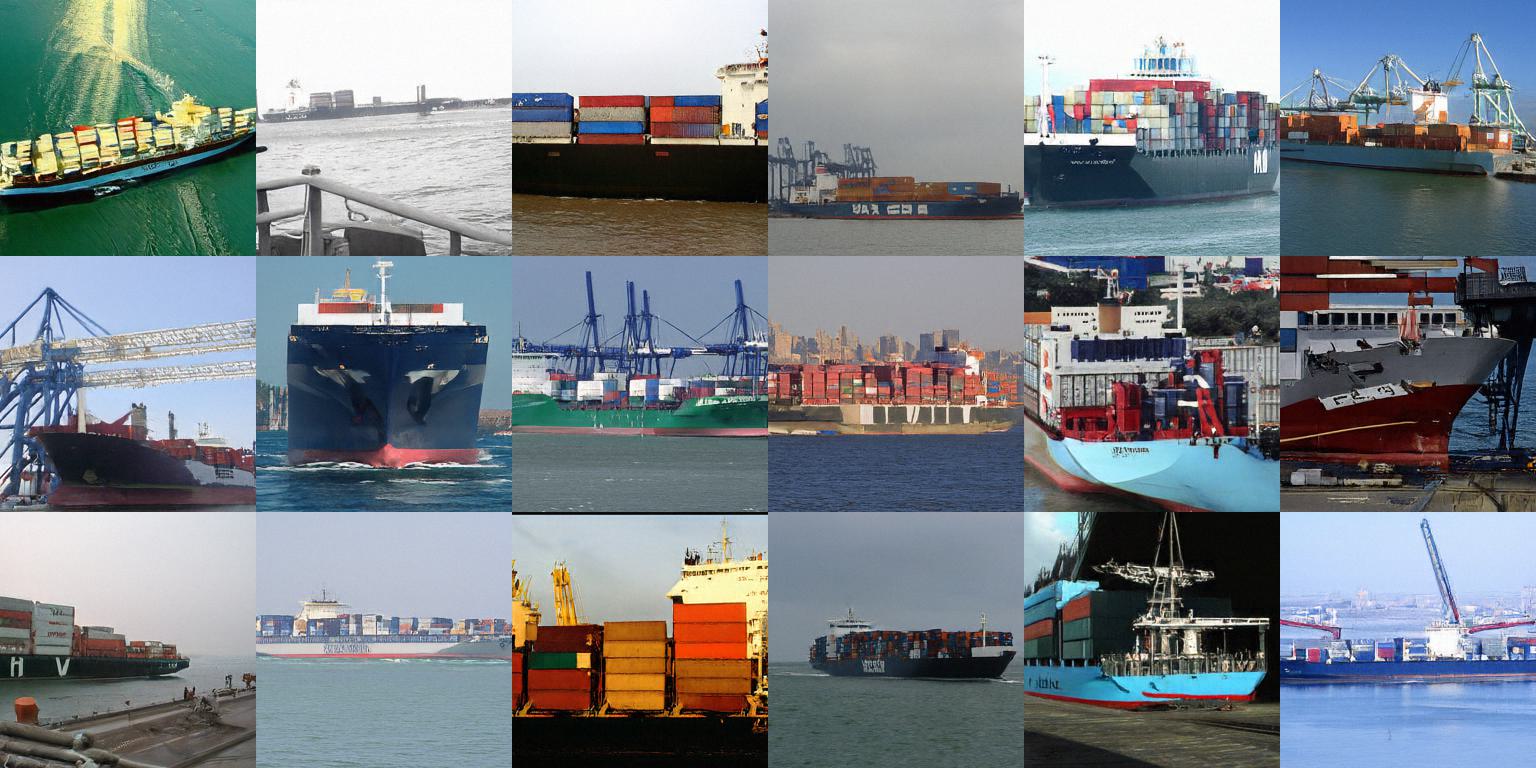}
  \vspace{-2.1em}
  \caption*{class n03095699}
\end{minipage}
&
\begin{minipage}[t]{0.48\textwidth}
  \centering
  \includegraphics[width=\linewidth]{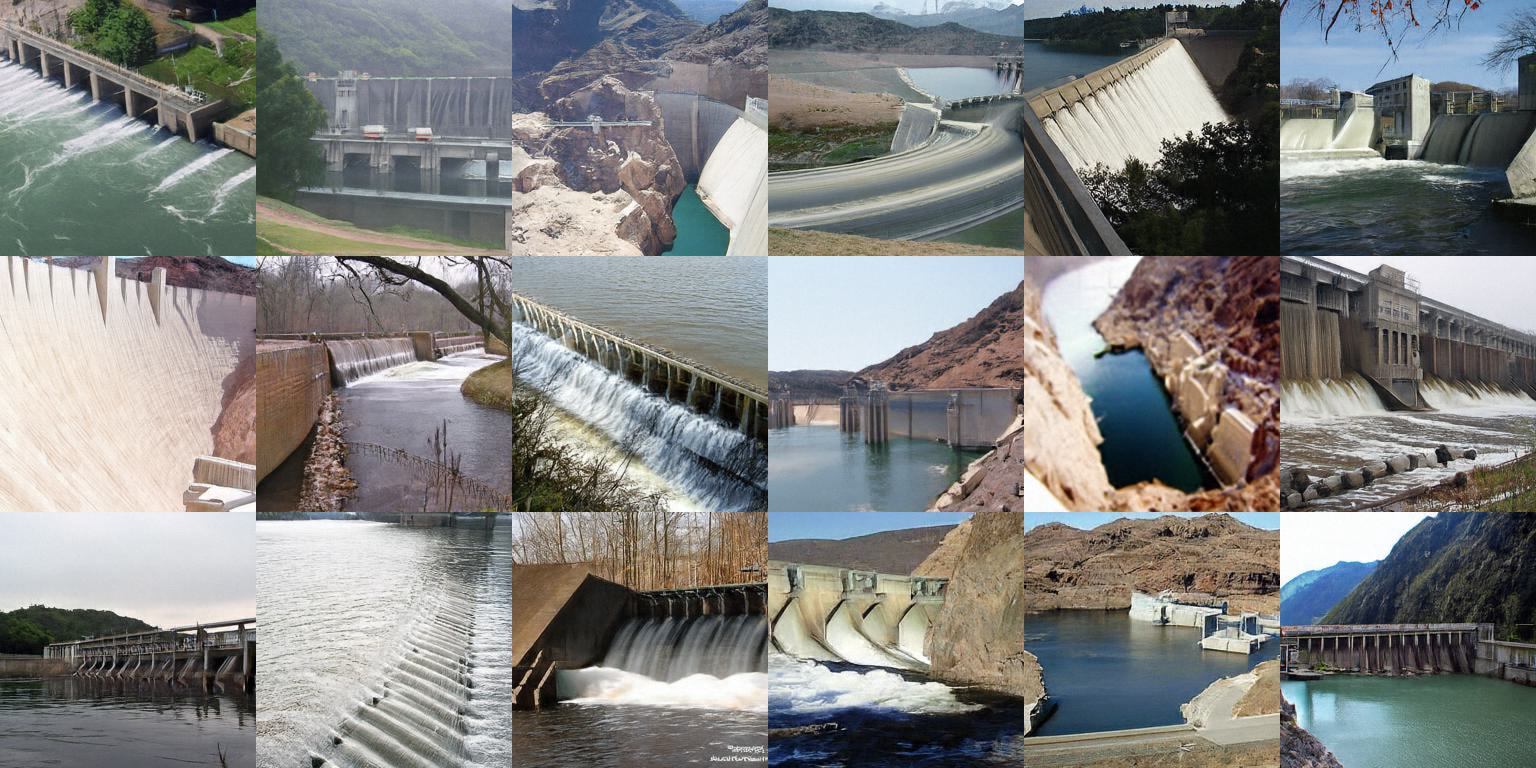}
  \vspace{-2.1em}
  \caption*{class n03160309}
\end{minipage}
\\[1.5pt] 

\end{tabular}

\caption{\textbf{Uncurated samples of PixelREPA/H-16 on ImageNet 256$\times$256.}}
\label{fig:appen_uncurated_3}
\end{figure*}

\begin{figure*}[h]
\centering
\setlength{\tabcolsep}{10pt}  
\renewcommand{\arraystretch}{1.0}

\begin{tabular}{@{}cc@{}}

\begin{minipage}[t]{0.48\textwidth}
  \centering
  \includegraphics[width=\linewidth]{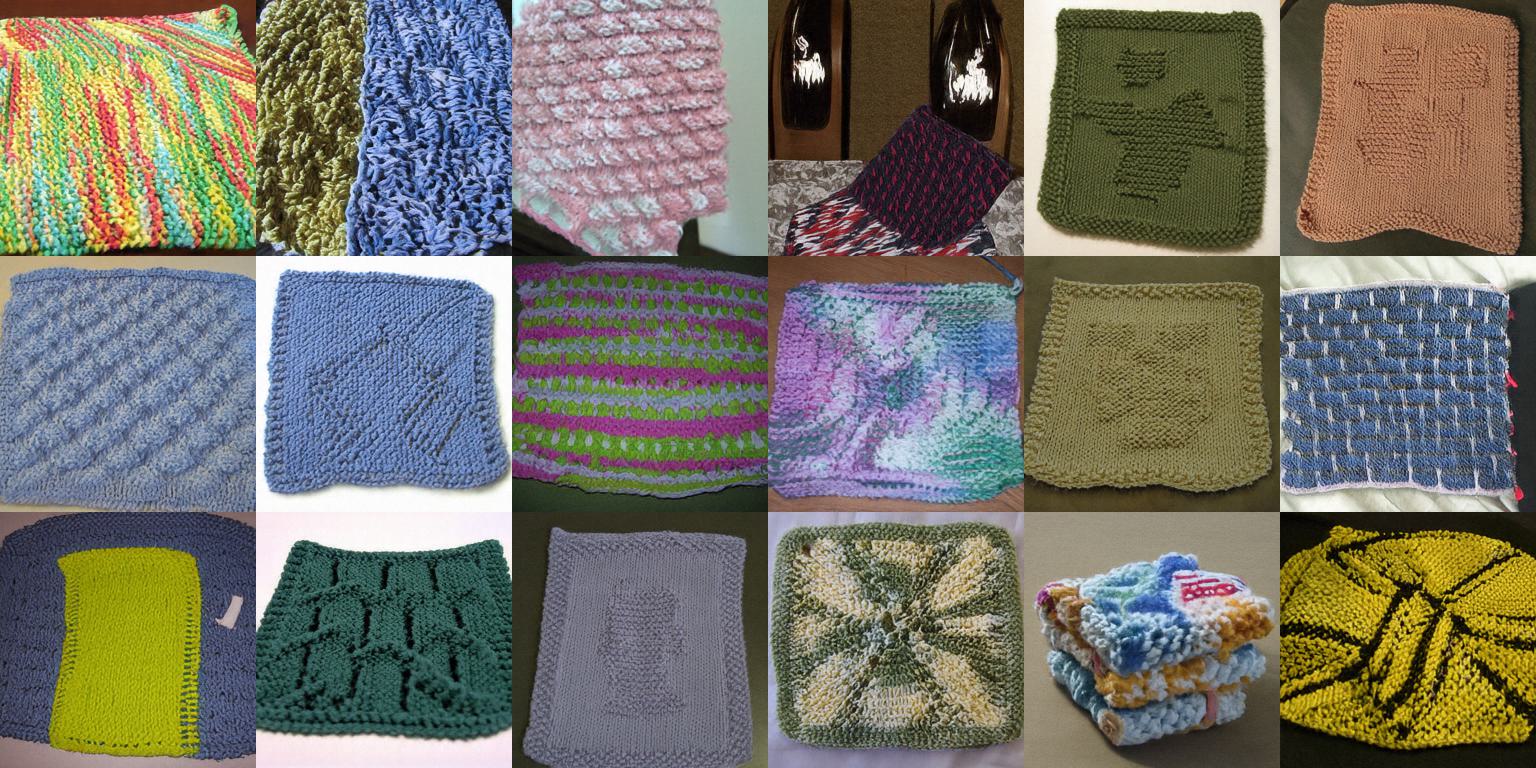}
  \vspace{-2.1em}
  \caption*{class n03207743}
\end{minipage}
&
\begin{minipage}[t]{0.48\textwidth}
  \centering
  \includegraphics[width=\linewidth]{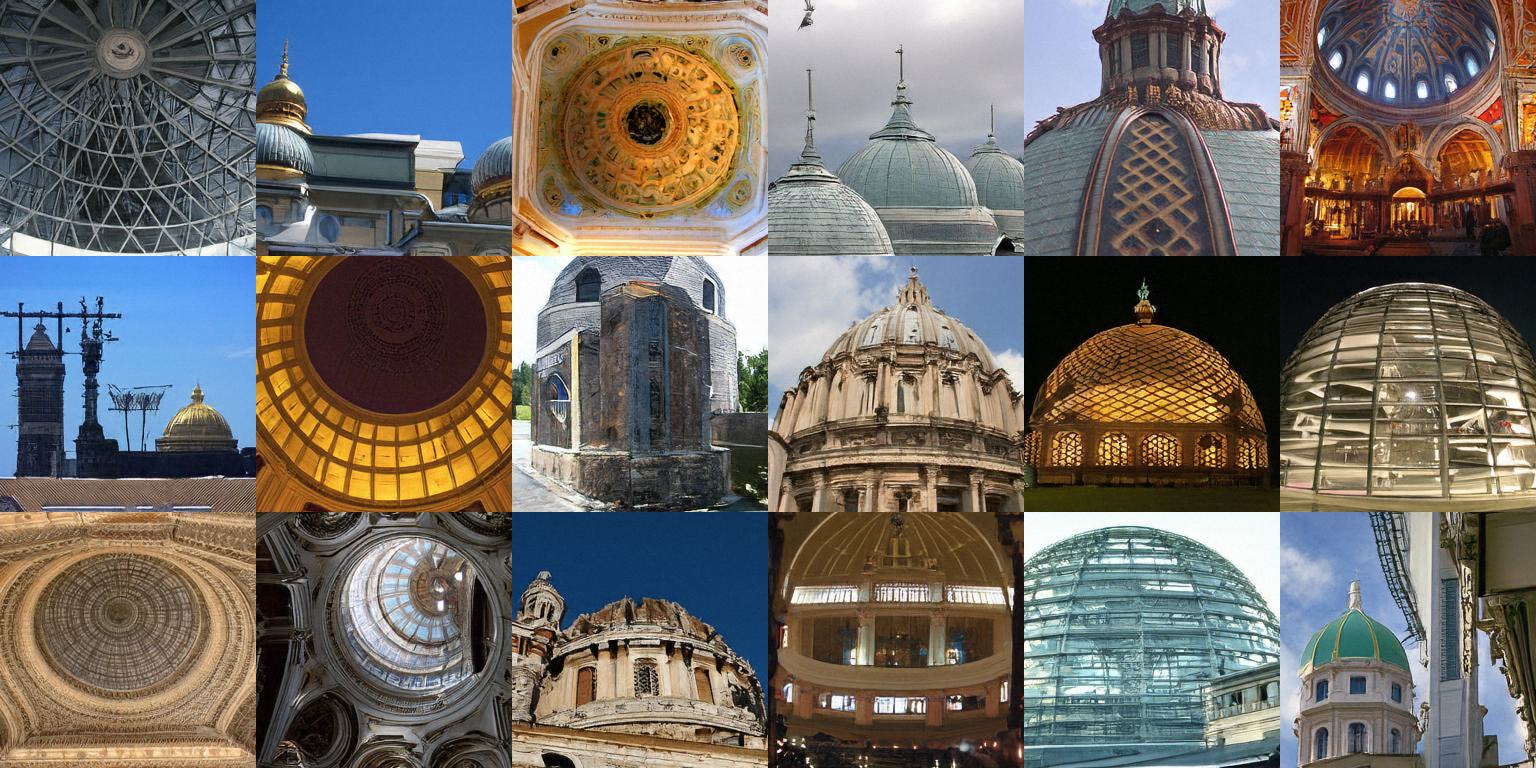}
  \vspace{-2.1em}
  \caption*{class n03220513}
\end{minipage}
\\[1.5pt] 

\begin{minipage}[t]{0.48\textwidth}
  \centering
  \includegraphics[width=\linewidth]{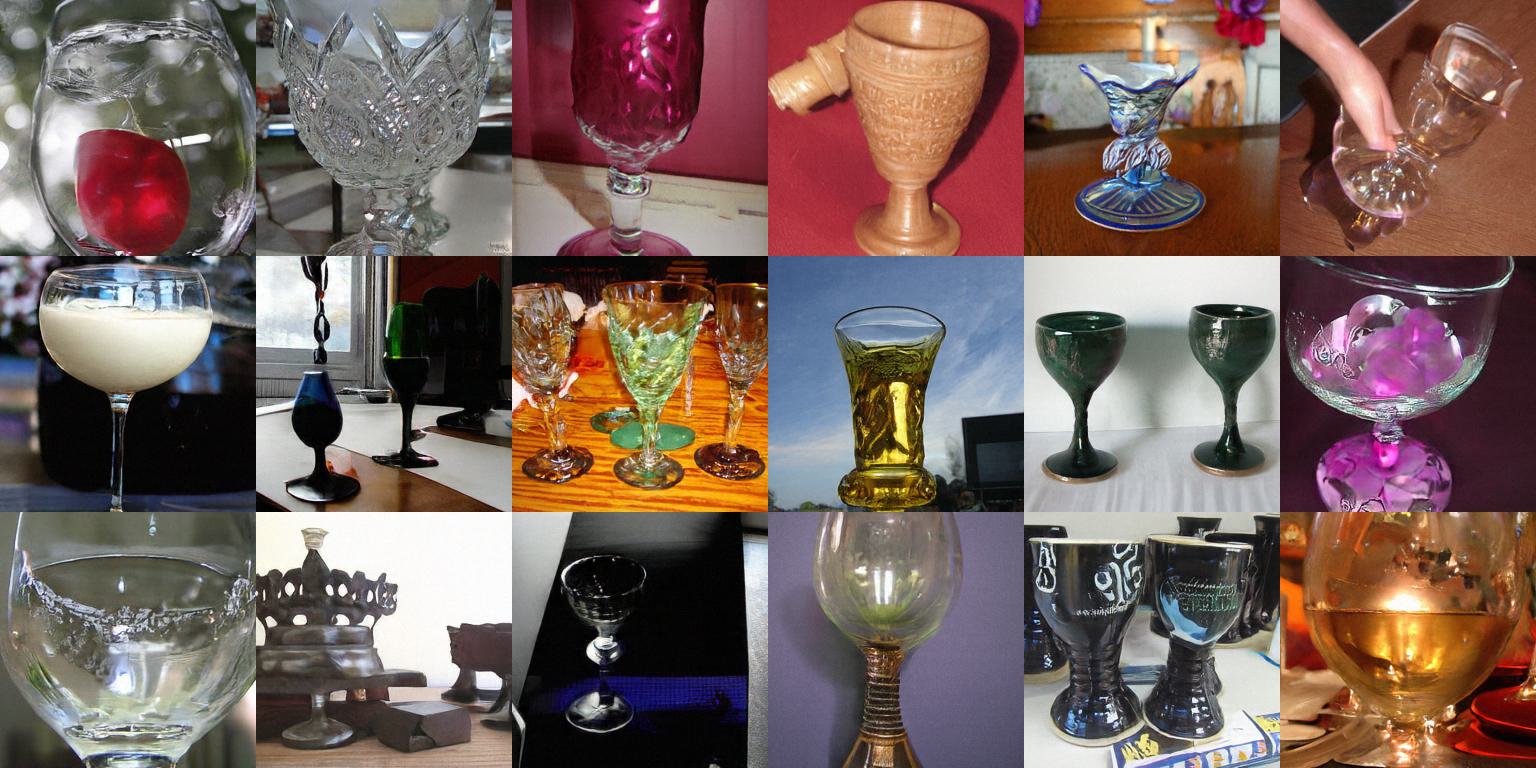}
  \vspace{-2.1em}
  \caption*{class n03443371}
\end{minipage}
&
\begin{minipage}[t]{0.48\textwidth}
  \centering
  \includegraphics[width=\linewidth]{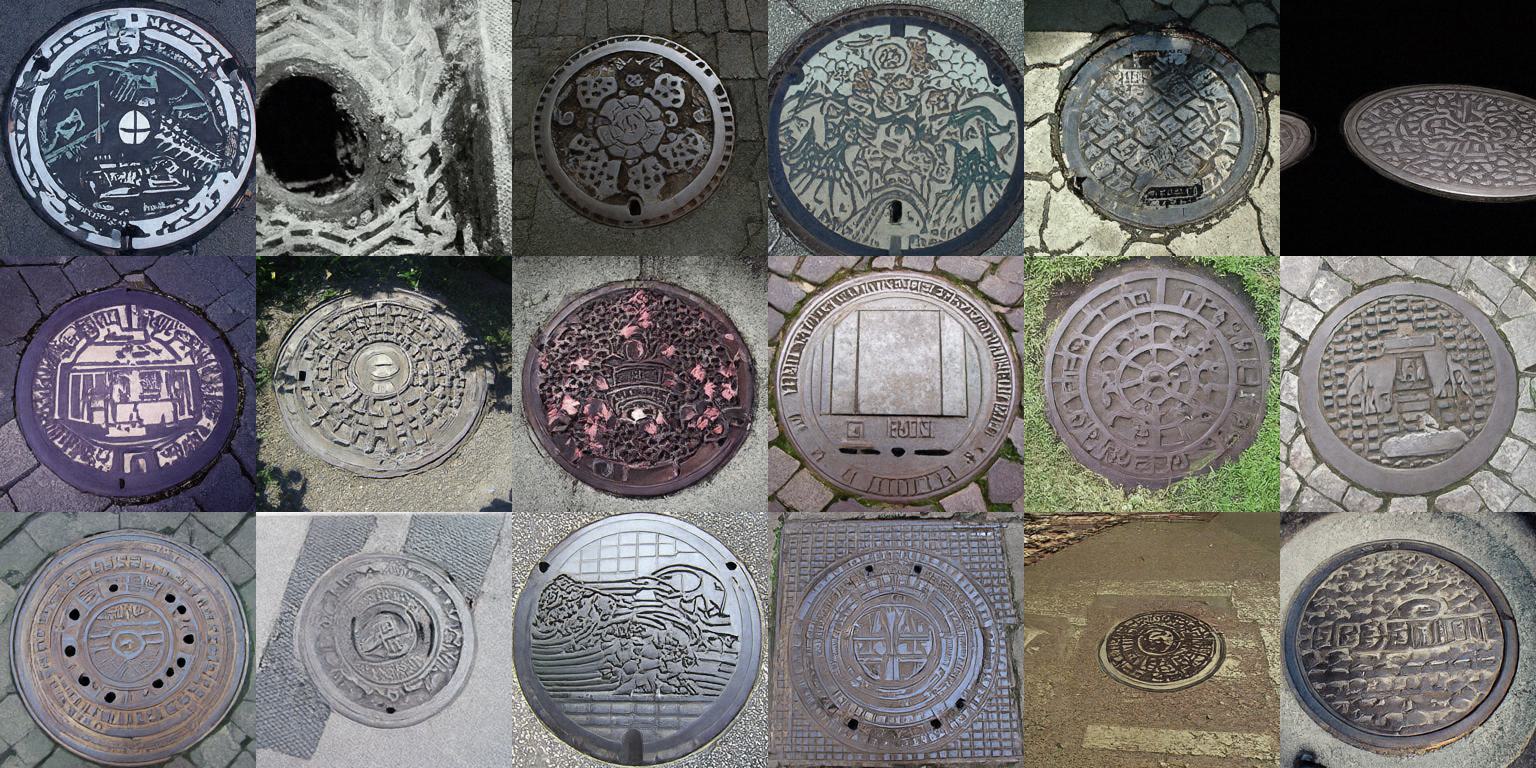}
  \vspace{-2.1em}
  \caption*{class n03717622}
\end{minipage}
\\[1.5pt] 

\begin{minipage}[t]{0.48\textwidth}
  \centering
  \includegraphics[width=\linewidth]{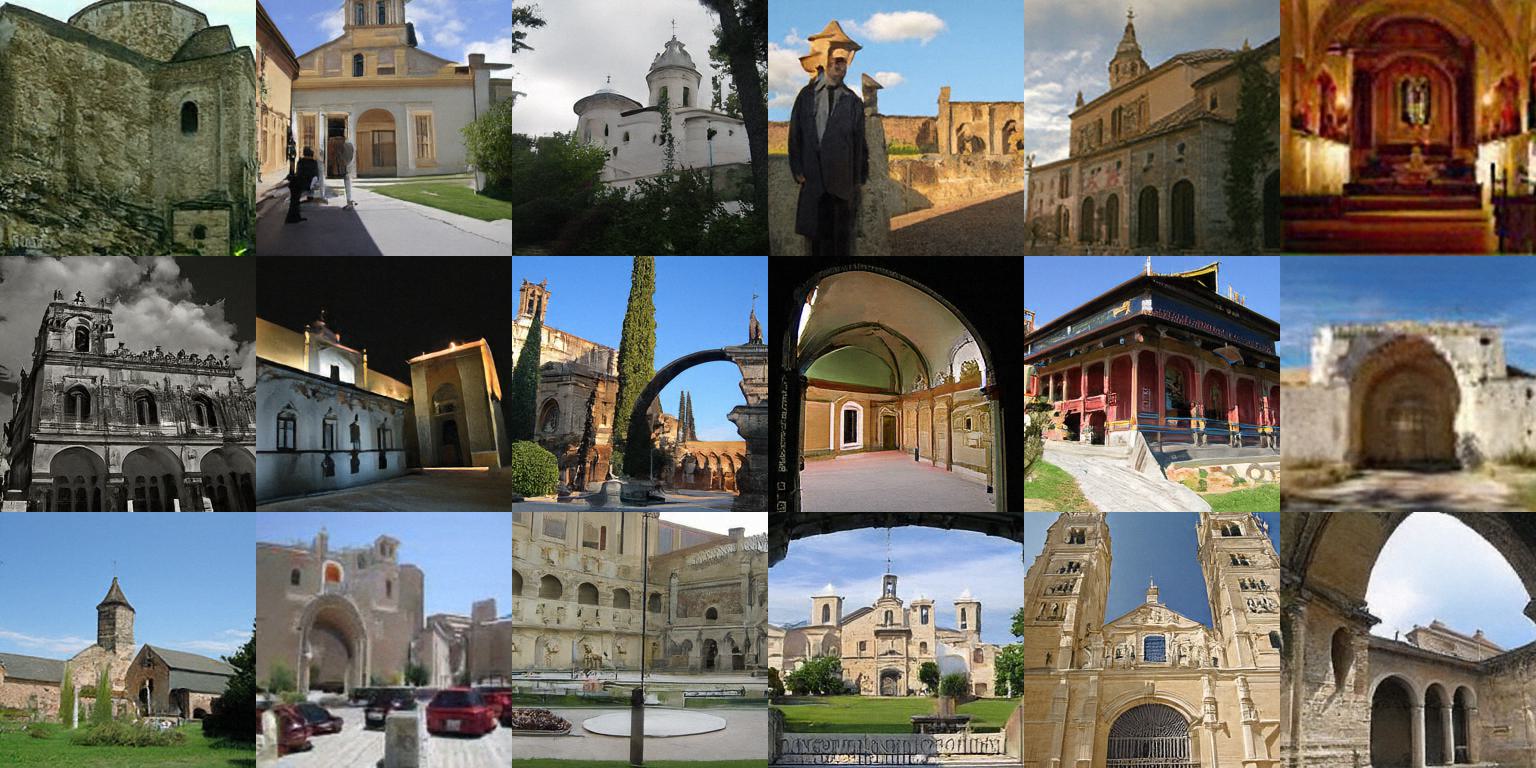}
  \vspace{-2.1em}
  \caption*{class n03781244}
\end{minipage}
&
\begin{minipage}[t]{0.48\textwidth}
  \centering
  \includegraphics[width=\linewidth]{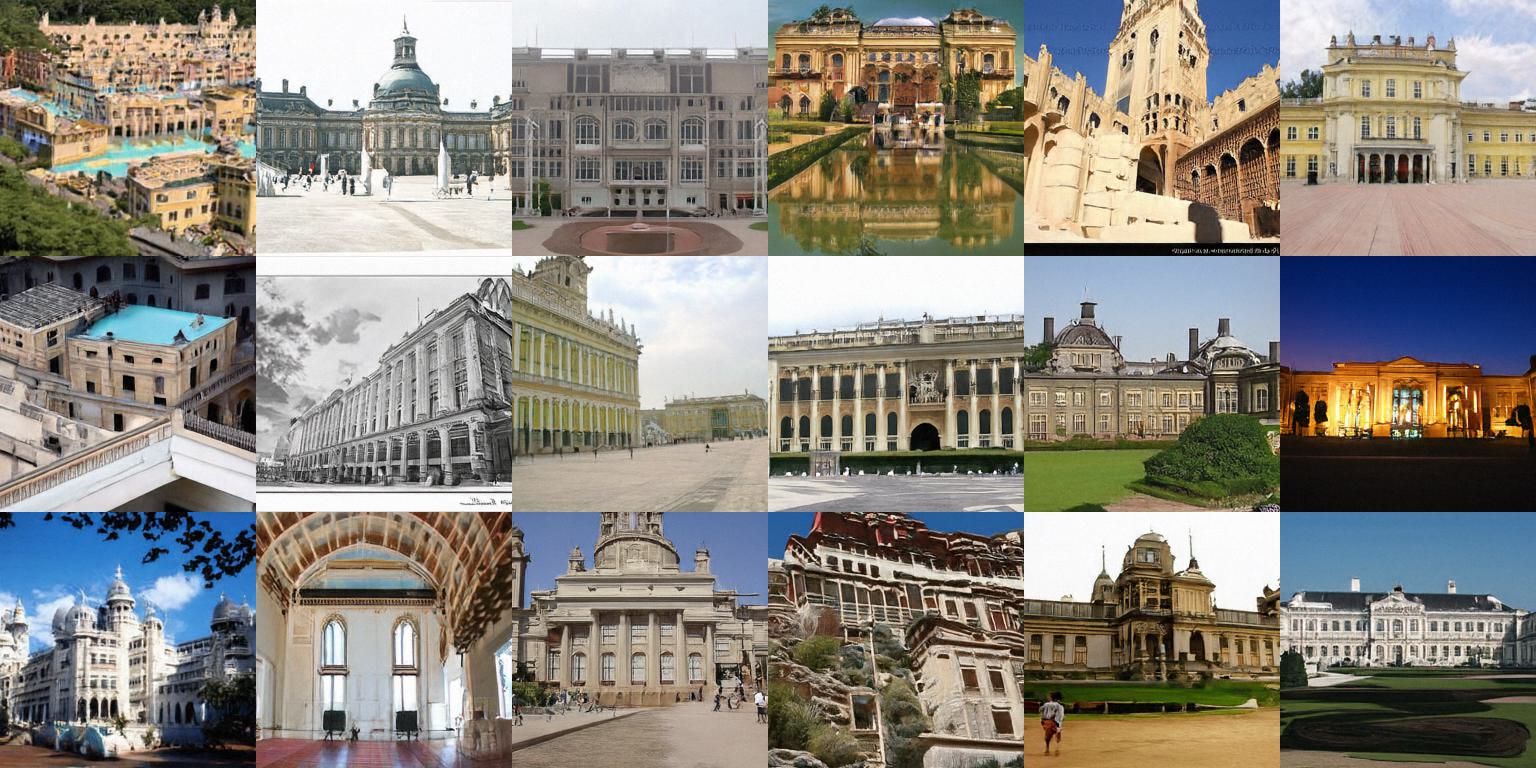}
  \vspace{-2.1em}
  \caption*{class n03877845}
\end{minipage}
\\[1.5pt] 

\begin{minipage}[t]{0.48\textwidth}
  \centering
  \includegraphics[width=\linewidth]{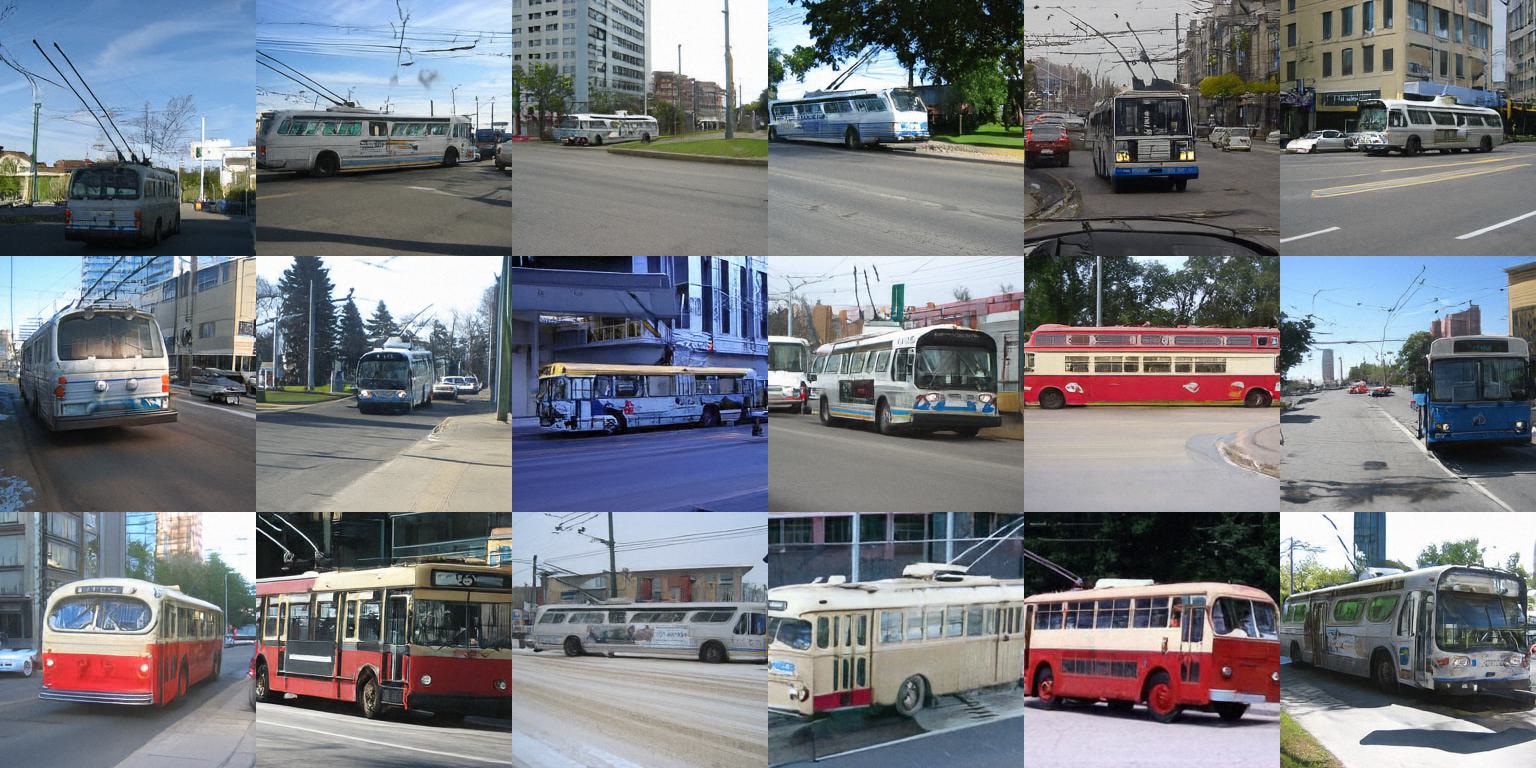}
  \vspace{-2.1em}
  \caption*{class n04487081}
\end{minipage}
&
\begin{minipage}[t]{0.48\textwidth}
  \centering
  \includegraphics[width=\linewidth]{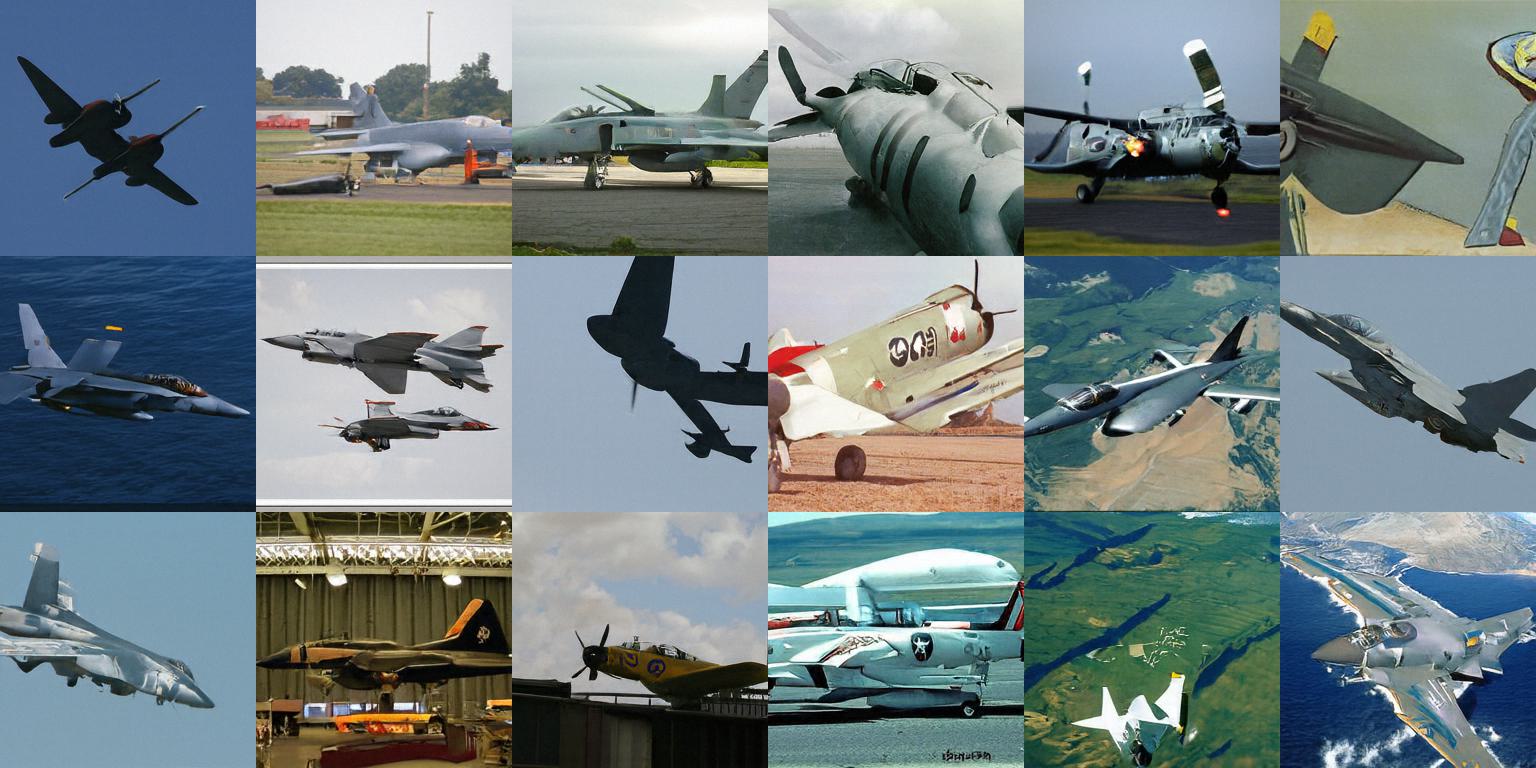}
  \vspace{-2.1em}
  \caption*{class n04552348}
\end{minipage}
\\[1.5pt] 

\begin{minipage}[t]{0.48\textwidth}
  \centering
  \includegraphics[width=\linewidth]{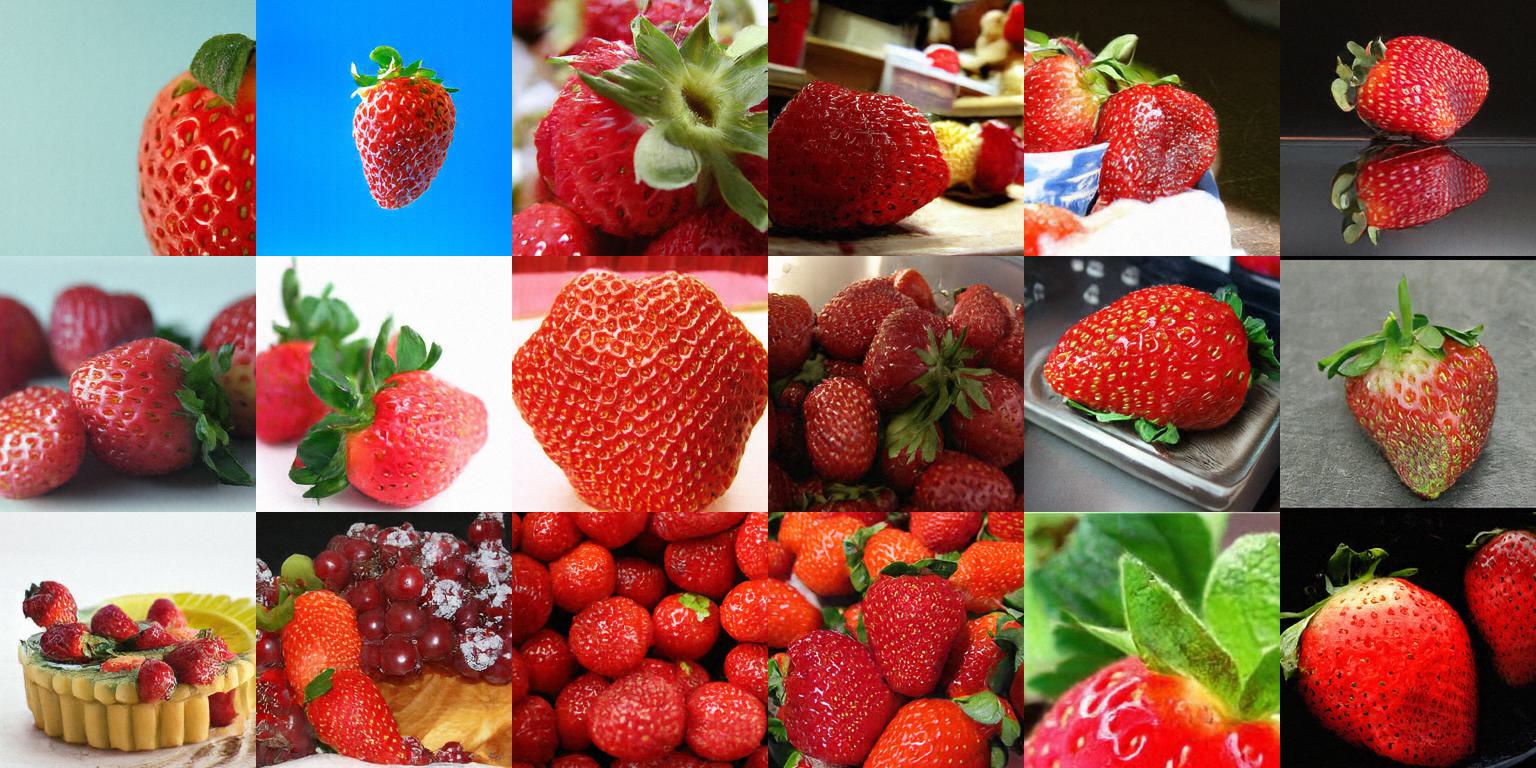}
  \vspace{-2.1em}
  \caption*{class n07745940}
\end{minipage}
&
\begin{minipage}[t]{0.48\textwidth}
  \centering
  \includegraphics[width=\linewidth]{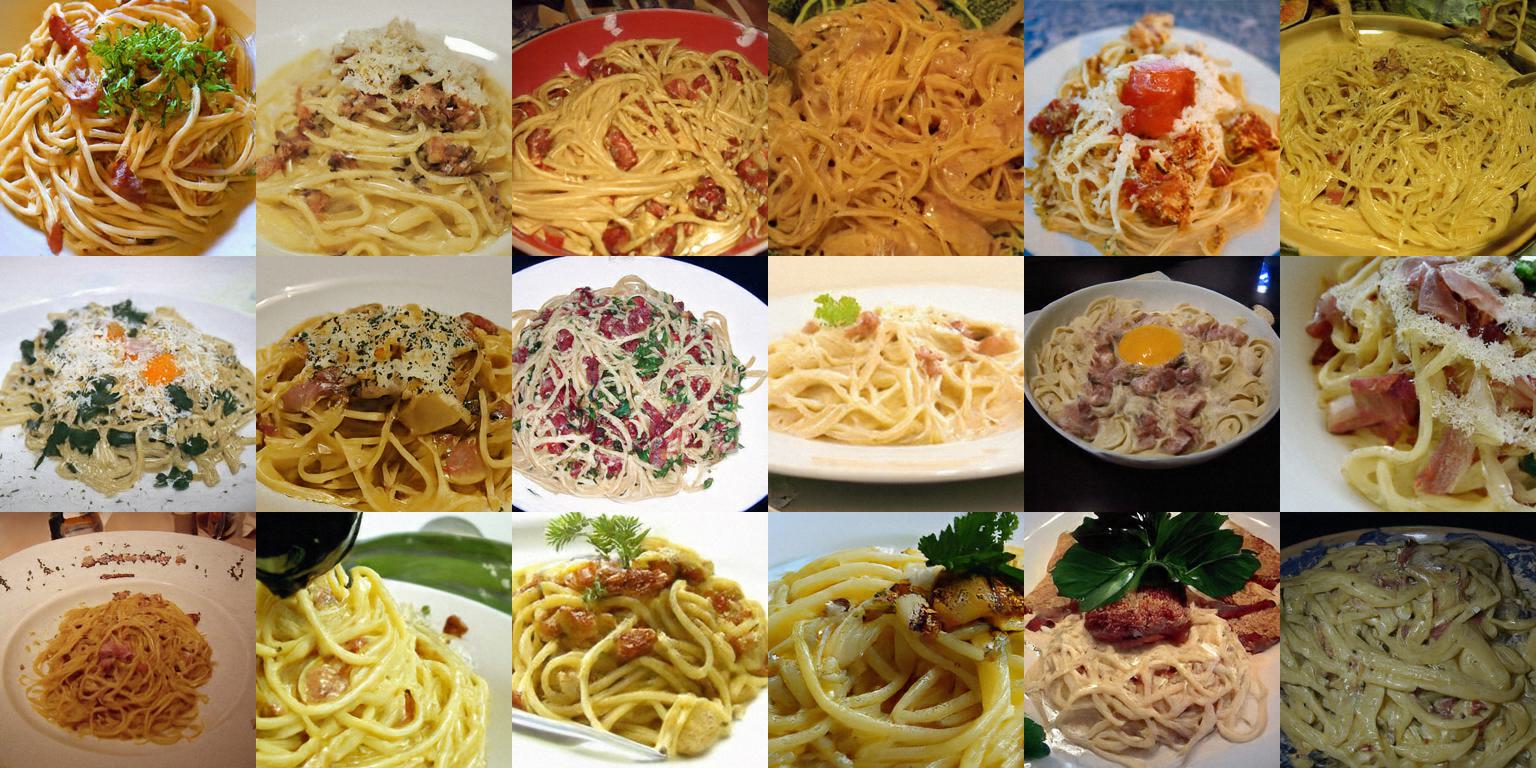}
  \vspace{-2.1em}
  \caption*{class n07831146}
\end{minipage}
\\[1.5pt] 

\end{tabular}

\caption{\textbf{Uncurated samples of PixelREPA/H-16 on ImageNet 256$\times$256.}}
\label{fig:appen_uncurated_4}
\end{figure*}

We provide uncurated qualitative results for various classes, as shown in \cref{fig:appen_uncurated_1,fig:appen_uncurated_2,fig:appen_uncurated_3,fig:appen_uncurated_4}.
These results are evaluated with PixelREPA-H and share same classifier free guidance.

